\documentclass[12pt]{article}
\usepackage{amsmath}
\usepackage{graphicx}
\usepackage{enumerate}

\usepackage{natbib} 

\newcommand{\blind}{1}

\usepackage{amsmath,amssymb,amsfonts,amsthm,bbm}

\usepackage{mathtools}   % for \norm
\usepackage{stmaryrd} % for \llbracket     1 \rrbracket

\usepackage{enumitem}   % for customized enumerate
\usepackage{romannum}

\usepackage[titletoc,title]{appendix}  % Appendix A

\usepackage[table,xcdraw,dvipsnames]{xcolor}   % for \textcolor, \colorlet, ...
\usepackage{hyperref}
\newcommand\myshade{85}
\colorlet{mylinkcolor}{YellowOrange}
\colorlet{mycitecolor}{Aquamarine}
\colorlet{myurlcolor}{violet}

\hypersetup{
  linkcolor  = mylinkcolor!\myshade!black,
  citecolor  = mycitecolor!\myshade!black,
  urlcolor   = myurlcolor!\myshade!black,
  colorlinks = true,
}

\usepackage{caption}
\usepackage{subcaption}

\usepackage[normalem]{ulem}
\usepackage{setspace}

\usepackage{graphicx}
\usepackage{comment}
\graphicspath{{figs/}}

\newcommand{\bfm}[1]{\ensuremath{\boldsymbol{#1}}} % bm

   \def\bA{\bfm A}  
     
   \def\bC{\bfm C}  
     
\def\be{\bfm e}   \def\bE{\bfm E}  \def\EE{\mathbb{E}}
\def\bff{\bfm f}

     \def\II{\mathbb{I}}

     \def\PP{\mathbb{P}}
     
     \def\RR{\mathbb{R}}

\def\bv{\bfm v}     
\def\bw{\bfm w}     
\def\bx{\bfm x}   \def\bX{\bfm X}  
\def\by{\bfm y}     
   \def\bZ{\bfm Z}  \def\ZZ{\mathbb{Z}}

\def\calA{{\cal  A}} \def\cA{{\cal  A}}

\def\calG{{\cal  G}} 
\def\calH{{\cal  H}}

\def\calK{{\cal  K}} \def\cK{{\cal  K}}
 
\def\calM{{\cal  M}} 
\def\calN{{\cal  N}} 
 
\def\calP{{\cal  P}} 
\def\calQ{{\cal  Q}} \def\cQ{{\cal  Q}}
 
\def\calS{{\cal  S}} \def\cS{{\cal  S}}
\def\calT{{\cal  T}}

\def\calX{{\cal  X}} \def\cX{{\cal  X}}

\newcommand{\bfsym}[1]{\ensuremath{\boldsymbol{#1}}}

 \def\balpha{\bfsym \alpha}
 \def\bbeta{\bfsym \beta}
              
 \def\bdelta{\bfsym {\delta}}

              \def\bSigma{\bfsym \Sigma}

 \def\bxi{\bfsym {\xi}}

\def\bphi{\bfsym {\phi}}          \def\bPhi{\bfsym {\Phi}}

                  \def\hbbeta{\hat{\bfsym \beta}}

\providecommand{\abs}[1]{\left\lvert#1\right\rvert}
\providecommand{\norm}[1]{\left\lVert#1\right\rVert}

\providecommand{\paren}[1]{\left( #1 \right)}
\providecommand{\paran}[1]{\left( #1 \right)}
\providecommand{\brackets}[1]{\left[ #1 \right]}
\providecommand{\braces}[1]{\left\{ #1 \right\}}

\DeclarePairedDelimiter\ceil{\lceil}{\rceil}

\providecommand{\defeq}{\triangleq}

\usepackage{mathtools}
\DeclarePairedDelimiterX{\infdivx}[2]{(}{)}{%
  #1 \; \delimsize\| \; #2%
}

\DeclareMathOperator{\argmin}{argmin}

\newcommand{\E}[1]{{\mathbb{E}} \left[ #1 \right]}

\newtheorem{definition}{Definition}[section]
\newtheorem{assumption}[definition]{Assumption}
\newtheorem{lemma}[definition]{Lemma}

\newtheorem{theorem}[definition]{Theorem}

\newcommand{\Op}[1]{{\mathcal{O}_p} \left( #1 \right) }

\definecolor{royalpurple}{rgb}{0.47, 0.32, 0.66}

\def\lam {\lambda}

\renewcommand{\hat}{\widehat}

\usepackage{setspace}
\usepackage{etoolbox}
\BeforeBeginEnvironment{equation*}{\begin{singlespace}\vspace*{-\baselineskip}}
	\AfterEndEnvironment{equation*}{\end{singlespace}\noindent\ignorespaces}
\BeforeBeginEnvironment{equation}{\begin{singlespace}\vspace*{-\baselineskip}}
	\AfterEndEnvironment{equation}{\end{singlespace}\noindent\ignorespaces}
\BeforeBeginEnvironment{align}{\begin{singlespace}\vspace*{-\baselineskip}}
	\AfterEndEnvironment{align}{\end{singlespace}\noindent\ignorespaces}
\BeforeBeginEnvironment{align*}{\begin{singlespace}\vspace*{-\baselineskip}}
	\AfterEndEnvironment{align*}{\end{singlespace}\noindent\ignorespaces}
\BeforeBeginEnvironment{eqnarray}{\begin{singlespace}\vspace*{-\baselineskip}}
	\AfterEndEnvironment{eqnarray}{\end{singlespace}\noindent\ignorespaces}
\BeforeBeginEnvironment{eqnarray*}{\begin{singlespace}\vspace*{-\baselineskip}}
	\AfterEndEnvironment{eqnarray*}{\end{singlespace}\noindent\ignorespaces}

\usepackage[framemethod=TikZ]{mdframed} % box environment & flexible box designs

\usepackage{fancybox}

\usepackage[linesnumberedhidden,ruled,vlined]{algorithm2e}

\SetCommentSty{mycommfont}

\SetKwInput{KwInput}{Input}                % Set the Input
\SetKwInput{KwOutput}{Output}              % set the Output

\usepackage{listings}             % Include the listings-package
\newtheorem{rmk}{Remark}

\newcommand*{\rom}[1]{\uppercase\expandafter{\romannumeral #1\relax}} %

\begin{document}
\pagenumbering{arabic}

\def\spacingset#1{\renewcommand{\baselinestretch}%
{#1}\small\normalsize} \spacingset{1}

\def\cond{\;|\;}

%---------------------------------------------------
%
% Title page
%
% https://www.math.uh.edu/~torok/math_6298/latex/MANUALS/NASA_Hypertext-Help-with-LaTeX/latex/ltx-407.html
%

%\def\TITLE{Knowledge Transfer in Dynamic Decisions with Fitted $Q$-Learning}
%\def\TITLE{Data-Driven Knowledge Transfer in Infinite-Horizon Markov Decision Process}
\def\TITLE{Data-Driven Knowledge Transfer in \\ Batch $Q^*$ Learning}

\if1\blind
{
  \title{\bf \TITLE}
  \author{
  Elynn Chen$^1$ \hspace{6ex}
  Xi Chen$^1$ \thanks{Corresponding author.
  Email: \textit{xchen3@stern.nyu.edu}} \hspace{6ex}
  Wenbo Jing$^2$ \hspace{6ex}
  %(alphabet) $^\dag$ 
  \\  {\normalsize
  $^{1}$Stern School of Business, New York University, New York, NY}\\
  { \normalsize $^2$College of Business, City University of Hong Kong, Kowloon, Hong Kong}
  }
  \date{}
  \maketitle
} \fi

\if0\blind
{
  \bigskip
  \bigskip
  \bigskip
  \begin{center}
    {\LARGE\bf \TITLE}
 \end{center}
  \date{}
 \vspace{10em}
  \medskip
} \fi

\bigskip
\begin{abstract}
\spacingset{1.18}
%\begin{comment}
In data-driven decision-making across marketing, healthcare, and education, leveraging large datasets from existing ventures is crucial for navigating high-dimensional feature spaces and addressing data scarcity in new ventures. We investigate knowledge transfer in dynamic decision-making by focusing on batch stationary environments and formally defining task discrepancies through the framework of Markov decision processes (MDPs). We propose the Transfer Fitted $Q$-Iteration algorithm with general function approximation, which enables direct estimation of the optimal action-state function $Q^*$ using both target and source data. Under sieve approximation, we establish the relationship between statistical performance and the MDP task discrepancy, highlighting the influence of source and target sample sizes and task discrepancy on the effectiveness of knowledge transfer. Our theoretical and empirical results demonstrate that the final learning error of the function is significantly reduced compared to the single-task learning rate.

%\end{comment}
\end{abstract}

\noindent%
\vfill

%%---------------------------------------------------
%%
%% Main Text
%5

\newpage
\spacingset{1.78} % DON'T change the spacing!

\addtolength{\textheight}{.1in}%

\section{Introduction}  \label{sec:intro}

Data-driven sequential decision-making is gaining widespread prominence in real-world applications, including marketing \citep{liu2022dynamic}, healthcare \citep{komorowski2018artificial}, and education \citep{rafferty2016faster}.
A primary challenge in these areas is managing high-dimensional feature spaces, especially when personalizing services or navigating complex domains. 
Furthermore, societal applications often face a significant data scarcity issue when venturing into new locations, targeting different population groups, or introducing new products or services. 
Data scarcity, marked by high dimensionality or a lack of historical data, demands innovative methods for data aggregation and automatic knowledge transfer.
To tackle this crucial challenge, we introduce a knowledge transfer framework designed for data-driven sequential decision-making. This method can accelerate learning in a specific decision-making task by utilizing related source tasks from large-scale observational or simulated datasets. 

The formal study of data-driven sequential decision-making is conducted within the broad framework of \textit{reinforcement learning} (RL) \citep{sutton2018reinforcement}. %{levine2020offline}.  
Within RL, numerous model assumptions and methods exist for estimation and decision-making. 
In this paper, we focus on sample-transferred estimation of the optimal action-value function, i.e., the $Q^*$ function, for stationary \textit{Markov decision processes} (MDPs).%, employing the classic framework of {\em Fitted $Q$-Iteration} (FQI)  \citep{ernst2005tree}.

The literature lacks a thorough examination of \emph{transfer learning} (TL) for RL with $Q^*$ estimation that is supported by theoretical guarantees. We pioneer this investigation by first delineating the transferred RL problem within the framework of MDPs, where we formally define the \emph{RL task discrepancy} based on differences in reward functions and transition probabilities. To facilitate a transfer algorithm that aims at direct estimation of the $Q^*$ function, we derive a theoretical result that explicates the relationship between task discrepancy and the divergence of $Q^*$ functions of different MDPs. 
Given that reward functions and transition probabilities can be readily estimated in practice, this also guarantees the tangible efficacy of transferring MDP tasks that display minor discrepancies.
%This precise definition of the TL for RL, along with our theoretical finding, lays the groundwork for both the development of an estimation procedure and subsequent theoretical analysis. 

Based on this formal characterization, we introduce a general framework of Transfer FQI algorithm (Algorithm \ref{algo:fitted-q-transfer}). 
It is, in essence, an {\em iterative fixed-point algorithm} with knowledge transfer built on general function approximations.
While it adopts the core transfer-learning concept of initial learning of commonalities followed by adjustments for idiosyncratic biases, the Transfer FQI algorithm notably diverges from current transfer algorithms applied in supervised or unsupervised learning in two principal respects.
First, the Transfer FQI adopts an iterative approach to knowledge transfer, in contrast to the one-off nature of transferred learning in supervised or unsupervised settings. This necessitates meticulous attention to mitigate estimation biases and transfer-induced errors across each iteration, a challenge not previously tackled by non-iterative transferred learning algorithms in existing literature.
Second, due to its self-iterative nature, the response variables in FQI is not observable and needs to be re-constructed for each task in each iteration using the estimators obtained in the previous step. Therefore, to enable the benefit of transfer learning, we take extra steps to simultaneously build improved estimators for both the target and source tasks. This contrasts with the conventional focus of transferred supervised learning algorithms, which  aim to refine estimations solely for the target task.

For theoretical analysis, we instantiate the general framework using semi-parametric sieve approximation, which is widely employed in societal applications, and establish rigorous theoretical guarantees. The developed theoretical analysis framework for an iterative fixed-point algorithm with knowledge transfer generally applies to other types of function approximation and similarity characterization. We first show that when the transition dynamics are shared across tasks, the regret of our algorithm decomposes into three components: the approximation bias determined by the number of sieve basis functions, the commonality estimation error depending on the total sample size across all tasks, and the task-difference bias arising from discrepancies in reward functions. This decomposition provides a precise insight that the knowledge transfer yields improvement whenever the total source sample size
is larger than the target sample size, and the discrepancy level between the reward functions is sufficiently small such that bias correction is estimable from the limited target data.

We further extend this analysis to the transition-heterogeneous setting, where both reward and transition kernels differ across tasks. In this more general case, the task-difference bias is characterized jointly by the reward and transition discrepancies across tasks. The resulting regret rate exhibits the same structure as in the homogeneous case, but with the task-difference bias term inflated by a heterogeneity factor. Importantly, our analysis provides explicit sample size conditions under which the task-difference bias becomes dominated by the commonality estimation error term, ensuring that the transfer benefit is preserved. Moreover, we develop a data-driven procedure for selecting the number of sieve basis functions that automatically balances estimation variance and approximation bias without requiring knowledge of underlying smoothness parameters, and we show that this procedure simultaneously guards against negative transfer when source tasks are insufficiently similar. Both synthetic and real-world experiments show that our proposed method consistently outperforms single-task and naive aggregation baselines, especially when source tasks are informative and task discrepancy is moderate.%Together, these results deliver a complete, verifiable, and practically implementable theory for when and how transfer learning can provably improve offline reinforcement learning.

\subsection{Literature and Organization}

This paper is situated at the intersection of two bodies of literature: batch reinforcement learning and transfer learning. 
The literature on reinforcement learning is broad and vast. 
The readers are referred to \cite{sutton2018reinforcement} for comprehensive reviews of RL.  
We review only the most relevant studies with theoretical guarantees. 

\medskip
\noindent
\textbf{Batch Reinforcement Learning.}  
We work under the setting of batch reinforcement learning \citep{chen2019information,xie2021batch,shi2022pessimistic, foster2022offline,yan2022model,li2024settling,jia2024offline}, where a sufficient amount of {\em source} data, usually a set of transitions sampled from the {\em source} MDP,  is available. 
\textit{Fitted Q-Iteration} (FQI) is an iterative framework that is the prototype of many batch RL algorithms. 
%For example, the empirically successful Deep Q-Network (DQN) can be reduced to the neural FQI algorithm \citep{riedmiller2005neural} under the technique of experience replay and function approximation with ReLU deep neural network. %\citep{nair2010rectified}.
%FQI is perhaps the most popular algorithm in batch RL due to its fast convergence and stability \citep{lange2012batch}. 
\cite{murphy2005generalization} and \cite{munos2008finite} established the finite sample bounds for FQI for a general class of regression functions.
Various variations of FQI have been studied in the literature. 
For example, %\cite{antos2007fitted} %and \cite{massoud2009regularized} 
%studied FQI with continuous action space; \cite{tosatto2017boosted} considered ensemble learning with FQI; \cite{geist2019theory} proposed FQI with entropy regularization; 
\cite{chen2019information} and \cite{xie2020q,xie2021batch} studied the necessity of assumptions for polynomial sample complexity and developed an algorithm under relaxed assumptions, and \cite{fan2020theoretical} studied the Deep Q-Network (DQN) algorithm from both algorithmic and statistical perspectives. 

All the current literature in FQI considers only a single RL task.  In addition, recent federated RL work \citep{yang2023federated, zheng2024federated} focused on multi-agent cooperation rather than cross-task knowledge transfer.
Quite differently, the present paper considers the assistance of multiple RL source tasks to a target task with rigorous formulation, estimation algorithm, and theoretical guarantees. 
The iterative nature of FQI, together with semi-parametric function approximation and penalization, brings new challenges.
%In the trivial case where the source tasks are empty, the statistical rate of our proposed algorithm (Algorithm \ref{algo:fitted-q-transfer}) recovers the statistical convergence rate of non-parametric estimation on a single RL task. 
%When the source sample size is larger than the target sample size, i.e., $n_{\calK} \gtrsim n_0$, the statistical rate becomes sharper as long as the task discrepancy satisfies a mild condition. % $h_r \lesssim \frac{p}{\sqrt{n_0\log p}}$. 
%Therefore, for a fixed target task, our proposed algorithm achieves improvement by transferring as long as the source tasks are sufficiently similar to the target task and the source sample size is sufficiently large compared to the target sample size.% We note that the above requirement for the discrepancy, $h_r \lesssim \frac{p}{\sqrt{n_0\log p}}$, is mild, {\red since the number of basis $p$ is allowed to be as large as $n_{\calK}$}, which allows the upper bound of $h_r$ to be of the order $\frac{n_{\calK}}{\sqrt{n_0}}$, ignoring the logarithm term. 

\medskip
\noindent
\textbf{Transfer Learning.}  
Transfer learning has been studied under both conditional and marginal shifts between source and target domains. Conditional shift includes {\em posterior drift} \citep{li2022transfer-jrssb}, where the conditional distribution of $Y$ given $X$ differs across tasks. Marginal shifts include {\em covariate shift} \citep{wang2023pseudo}, where the distribution of $X$ changes, and {\em label shift} \citep{maity2022minimax}, where the distribution of $Y$ differs.
The present paper is most related to the literature of transfer learning under {\em posterior drifts}, which has been studied in different contexts across a spectrum of supervised learning (SL) problems, including classification \citep{cai2021transfer}, high-dimensional linear regression \citep{li2022transfer-jrssb}, %{gu2022robust}, 
and generalized linear models \citep{tian2023transfer, li2023estimation}. %, and there are also applications to problems in unsupervised learning (UL) scenarios~\citep{li2022transfer-jasa}. 
%Moreover, this work is also related to other multi-dataset problems such as multi-task learning  (see, e.g., \citealp{duan2023adaptive}  and references therein) and multi-distribution learning (see, e.g., \citealp{zhang2023optimal}).

This paper uniquely explores TL within offline RL contexts, particularly focusing on {\em posterior drifts}. Our objective is to estimate the optimal $Q^*$ function through sample transfers directly. Unlike supervised learning, offline RL estimation of $Q^*$ involves no direct observation but instead seeks to approximate the fixed point of the population-level Bellman optimality equation through iterative updates of a sample-level version. This process requires a sophisticated de-biasing method in each iteration to counter sequential bias from task variances. Additionally, our theoretical exploration into TL using semi-parametric sieve approximation presents novel insights into RL-based transfer learning, uncovering phenomena not previously identified in supervised or unsupervised learning contexts. A thorough discussion is provided in Section B of the supplementary materials. 

\medskip
\noindent
\textbf{Transfer Learning for RL.} The RL framework, inclusive of various elements within an MDP, leads to empirical TL studies in deep RL adopting different assumptions about task similarities across MDP components, resulting in diverse research focuses. For instance, learning from demonstration assumes identical source and target MDPs \citep{ma2019imitation}. %\citep{piot2014boosted,ma2019imitation}. 
In contrast, policy transfer research often considers variations in state and action spaces \citep{yin2017knowledge} or reward functions \citep{barreto2017successor}, while reward shaping studies presuppose differences in reward functions defined by a specific function \citep{vecerik2017leveraging}.  Representation transfer research posits that state, action, or reward spaces can be divided into orthogonal, task-invariant subspaces, facilitating knowledge transfer across domains \citep{chai2025transition,zhang2025transfer}. For recent reviews on TL in deep RL, see \cite{zhu2023transfer} and references therein.

Theoretical RL research has begun to rigorously address TL with formal proofs, concentrating on non-stationary finite-horizon MDPs. Under linear MDP settings with varying reward functions, \cite{chen2022transferred} and \cite{chai2025deep} introduced a transfer algorithm leveraging backward-style dynamic programming and one-step least-square regression, contrasting with our iterative approach. \cite{agarwal2023provable} and \cite{chai2025transition} explored representation transfer, assuming a low-rank or low-rank plus sparse transition model, whereas we avoid specific assumptions about transition probabilities. 
\cite{zhou2025prior} studied prior-aligned meta-RL under the Baysian framework. 
Moreover, \cite{qu2024hybrid} formulated a hybrid transfer RL problem, where the agent transfers knowledge on offline source tasks to learn in an online target environment. Our theoretical contributions, highlighting the impact of task discrepancies on TL's statistical benefits in RL, enrich this body of work and enhance the understanding of TL in sequential decision-making.

\medskip
\noindent
\textbf{Organization.} 
The remainder of the paper is organized as follows. Section \ref{sec:model} formulates transfer learning in stationary sequential decision settings and defines task discrepancy over MDPs. Section \ref{sec:estimation} introduces our transferred $Q^*$ learning algorithm under general and sieve-based function approximation. Section \ref{sec:theory} provides theoretical guarantees under both homogeneous and heterogeneous transitions, along with a data-driven method for selecting model complexity. Sections \ref{sec:simul} and \ref{sec:appl} present empirical results on synthetic and real data. Additional computational details and proofs are provided in the supplementary materials.

\section{Statistical Framework} \label{sec:model}

\paragraph{Mathematical Framework for RL} %under a Stationary Environment.}
%We focus on data-driven decision-making, or reinforcement learning, under a stationary environment. 
The mathematical model for studying RL is the discounted \textit{Markov Decision Process} (MDP), characterized by a tuple $\calM = \braces{\calX, \calA, P, r, \gamma, \nu}$. 
We specifically focus on the setting with finite action space $\calA$, i.e., $\calA=\{1,2,\dots, m\}$ for a constant $m$. 
For a fixed trajectory index $i = 1, 2, \dots, I$, at time $t=0, 1, \dots, T$, an agent observes the current system state $\bX_{i, t}$ supported on the state space $\calX$, chooses a decision $A_{i, t}$ supported on the action space $\calA$, transits to the next state $\bX_{i, t+1}$ according to the system transition probability $P\paren{\cdot\,|\,\bX_{i, t}, A_{i, t}}$, and receives an immediate reward $R_{i,t}=r(\bX_{i, t}, A_{i, t})+\eta_{i,t}$, where $r(\bx, a)$ is a  reward function, and $\eta_{i,t}$ denotes a zero-mean noise. The distribution of the initial state $\bX_{i, 0}$ is denoted by $\nu$.

An agent's decision-making rule is characterized by a policy $\pi\paren{a \mid \bx}$ that defines a distribution over actions conditional on states.
%\subsection{Policy and value functions}
Formally, a policy $\pi\paren{a\mid\bx}: \calX \mapsto \calP(\calA)$ is a function that maps the state space $\calX$ to probability mass functions on the action space $\calA$.
It satisfies $\pi(a\mid\bx)\ge 0$ for any $a\in\calA$, $\bx\in\calX$, and $\underset{a\in\calA}{\sum} \pi(a\mid\bx) = 1$ for any $\bx\in\calX$. Under policy $\pi$, at time $t$, a decision maker chooses action $A_{i,t} = j$ at state $\bX_{i,t} = \bx$ with probability $\pi(j \mid \bx)$. %and receives an immediate reward $R_{i,t} = r\paren{\bX_{i,t}, A_{i,t}} + \eta_{i,t}$.
%The trajectory $i$ is then given by $\calH_i= \braces{(\bX_{i,0}, A_{i,0}), \cdots, (\bX_{i,T}, A_{i,T})}$, where the horizon $T$ can  be either finite or infinite. Let $\nu$ denote the distribution of the initial state $\bX_{i, 0}$.
%The trajectory distribution $\calP_\pi$ given policy $\pi$ is then given by
%\begin{equation}  \label{eqn:probtrajectory}
%\calP_\pi(\calH_i) = \nu(\bX_{i,0}) \prod_{t=0}^{\infty} \pi\paren{A_{i,t} | \bX_{i,t}} P\paren{\bX_{i,t+1} | \bX_{i,t}, A_{i,t}}.
%\end{equation}
The goal of RL is to learn an optimal policy that maximizes the expected discounted accumulative reward, or {\em expected return}, defined as
\begin{equation} \label{eqn:expected-return}
v^{\pi}: = \EE_{\pi}\left[\sum_{t=0}^{\infty} \gamma^t R_{i, t}\right],
\end{equation}
where the expectation is taken under the trajectory distribution generated by policy $\pi$ on MDP $\calM$.
%The horizon $T$ can be either finite (episodic) or infinite (continuing), and 
The discount factor $\gamma\in[0,1)$ reflects a trade-off between immediate and future rewards.
%The total return collected over time on a trajectory $\calH_i$ is the accumulated discounted rewards defined as
%\begin{equation}  \label{eqn:return}
%    R^{\pi}_T\paren{\calH_i} = \sum_{t=0}^T \gamma^t R_{i,t},
%\end{equation}
%where the discount rate $\gamma$ reflects a trade-off between immediate and future rewards.
If $\gamma=0$, the decision maker chooses actions that maximize the immediate reward $R_{i, 0}$.
As $\gamma$ increases, the decision maker puts more weight on future rewards.
%The goal of reinforcement learning is to learn the optimal policy $\pi^{opt}(a \mid \bx)$ that maximizes the total return $R^{\pi}\paren{\calH_i}$.
%The value $T\cdot(1-\gamma)^{-1}$ can be viewed as the effective horizon length of an MDP. 

Given a policy $\pi$ and a discount factor $\gamma \in [0,1)$, the {\em state-value function} is the expectation of the total return starting from a state $\bx$:
\begin{equation} \label{eqn:v-func-0}
V^{\pi}(\bx) = \EE_{\pi}\brackets{ \sum_{t=0}^{\infty} \gamma^t R_{i,t} \bigg| \bX_{i,0} = \bx}.
\end{equation}
The {\em action-value function} or {\em $Q$-function} of a given policy $\pi$ is defined as the expectation of the accumulated discounted rewards starting from a state $\bx$ with action $a$:
\begin{equation} \label{eqn:q-func-0}
Q^{\pi}(\bx, a) = \EE_{\pi}\brackets{ \sum_{t= 0}^{\infty} \gamma^t R_{i,t} \bigg| \bX_{i,0} = \bx, A_{i,0} = a},
\end{equation}
where the expectation is taken by assuming that the dynamic system follows the given policy $\pi$ after the initial state. The {\em optimal action-value function $Q^*$} is defined as
\begin{equation} \label{eqn:opt-q}
Q^*\paren{\bx, a} = \underset{\pi}{\sup}\; Q^\pi\paren{\bx, a}, \quad \forall \paren{\bx, a} \in \calX \times \calA.
\end{equation}
where the supremum is taken over all policies. Moreover, for any given action-value function $Q:\calX \times \calA \mapsto \RR$, the {\em greedy policy} $\pi^Q$ is defined as the policy that selects the action with the largest Q-value, i.e., 
%\begin{equation} \label{eqn:greedy-policy}
$\pi^Q\paren{a \mid \bx} =0$ if $a \notin \underset{a^{\prime}}{\arg\!\max} \; Q\paren{\bx, a'}$.
%\end{equation}
It is well known that the optimal policy $\pi^*$ that maximizes the expected return in \eqref{eqn:expected-return} is the greedy policy of $Q^*$, i.e., 
\[\pi^*=\pi^{Q^*}.\]
Furthermore, one important property of $Q^*$ is the {\em Bellman optimal equation}:
\begin{equation} \label{eqn:bellman-opt}
\EE\Big[R_{i, t} + \gamma\; \underset{a'\in\calA}{\max}\; Q^*\paren{\bX_{i, t+1}, a'} - Q^*\paren{\bX_{i, t}, A_{i, t}} \big| \bX_{i, t}, A_{i, t} \Big] = 0.
\end{equation}

The goal of this paper is to  improve the learning of the $Q^*$ function by knowledge transfer. Once a better estimator of the $Q^*$ function is constructed with the assistance from source data, a better estimator for the optimal policy $\pi^*$ can be derived as the greedy policy with respect to the estimated $Q^*$.

\paragraph{The Target and Source RL Data. }
Transfer RL aims to improve the learning on a target RL task by leveraging data from similar source RL tasks.
We consider the case where we have abundant source data from offline observational data or simulated data, while the target task only has a small amount of offline data. Specifically, 
we have a target task and $K$ source tasks, which are characterized by MDPs $\calM^{(k)} = \braces{\calX, \calA, P^{(k)}, r^{(k)}, \gamma, \nu^{(k)}}$ for $k\in\{0\}\cup[K]$. 
The target RL task of interest is referred to as the $0$-th task and denoted by a superscript ``${(0)}$,'' while the source RL tasks are denoted by a superscript ``${(k)}$,'' for $k\in[K]$. 

Without loss of generality, we assume the horizon length of all tasks is the same,  denoted as $T$.
For each task $k\in\{0\}\cup[K]$, we collect $I^{(k)}$ i.i.d. trajectories of length $T$, denoted as $\big\{\big(\bX_{i, t}^{(k)},A_{i, t}^{(k)},R_{i, t}^{(k)}, \bX_{i, t+1}^{(k)}\big)\big\}$, $1\leq i \leq I^{(k)}$, $0\leq t \leq T-1$.
We also assume that the trajectories in different tasks are independent.

Single-task RL considers each task $k\in \{0\} \cup [K]$ separately and defines the underlying {\em true response} of interest at step $t$ as
\vspace{-1ex}
\begin{equation}  \label{eqn-yk}
Y^{(k)}_{i, t}
:= R^{(k)}_{i, t}
+ \gamma \cdot \max_{a'\in\calA} Q^{*(k)}\big(\bX_{i, t+1}^{(k)}, a'\big),
\end{equation}
where $Q^{*(k)}$ denotes the optimal action-value function of task $k$. 
According to the Bellman optimal equation \eqref{eqn:bellman-opt}, we have
\begin{equation}
\label{bellman0}
Q^{*(k)}\paren{\bx, a} = \EE\brackets{ Y^{(k)}_{i, t}
	\,\big|\, \bX^{(k)}_{i, t}=\bx,\; A_{i, t}^{(k)}=a },
\quad\text{for}\quad k\in \{0\} \cup [K],
\end{equation}
which provides a moment condition for the estimation of $Q^{*(k)}\paren{\bx, a}$.
If $Y^{(k)}_{i, t}$ is directly observable, then $Q^{*(k)}\paren{\bx, a}$ can be estimated via regression.
However, what we observe in the RL setting is only a ``partial response'' $R^{(k)}_{i, t}$.
The other component of $Y^{(k)}_{i, t}$, as shown in the second term on the RHS of \eqref{eqn-yk}, depends on the unknown $Q^*$ function and future observations. As will be discussed in detail in Section \ref{sec:estimation}, we estimate $Q^{*(k)}\paren{\bx, a}$ in an iterative fashion.

\paragraph{Similarity Measure for Transferring between Different MDPs.}
The study of transfer learning necessitates a formal characterization of task similarity or difference. 
Since MDPs are characterized by tuples $\calM^{(k)} = \braces{\calX, \calA, P^{(k)}, r^{(k)}, \gamma}$ for $k\in\{0\}\cup[K]$, we characterize the similarity between the target and the source tasks through the difference between the reward functions $r^{(k)}$ and the transition probabilities $P^{(k)}$. Specifically, we denote $\rho^{(k)}$ as the density of the transition kernel $P^{(k)}$ and assume it exists almost everywhere for all $k$. The discrepancy between any source task $k$ and the target task $0$ is quantified by  
\begin{align}
\delta^{(k)}_r(\bx, a) &:= r^{(k)}(\bx, a) - r^{(0)}(\bx, a), \label{eqn:delta-r} \\
\delta^{(k)}_\rho(\bx'\,|\,\bx, a) &:= \rho^{(k)}(\bx'\,|\, \bx, a) - \rho^{(0)}(\bx'\,|\,\bx, a).   \label{eqn:delta-p} 
\end{align}
Since our estimation target is the $Q^*$ function, we next establish the relationship between the difference of the $Q^*$ function and task discrepancy defined on MDP tuples. 

\begin{lemma}[Difference of $Q^*$] \label{lem:diff-Q*}
Let the difference between the optimal action-value functions across different tasks be defined as
\begin{equation} \label{eqn:delta-q}
	\delta_Q^{(k)}(\bx, a) := Q^{*(k)}(\bx,a) - Q^{*(0)}(\bx,a).
\end{equation}
Assume that the reward functions $r^{(k)}(\bx, a)$ are uniformly upper bounded by a constant $R_{\max}$. Then we have
\begin{equation}\label{eq: delta-Q^*-r-P}
	\sup_{\bx, a}\abs{\delta_{Q}^{(k)}\paran{\bx, a} }\leq \frac{1}{1-\gamma}\sup_{\bx, a}\abs{\delta_r^{(k)}\paran{\bx, a}}+\frac{\gamma R_{\max}}{(1-\gamma)^2}\int_{\calX} \sup_{\bx, a}\abs{\delta_\rho^{(k)}\paran{\bx'\mid \bx, a }}\mathrm{d} \bx'.
\end{equation}
\end{lemma}
Lemma \ref{lem:diff-Q*} shows that the magnitude of the difference of $Q^*$ functions can be upper bounded by that of $\delta_r$ and $\delta_\rho$, which theoretically guarantees the transferability across RL tasks that are similar in reward functions and transition kernels for optimal $Q^*$ learning. 

Indeed, the premise of transfer learning is that the differences $\delta_r$ and $\delta_\rho$ are ``small'' in the sense that the bias incurred from different source tasks can be ``easily'' corrected even with a small amount of target data, which will be shown later in the theoretical analysis in Section \ref{sec:theory}. For a better illustration, the magnitude of $\delta_r$ and $\delta_\rho$ will be instantiated under a specific function class and will show up in the statistical learning rate of our proposed algorithm, where we provide a formal quantification for how ``small" the differences should be to benefit from transferring. As the rewards and transition pairs for all of the stages are {\em directly observable}, it can be {\em verified} in practice whether similarity assumptions to be imposed on \eqref{eqn:delta-r} and \eqref{eqn:delta-p} are satisfied \citep{silver2021reward}.

%%
%%  Fitted Q learning + Transfer
%%
\section{Batch $Q^*$ Learning with Knowledge Transfer}
\label{sec:estimation}

\subsection{Transfer FQI with General Function Approximation}

The proposed knowledge transfer algorithm is based on the framework of {\em Fitted $Q$-Iteration} (FQI) due to its wide applications in offline RL. 
The framework of FQI aims to minimize the \textit{Bellman error} by bootstrapping and semi-gradient method. Inspired by the Bellman optimal equation
\begin{equation*} 
\EE\brackets{ R_{i,t} + \gamma \underset{a'\in\calA}{\max}\; Q^*\paran{\bX_{i,t+1}, a'} \big| \bX_{i,t}, A_{i,t} } = Q^*\paran{\bX_{i,t}, A_{i,t}},
\end{equation*}
FQI proceeds iteratively with a function class $\hat{Q}\big(\bx, a;\bbeta\big)$ parameterized by $\bbeta$ to approximate $Q^*\paran{\bx, a}$. 
In the $\tau$-th iteration, given an estimator $\hat{Q}\big(\bx, a;\hat\bbeta_{\tau-1}\big)$, the FQI computes $Y^{\tau}_{i,t}=R_{i,t} + \gamma\underset{a'\in\calA}{\max}\;\hat{Q}\big(\bX_{i,t+1}, a';\hat\bbeta_{\tau-1}\big) $ as a pseudo-response variable and regresses $\{Y^{\tau}_{i,t}\}$ on $\{(\bX_{i,t}, A_{i, t})\}$ to obtain an updated estimator $\hat{Q}\big(\bx, a;\hat\bbeta_{\tau}\big)$. Specifically,  
\begin{equation} \label{eqn:fqi-update}
\hat\bbeta_{\tau} = \underset{\bbeta}{\arg\!\min}\; \sum_{i=1}^{I}\sum_{t=0}^{T-1} \big( Y^{\tau}_{i,t} - \hat{Q}\paran{\bX_{i,t}, A_{i,t}; \bbeta} \big)^2.
\end{equation}
Based on the iterative framework of FQI, we develop the Transfer FQI algorithm (Algorithm \ref{algo:fitted-q-transfer}) to apply knowledge transfer across different batch RL tasks.
Suppose that we obtain samples $\calS^{(k)}=\big\{\big(\bX^{(k)}_{i,t}, A^{(k)}_{i,t}, R^{(k)}_{i,t}, \bX^{(k)}_{i,t+1}\big)\big\}$ for $1 \le i \le I^{(k)}, 0 \le t \le T-1$ that are randomly sampled from the $k$-th task, where $k=0$ represents the target task and $k\in[K]$ represents the $k$-th source task. As illustrated in Section \ref{sec:model}, the samples are assumed to be i.i.d. across trajectory $i$ but correlated across $t$ for the same $i$. For each $k$, we evenly divide $\calS^{(k)}$ into $\Upsilon$ disjoint subsets $\calS_1^{(k)},\calS_2^{(k)}, \dots, \calS_\Upsilon^{(k)}$, where the subset $\calS_{\tau}^{(k)}$ contains samples with indices $1+(\tau-1)I^{(k)}/\Upsilon \leq i \leq \tau I^{(k)}/\Upsilon, 0\leq t\leq T-1$. We use $n_k=I^{(k)}T/\Upsilon$ to denote the sample size of subset $\cS_{\tau}^{(k)}$. The subsets $\{\calS_\tau^{(k)}\}_{k=0}^K$ contain the data we used in the $\tau$-th iteration of our proposed algorithm, for $\tau=1,2,\dots, \Upsilon$. This sample splitting procedure ensures the independence between the samples used in different iterations.

On the population level, there exists a  center of the $Q^*$ functions $\{Q^{*(k)}\}_{k=0}^K$, defined as
\begin{equation*} 
W^* = \underset{W}{\arg\!\min}\;\EE\bigg[
\sum_{i, t, k} \Big(Y_{i,t}^{(k)} - W\big(\bX_{i,t}^{(k)}, A_{i,t}^{(k)}\big)\Big)^2 \bigg],% \cond \bX_{i,t}^{(k)}=\bx, A_{i,t}^{(k)}=a \bigg],
\end{equation*}
where $Y_{i, t}^{(k)}$ is the true response defined by \eqref{eqn-yk}.
It is straightforward to derive that 
\begin{equation}\label{eqn:W*}
W^*=(n_{\cK})^{-1}\sum_{k=0}^K n_k\cdot Q^{*(k)} = Q^{*(0)} + (n_{\cK})^{-1}\sum_{k=0}^K n_k\cdot\delta_Q^{(k)}, 
\end{equation}
where $n_{\calK}:=\sum_{k=0}^{K}n_k$, and $\delta_Q^{(k)}$ is the defined in \eqref{eqn:delta-q}. 
The weighted average $\delta^{(0)}:= (n_{\cK})^{-1}\sum_k n_k\cdot\delta_Q^{(k)}$ characterizes the bias of the center $W^*$ from the target function $Q^{*(0)}$.
If each $\delta_Q^{(k)}$ is ``sufficiently small,'' then their weighted average is also small,  and we expect to learn the bias $\delta^{(0)}$ well even with a small amount of target data. 

To estimate the optimal action-value function $Q^*$, we further need an \textit{approximating space} $\calQ$, a well-defined function class on $\calX \times \calA$. Given a general approximating space $\cQ$, we denote the projection of the optimal action-value function $Q^{*(k)}$ and the center $W^*$ on $\cQ$ by $\hat{Q^{*(k)}}$ and $\hat{W^*}$, respectively. 
The above equations also hold for the projections $\hat{Q^{*(k)}}$ and $\hat{W^*}$, so we estimate $\hat{W^*}$ by minimizing the empirical $L_2$ loss in Step \rom{1} (Equation \eqref{eqn:pool-gen}) of Algorithm \ref{algo:fitted-q-transfer}. 
Since an informative source task must be similar to the target task, the approximating space for the bias $\delta^{(0)}$, denoted as $\cQ' \subset \cQ$, is often more restrictive such that we can employ this restrictive structure to estimate $\delta^{(0)}$ well even with a small amount of target data. 
In literature, restrictive structures that characterize task similarity include the sparse condition \citep{li2022transfer-jrssb}, smoothness condition \citep{cai2021transfer}, polynomial order \citep{cai2024transfer}, and RKHS norms \citep{wang2023minimax}. In Step \rom{2} of Algorithm \ref{algo:fitted-q-transfer}, we denote the restrictive structure imposed by $\cQ'$ as a norm $\norm{\cdot}_{\cQ'}$ and minimize a $\norm{\cdot}_{\cQ'}$-penalized objective in \eqref{eqn:correct-gen} to obtain an estimator for the bias $\delta^{(k)}$ on each task.
At the end of each iteration, the optimal action-value functions $Q^{*(k)}$ are estimated by combining the center estimator \eqref{eqn:pool-gen} with the bias-correction estimator \eqref{eqn:correct-gen} on each task, where, in particular, $\widehat{Q}^{(0)}$ is our goal estimator for the target task. These estimators are further refined as the iterations proceed.

\begin{algorithm}[!ht]
\caption{{\sc TransFQI}: Transfer Fitted $Q$-Iteration (General)} \label{algo:fitted-q-transfer}
\DontPrintSemicolon % Some LaTeX compilers require you to use \dontprintsemicolon instead

\KwIn{
	%Target data $\calS^{(0)}=\big\{\big(\bX_{i,t}^{(0)}, A_{i,t}^{(0)}, R_{i,t}^{(0)}, \bX_{i, t+1}^{(0)}\big)\big\}$, $1 \leq i \leq I^{(0)}$, $0\leq t \leq T-1$;  
	%Informative source data $\calS^{(k)}=\big\{\big(\bX_{i,t}^{(k)}, A_{i,t}^{(k)}, R_{i,t}^{(k)}, \bX_{i,t+1}^{(k)}\big)\big\}$, $1 \leq i \leq I^{(k)}$, $0\leq t \leq T-1$, $1 \leq k \leq K$; 
	%An approximation space $\cQ$ for $Q^*$ and a function class $\cQ'$ for the difference $\delta$; 
	% Initial estimators $\hat{Q}_0^{(k)}\in\cQ$;
	%Maximum number of iterations $\Upsilon$; 
	%Regularization parameters $\lambda_{\delta}^{(k)}$ for $0 \leq k\leq K$.
	\begin{itemize}
		\item Target data $\calS^{(0)}=\big\{\big(\bX_{i,t}^{(0)}, A_{i,t}^{(0)}, R_{i,t}^{(0)}, \bX_{i, t+1}^{(0)}\big)\big\}$, $1 \leq i \leq I^{(0)}$, $0\leq t \leq T-1$;  
		\item Informative source data $\calS^{(k)}=\big\{\big(\bX_{i,t}^{(k)}, A_{i,t}^{(k)}, R_{i,t}^{(k)}, \bX_{i,t+1}^{(k)}\big)\big\}$, $1 \leq i \leq I^{(k)}$, $0\leq t \leq T-1$, $1 \leq k \leq K$; 
		\item An approximation space $\cQ$ for $Q^*$ and a function class $\cQ'$ for the difference $\delta$; \item Initial estimators $\hat{Q}_0^{(k)}\in\cQ$;
		\item Maximum number of iterations $\Upsilon$; 
		\item Regularization parameters $\lambda_{\delta}^{(k)}$ for $0 \leq k\leq K$.
	\end{itemize}
}

\vspace{1ex}

\KwOut{An estimator $\hat{Q}^{(0)}_\Upsilon$ and the corresponding greedy policy $\hat\pi^{(0)}_\Upsilon=\pi^{\hat{Q}^{(0)}_\Upsilon}$.}

\vspace{1ex}

For $k = 0,1, \dots, K$, evenly divide $\cS^{(k)}$ into $\Upsilon$ disjoint subsets $\calS_1^{(k)},\calS_2^{(k)}, \dots, \calS_\Upsilon^{(k)}$;

\For{$\tau = 1, 2,  \dots, \Upsilon$}{
	
	Compute $Y_{i,t}^{(k), \tau} = R_{i,t}^{(k)} + \gamma\cdot\max\limits_{a'\in\calA}\hat{Q}_{\tau-1}^{(k)}\big(\bX_{i, t+1}^{(k)}, a'\big)$ for all $(i, t) \in \calS_{\tau}^{(k)}$ and $0 \leq k \leq K$. %Compute the sample size $n_k=\sum_{i=1}^{I^{(k)}} T_i^{(k)}$ and $n_{\calK}=\sum_{k}n^{(k)}$.
	
	\vspace{1ex}
	
	\underline{\sc Step I.}  Compute an aggregated estimator for all tasks: 
	\vspace{1ex}
	\begin{equation}\label{eqn:pool-gen}
		\hat{W}_{\tau} =
		\underset{W\in\cQ}{\arg\!\min}\, \Big(
		\frac{1}{2n_{\calK}}
		\sum_{k=0}^K\sum_{(i, t) \in \calS_\tau^{(k)}} \brackets{Y_{i,t}^{(k), \tau} - W\big(\bX_{i,t}^{(k)}, A_{i,t}^{(k)}\big)}_2^2 \Big).
	\end{equation}
	
	\vspace{1ex}
	
	\underline{\sc Step II.}  Compute a corrected target estimator for each task:
	
	\vspace{1ex}
	
	\For{$k = 0, 1, \dots, K$}{
		\vspace{1ex}
		Obtain
		\begin{equation}
			\hat{Q}_{\tau}^{(k)}= \hat{W}_{\tau} + \hat\delta_{\tau}^{(k)},
		\end{equation}
		where
		\begin{equation} \label{eqn:correct-gen}
			\hat\delta_{\tau}^{(k)} =
			\underset{\delta\in\calQ'}{\arg\!\min}\, \Big(
			\frac{1}{2 n_k} \sum_{(i, t) \in \calS_\tau^{(k)}}
			\brackets{Y_{i,t}^{(k), \tau}-\hat{W}_{\tau}(\bX_{i, t}^{(k)}, A_{i, t}^{(k)})- \delta(\bX_{i, t}^{(k)}, A_{i, t}^{(k)})}_2^2 + \lam^{(k)}_{\delta}\norm{\delta}_{\cQ'} \Big),
		\end{equation}
		and $\norm{\cdot}_{\cQ'}$ is a function norm that imposes structures on the task difference. 
	}
	\vspace{1ex}
}
\end{algorithm}

\subsection{Transfer FQI with Sieve Function Approximation}

Algorithm \ref{algo:fitted-q-transfer} establishes a general framework for transferred FQI. 
With different applications, the approximating space $\calQ$ can be chosen according to the norm of the domain. 
For example, for language and vision tasks, neural networks are usually chosen for the approximating space $\cQ$ due to the intrinsic data structure and the availability of massive training data. 
However, semi-parametric sieve approximation has been proven to offer better approximation and sensible interpretations for applications of business, economics, and finance \citep{chen2007large}. 
To propel data-driven decision in societal applications, we focus our attention on sieve function approximation hereafter. 

Now we instantiate Algorithm \ref{algo:fitted-q-transfer} with sieve function approximation.
The approximating space $\cQ$ is chosen to be linear combinations of sieve basis functions, i.e., $\cQ$ contains functions of the form $\hat Q(\bx, a; \bbeta)=\bxi^\top(\bx, a)\bbeta$, where 
\begin{equation}\label{eq:def-xi}
\bxi(\bx, a):=\left[\bphi^{\top}(\bx)\II(a=1),\bphi^{\top}(\bx)\II(a=2),\dots,\bphi^{\top}(\bx)\II(a=m)\right]^{\top},
\end{equation}
and $\bphi(\cdot)=\paran{\phi_1(\cdot),\dots, \phi_p(\cdot)}^{\top}$ denotes a set of pre-selected sieve basis functions such as B-splines or wavelets. %{\blue For the convenience of constructing sieve basis functions, we hereafter assume that the state space $\calX$ is compact.}

With sieve approximation, Algorithm \ref{algo:fitted-q-transfer} is instantiated in several aspects. 
Firstly, the initial estimator is characterized by   $\hat{Q}_0^{(k)}\paran{\bx, a} = \bxi^\top\paran{\bx, a}\hat\bbeta_0^{(k)}$, where $\hat\bbeta_0^{(k)}$ is an $(mp)$-dimensional initial  coefficient vector for the $k$-th task. In the $\tau$-th iteration, after computing the pseudo response variable $\{Y_{i, t}^{(k), \tau}\}$ for all of the tasks, the aggregated estimator %$\hat{W}_{\tau}\paran{\bx, a}$ is instantiated as
$\hat{W}_{\tau}\paran{\bx, a} = \bxi^\top\paran{\bx, a} \hat{\bw}_{\tau}$, where $\hat{\bw}_{\tau}$ is obtained from sieve-instantiated Equation \eqref{eqn:pool-gen} in Step \rom{1}, that is,
\begin{equation}
\hat{\bw}_{\tau} =
\underset{\bw\in\RR^{mp}}{\arg\!\min}\, 
\frac{1}{2n_{\calK}}
\sum_{k=0}^K\sum_{(i, t) \in \calS_\tau^{(k)}} \brackets{Y_{i,t}^{(k), \tau} - \bxi^\top\big(\bX_{i,t}^{(k)}, A_{i,t}^{(k)}\big) \bw}_2^2 .
\end{equation}
Due to the linearity of sieve spaces, the aggregated estimator $\hat{\bw}_{\tau}$ can be viewed as an estimator for a weighted average of the underlying parameters of the $K$ tasks, which is biased from the parameter of each task. In Step \rom{2}, we estimate the bias of $\hat{\bw}_{\tau}$ on each task by an $\ell_1$-regularized estimator $\hat{\bdelta}_{\tau}^{(k)}$, as we use $\ell_1$ distance to measure the difference across tasks under sieve function approximation. That is, 
\begin{equation}
\hat\bdelta_{\tau}^{(k)} =
\underset{\bdelta\in\RR^{mp}}{\arg\!\min}\, \bigg(
\frac{1}{2 n_k}\sum_{(i, t) \in \calS_\tau^{(k)}}
\brackets{Y_{i,t}^{(k), \tau}- \bxi^\top\big(\bX_{i,t}^{(k)}, A_{i,t}^{(k)}\big) \paran{\hat\bw_{\tau} + \bdelta }}_2^2 + \lam^{(k)}_{\delta}\norm{\bdelta}_1 \bigg).
\end{equation}
Then we correct the aggregated estimator by $\hbbeta_{\tau}^{(k)}=\hat{\bw}_{\tau}+\hat{\bdelta}_{\tau}^{(k)}$, and the corrected estimator $\hbbeta_{\tau}^{(k)}$ is input into the next iteration for further refinement. 

\section{Theory}
\label{sec:theory}

In this section, we establish statistical guarantees for our proposed transferred FQI algorithm (Algorithm \ref{algo:fitted-q-transfer}) under sieve function approximation.  As defined in Section \ref{sec:model}, the dataset of task  $k\in\{0\}\cup[K]$ contains $I^{(k)}$ independent trajectories, denoted by $\big\{\big(\bX_{i, t}^{(k)}, A_{i, t}^{(k)}, R_{i, t}^{(k)}, \bX_{i, t+1}^{(k)}\big)\big\}$ ($1 \leq i \leq I^{(k)}$, $0 \leq t \leq T-1$), each with length $T$.  We first introduce some common regularity conditions for the theoretical development. 

\begin{assumption}
\label{assum:distribution}
For all tasks $k\in \{0\}\cup[K]$, assume the following conditions hold:
\begin{enumerate}
	%\item[(a)] There exists a behavior policy $b^{(k)}(\cdot \,|\, \bx)$ such that $\PP\left(A^{(k)}_{i, t}=a \mid \bX^{(k)}_{i, t}=\bx\right) = b^{(k)}(a \mid \bx)$ for all $i \in [I^{(k)}]$, $t\in [T]$. Moreover, assume that the behavior distribution is bounded away from 0, i.e., $\inf\limits_{a, \bx} b^{(k)}(a \,|\, \bx) \geq \underline{b}$ for a constant $\underline{b} > 0$. 
	
	\item[(a)] Given $\bX^{(k)}_{i,t}=\bx$ and $A^{(k)}_{i,t}=a$, assume that the distribution of the next state $\bX_{i,t+1}^{(k)}$ is determined by a time-invariant transition kernel $P^{(k)}\big(\cdot \,|\, \bx, a\big)$  with density $\rho^{(k)}(\cdot \,|\, \bx, a)$ almost everywhere. Moreover, assume that there exists a behavior policy $b^{(k)}(\cdot \,|\, \bx)$ such that $\PP\big(A^{(k)}_{i, t}=a \,|\, \bX^{(k)}_{i, t}=\bx\big) = b^{(k)}(a \,|\, \bx)$ for all $i, t$.
	
	\item[(b)] Assume that the Markov chain $\big\{\bX^{(k)}_{i,t}\big\}_{t =0}^{T-1}$ has a unique stationary distribution with a density $\mu^{(k)}$ almost everywhere. Let $\nu^{(k)}$ denote the probability density of the initial state $\bX_{i,0}^{(k)}$. Assume that $\mu^{(k)}$ and $\nu^{(k)}$ are  bounded away from 0 and $\infty$. Furthermore, assume that the Markov chain $\big\{\bX^{(k)}_{i,t}\big\}_{t =0}^{T-1}$ is geometrically ergodic as $T\to\infty$.
	
	\item[(c)]
	Let $\bSigma^{(k)}:=\frac{1}{T}\EE\Big[\sum_{t=0}^{T-1}\bxi\big(\bX_{i, t}^{(k)}, A_{i, t}^{(k)}\big)\bxi^{\top}\big(\bX_{i, t}^{(k)}, A_{i, t}^{(k)}\big)\Big]$, where $\bxi(\bx, a)$ is the basis function defined in \eqref{eq:def-xi}.  Assume that there exists a constant $c_{\Sigma} \geq 1$ such that $c_{\Sigma}^{-1} \leq \lambda_{\min}(\bSigma^{(k)})\leq\lambda_{\max}(\bSigma^{(k)}) \leq c_{\Sigma}$ for all $k$.
	
	\item[(d)] 	Assume that the reward $R_{i, t}^{(k)}=r^{(k)}\big(\bX_{i, t}^{(k)}, A_{i, t}^{(k)}\big) + \eta^{(k)}_{i,t}$, where the noise $\eta^{(k)}_{i,t}$ is $\sigma_{\eta}^2$-sub-Gaussian with a constant $\sigma_{\eta}>0$ that does not depend on $i,t,k$.
\end{enumerate}
\end{assumption}	

In Assumption \ref{assum:distribution}, condition (a) ensures that each task $k$ has time-invariant MDP dynamics, with both the transition kernel $P^{(k)}$ and the behavior policy $b^{(k)}$ of the offline dataset independent of time $t$. This time-homogeneity implies that for any trajectory $i$, the sequence $\big\{\big(\bX^{(k)}_{i,t}, A^{(k)}_{i,t} \big)\big\}$ forms a time-invariant Markov chain on $\calX \times \calA$ with transition kernel $P^{(k)}(\bx^{\prime}\,|\, \bx, a)b^{(k)}(a^{\prime}\,|\,\bx^{\prime})$. Consequently, the state sequence $\big\{\bX^{(k)}_{i,t}\big\}$ also forms a time-invariant Markov chain with transition kernel $\sum_{a \in \calA} b^{(k)}(a \,|\,\bx)P^{(k)}(\bx^{\prime}\,|\, \bx, a)$.

Condition (b) further requires that $\big\{\bX^{(k)}_{i,t}\big\}$ has a stationary distribution with bounded density and exhibits geometric ergodicity, i.e., the Markov chain approaches stationarity at a geometric rate as $T\to \infty$. This property guarantees that the Markov chain has a stable long-term behavior regardless of its initial state and, crucially, ensures that the sample covariance matrix $\widehat{\bSigma}^{(k)}:=\frac{1}{n_k}\sum_{i=1}^{I^{(k)}}\sum_{t=0}^{T-1}\bxi\big(\bX_{i, t}^{(k)}, A_{i, t}^{(k)}\big)\bxi^{\top}\big(\bX_{i, t}^{(k)}, A_{i, t}^{(k)}\big)$ converges to its population counterpart $\bSigma^{(k)}=\EE\Big[\widehat{\bSigma}^{(k)}\Big]$ at the rate $O_{\PP}\big(\sqrt{p\log n_k/n_k}\big)$. This convergence is essential for establishing the consistency of our regression-based estimation procedure.

Moreover, condition (c) assumes the invertibility and boundedness of the population covariance matrices $\bSigma^{(k)}$ uniformly across all tasks. This assumption, combined with condition (b), ensures the invertibility of the sample covariance matrices $\widehat{\bSigma}^{(k)}$ and enables us to establish the consistency of our regression estimators $\widehat{\bbeta}^{(k)}_{\tau}$ to their theoretical counterparts $\bbeta^{(k)}_{\tau}$, which are key intermediate coefficients defined in the technical proof. Finally, condition (d) assumes sub-Gaussianity of the reward noise, a standard assumption that facilitates the derivation of concentration bounds for $\widehat{\bbeta}^{(k)}_{\tau}$.

\begin{rmk}\label{rmk:data-coverage}
	Conditions (a), (b), and (c) are also related to the concentratability assumption \citep{chen2019information, xie2021batch, jia2024offline}. Specifically,  define an admissible distribution $\nu_{t}^{\pi}$ as the distribution of $(\bX_t, A_t)$ induced by initial distribution $\nu$ and policy $\pi$. The concentratability assumption in \cite{chen2019information} requires that $\nu_{t}^{\pi}(\bx, a) / \mu(\bx, a) \leq C$  for an absolute constant $C$ that is independent of $\nu$, $\pi$, and $t$, where $\mu$ is the distribution of i.i.d. data points. This assumption ensures that all admissible distributions can be effectively ``covered'' by the data distribution.  In contrast, our method does not assume i.i.d. sampling. Instead, we require that each Markov chain has a lower-bounded stationary distribution $\mu$ and an upper-bounded initial distribution $\nu$, and thus satisfies the concentratability assumption in the long run, enabling sufficient data coverage for learning the optimal policy.
\end{rmk}

We then introduce the notion of H\"{o}lder $\kappa$-smooth function, which is a generalization of Lipschitz continuity and is widely used to characterize the regularity of functions.

\begin{definition}[H\"{o}lder $\kappa$-smooth functions]
Let $f\paran{\cdot}$ be an arbitrary function on $\calX\in\RR^{d}$.
For a $d$-tuple $\balpha = (\alpha_1, \cdots, \alpha_d)$ of non-negative integers, let $D^{\alpha}$ denote the differential operator
$
D^{\balpha} f(\bx) = \frac{{\partial}^{\norm{\balpha}_1} f(\bx)}{\partial x_1^{\alpha_1}\cdots \partial x_d^{\alpha_d}}
$,
where $\bx=(x_1, \cdots, x_d)^\top$.
For $\kappa>0$, the class of $\kappa$-smooth functions is defined as
\begin{equation*}
	\Lambda \paran{\kappa, c} = \Big\{ f: \underset{\norm{\balpha}_1\le \ceil{\kappa}-1}{\sup} \; \underset{\bx\in\calX}{\sup} \; \abs{D^{\balpha} f(\bx)} \le c \text{ \,and } \underset{\norm{\balpha}_1= \ceil{\kappa}-1}{\sup}\; \underset{\bx_1 \ne \bx_2}{\sup} \; \frac{\abs{D^{\balpha} f(\bx_1) - D^{\balpha} f(\bx_2)}}{\norm{\bx_1-\bx_2}_2^{\kappa - \ceil{\kappa}+1}} \leq c\Big\},
\end{equation*}
where $\ceil{\kappa}$ denotes the the least integer greater than or equal to $\kappa$.
\end{definition}

With a set of typical sieve basis functions $\bphi(\bx)=(\phi_1,\dots,\phi_p)$ such as B-splines and wavelets, the H\"{o}lder $\kappa$-smooth functions satisfy the following property \citep{huang1998projection}: for any function $f(\cdot) \in \Lambda(\kappa, c)$,  there exist coefficients $\bbeta$ such that %$\braces{\bbeta_a\in\RR^p}_{a \in \calA}$ that 
\begin{equation}\label{eq:property}
\underset{\bx\in\calX}{\sup} \abs{  f\paran{\bx} - \bbeta^\top\bphi(\bx)} 
\le C p^{-\kappa/d},
\end{equation}
for some positive constant $C$. %Let $\bbeta=\paran{\bbeta_1^{\top},\dots,\bbeta_m^{\top}}^{\top}$, and then we have
%\[	\underset{\bx\in\calX, a\in \calA}{\sup} \abs{  f\paran{\bx, a} - \bbeta^\top\bxi(\bx, a)} 
%\le C p^{-\kappa/d}.\]
We then assume the following regularity condition on the reward functions and the transition kernels.

\begin{assumption} \label{assum:kappa-smooth}
Assume that there exist some constants $\kappa$, $c>0$ such that $r^{(k)}(\cdot, a)$ and $\rho^{(k)}(\bx' \,|\, \cdot, a)$ belong to $\Lambda(\kappa, c)$ for any $a\in\calA$, $ k \in \{0\} \cup [K]$, and $\bx'\in\calX$. In particular, there exists a uniform upper bound $R_{\max}$ on the reward functions, i.e., $\sup\limits_{\bx, a, k}\abs{r^{(k)}\paran{\bx, a}} \le R_{\max}$.
\end{assumption}

Assumption \ref{assum:kappa-smooth}, which requires $\kappa$-smoothness of both reward functions and transition densities for all tasks, is fundamental for ensuring the representation power of the approximation function space $\calQ$, i.e., the set of linear combinations of sieve basis functions. As highlighted in \cite{chen2019information}, theoretical guarantees for FQI with finite approximation function space rest on two key assumptions: realizability ($Q^* \in \calQ$) and completeness ($\calT f \in \calQ$ for all $f \in \calQ$, where $\calT$ denotes the Bellman optimality operator). Subsequent research has expanded to RL tasks with more general $Q^*$ functions that require approximation by infinite function classes, where H\"{o}lder smoothness provides the theoretical foundation for realizability and completeness (see, for example, \citealp{fan2020theoretical, shi2020statistical, shi2024statistically, bian2024off, wang2025adaptive}). Within this theoretical framework, we demonstrate through the following Lemma that Assumption \ref{assum:kappa-smooth} guarantees realizability approximately.

\begin{lemma}\label{lem:Q-kappa-smooth}
Under Assumption \ref{assum:kappa-smooth},  there exists some constant $c' > 0$ such that the optimal action-value function $Q^{*(k)}\paran{\cdot, a}$ belongs to the class $\Lambda(\kappa, c')$ for any task $k \in \{0\} \cup  [K]$ and any action $a\in\calA$. In particular,  $\sup\limits_{\bx, a, k}\abs{Q^{*(k)}(\bx, a)} \leq R_{\max}/(1-\gamma)$. 
\end{lemma}

%Lemma 1 in \cite{shi2020statistical} shows that under Assumption \ref{assum:kappa-smooth}, . Together with the \eqref{eq:property}, Assumption \ref{assum:kappa-smooth} ensures that the optimal $Q^*$ function on each task can be approximated by a linear combination of sieve basis functions. As a result, .    

Lemma \ref{lem:Q-kappa-smooth} indicates that, under Assumption \ref{assum:kappa-smooth}, the optimal action-value functions $Q^{*(k)}$ inherit $\kappa$-smoothness and can thus be approximated by $\mathcal Q$ with controlled error in \eqref{eq:property}. Moreover, approximate completeness follows directly from Assumption \ref{assum:kappa-smooth} and the definition that $\calT f(\cdot, a):=r(\cdot, a)+\int \max_{a' \in \calA} f(\bx', a') \rho(\bx' \,|\, \cdot, a) \mathrm{d} \bx'$. For additional discussions on the relationship between $\kappa$-smoothness and approxiamate completeness, we refer readers to Section 4 of \cite{fan2020theoretical}.

%In particular, they uniformly upper bounded in $\ell_{\infty}$ norm. 

\begin{rmk} \label{rmk:kappa-smooth}
The $\kappa$-smoothness condition is a generalization of Lipschitz continuity, as these two conditions are equivalent when $\kappa=1$. Our theoretical analysis requires no strong assumptions on the smoothness order $\kappa$, making the Assumption \ref{assum:kappa-smooth} mild—it is satisfied for $\kappa=1$ whenever the reward function and transition density are Lipschitz continuous. Furthermore, in Section \ref{sec:choose-p}, we develop a novel parameter selection method that achieves the desired theoretical guarantees without requiring any prior knowledge of $\kappa$.
\end{rmk}

In the remainder of this section, we first analyze a simplified setting in Section \ref{sec:res-homo} where the transition kernels $P^{(k)}$ are identical across tasks. %establishing theoretical guarantees for Algorithm \ref{algo:fitted-q-transfer}. 
Section \ref{sec:choose-p} introduces a data-adaptive approach for selecting the number of basis functions $p$ in practice. Section \ref{sec:res-hetero} then handles the general setting where transition kernels differ across tasks.

\subsection{Theoretical Results for Transition Homogeneous Tasks}
\label{sec:res-homo}

To clearly present our theoretical results, we first focus on the transition homogeneous setting, where the state-action variables are assumed to share the same distribution across different tasks, i.e., the transition $P^{(k)}(\bx'\,|\,\bx, a)$ are the same across $k\in\{0\}\cup[K]$. 
In contrast, the reward functions $r^{(k)}(\bx, a)$ are different across different tasks.  %, but the differences defined by \eqref{eqn:delta-r} between the target and the source tasks are small. 

Using the sieve approximation, we characterize the transferability across different tasks. Recall that $\bxi(\bx, a):=\left[\bphi^{\top}(\bx)\II(a=1),\bphi^{\top}(\bx)\II(a=2),\dots,\bphi^{\top}(\bx)\II(a=m)\right]^{\top}$. From \eqref{eq:property} and Assumption  \ref{assum:kappa-smooth}, we have that there exist coefficients $\big\{\bbeta^{(k)}_{r}\big\}$ %and $\left\{\bbeta^{(k)}_{\rho}(\bx')\right\}$ 
such that 
\[\sup_{\bx, a, k} \abs{r^{(k)}\paran{\bx, a} - \bxi^{\top}(\bx, a)\bbeta^{(k)}_{r}} \leq Cp^{-\kappa/d},\]
%and
%\[\sup_{\bx, a, \bx', k} \abs{\rho^{(k)}\paran{\bx' \,|\, \bx, a} - \bxi^{\top}(\bx, a)\bbeta^{(k)}_{\rho}(\bx')} \leq Cp^{-\kappa/d},\]
for some constant $C$. 
The similarity between $r^{(k)}\paran{\bx, a}$ and $r^{(0)}\paran{\bx, a}$ is manifested through their projections on the sieve space. Specifically, we use 
\begin{equation}\label{eq:h-homo}
h_r:=\max_{k\in[K]}\big\|\bbeta^{(k)}_{r}-\bbeta^{(0)}_{r}\big\|_1 
\end{equation}
to measure the discrepancy between $K$ source tasks and the target task. 
% Under $L_1$-norm in definition \eqref{eqn:H-r}, it holds that
% \begin{equation} \red 
% \max_{k\in[K]} H^{(k)}_{r,\nu,1} \le h_r, 
% \end{equation}
The level $h_r$ quantifies the difference between the target and source tasks. As long as $h_r$ is small enough, the source tasks are sufficiently informative to improve the estimation performance on the target task, which is shown by the following theorem for the theoretical property of Algorithm \ref{algo:fitted-q-transfer} under transition homogeneity:

\begin{theorem} \label{thm:Q-converge-homo}
Suppose Assumptions \ref{assum:distribution} and \ref{assum:kappa-smooth} hold. Further assume that  the sample sizes $n_0$ and $n_{\calK}$ satisfy $\frac{p\log^2 n_{\calK}}{n_{\calK}}+h_r\sqrt{\frac{\log p}{n_0}}=o(1)$.
By choosing the initial estimators such that $\sup\limits_{\bx, a,k} \widehat{Q}_0^{(k)}(\bx, a) \leq R_{\max}/(1-\gamma)$ and $\sup\limits_{k}\big\|\hbbeta_0^{(k)}-\hbbeta_0^{(0)}\big\|_1 \leq h_r$ and choosing the tuning parameter $\lam^{(k)}_{\delta}=
c_\delta \sqrt{\frac{\log p}{ n_k}}$ for some sufficiently large constant $c_\delta$, it holds that

\begin{equation}\label{eqn:rate-homo}
	%\begin{aligned} 
	v^{\pi^*}-v^{\widehat{\pi}_{\Upsilon}} = O_{\PP} \bigg(\frac{1}{(1-\gamma)^2} \Big[ \underbrace{p^{-\kappa/d}}_{\substack{\text{function-}\\\text{approximation} \\ \text{bias}}} + \underbrace{\sqrt{\frac{p}{n_{\calK}}}}_{\substack{\text{commonality} \\ \text{estimation error} 
	}} + \underbrace{\sqrt{\frac{p\log p}{n_0}}\wedge \Delta_{h_r}}_\text{task-difference bias}
	\Big]  + \underbrace{\frac{\gamma^{\Upsilon}R_{\max}}{(1-\gamma)^2}}_\text{algorithmic error}\bigg),
	%\end{aligned} 
\end{equation}
where $ \Delta_{h_r}:=
\sqrt{h_r}\big(\frac{\log p}{n_0}\big)^{1/4}\wedge h_r$, and $\widehat{\pi}_{\Upsilon}$ is the output policy of Algorithm \ref{algo:fitted-q-transfer}. 
\end{theorem}
%\wenbo{The original equation,
%	\[\EE\abs{Q^*-\hat Q^{(0)}_{\Upsilon}}  = \cO_p \bigg(\frac{\gamma}{(1-\gamma)^2} \Big[ \underbrace{p^{-\kappa/d}}_{\substack{\text{function-}\\\text{approximation} \\ \text{bias}}} + \underbrace{\sqrt{\frac{p \log^2 n_{\calK}}{n_{\calK}}}}_{\substack{\text{commonality} \\ \text{estimation error} 
		%	}} + \underbrace{\sqrt{h_r}\paran{\frac{\log p}{n_0}}^{1/4}}_\text{task-difference bias}
%	\Big]  + \underbrace{\frac{\gamma^{\Upsilon+1}R_{\max}}{(1-\gamma)^2}}_\text{algorithmic error}\bigg),\] 
%	is actually a result on $\EE\abs{Q^*-Q^{\hat \pi_{\Upsilon}}}$, which is 0 when $\gamma=0$. The correct result for $\EE\abs{Q^*-\hat Q^{(0)}_{\Upsilon}}$ should be
%	\[\EE\abs{Q^*-\hat Q^{(0)}_{\Upsilon}}  = \cO_p \bigg(\frac{1}{1-\gamma} \Big[ \underbrace{p^{-\kappa/d}}_{\substack{\text{function-}\\\text{approximation} \\ \text{bias}}} + \underbrace{\sqrt{\frac{p \log^2 n_{\calK}}{n_{\calK}}}}_{\substack{\text{commonality} \\ \text{estimation error} 
		%	}} + \underbrace{\sqrt{h_r}\paran{\frac{\log p}{n_0}}^{1/4}}_\text{task-difference bias}
%	\Big]  + \underbrace{\frac{\gamma^{\Upsilon}R_{\max}}{(1-\gamma)}}_\text{algorithmic error}\bigg).\] 
%}

%\wenbo{RL:error propagation $\gamma$, semi-para:bias}
Theorem \ref{thm:Q-converge-homo} establishes an error bound for the expected return of the estimated policy $\widehat{\pi}_{\Upsilon}$, %$Q^*$ estimator 
containing a statistical rate (the first three terms) and an algorithmic rate (the last term). 
%The statistical rate is weighted by $\frac{1}{(1-\gamma)^2}$, which is incurred by the iterative nature of the fixed-point solution. 
The algorithmic error is due to the error of the initial estimator, which decreases exponentially as the iterations proceed since $\gamma<1$. 
By choosing sufficiently large the number of iterative steps $\Upsilon$, the statistical error will dominate the algorithmic error. 

We now focus on discussing the statistical rates in the brackets in \eqref{eqn:rate-homo}. The first term $p^{-\kappa/d}$ is the function approximation bias, instantiated as the bias of non-parametric estimation in \eqref{eq:property}. %Due to the careful design of our algorithm, this estimation bias does not accumulate over different tasks. 
It is consistent with the bias agreed in the literature on sieve approximation. We conjecture that this function approximation error will be instantiated by other estimation biases if we consider different function classes $\calQ$ and approximation techniques, such as neural network function approximation. We leave the detailed analysis of other function approximation methods for future works. 

The second term $\sqrt{\frac{p}{n_{\calK}}}$ represents the statistical convergence rate of estimating the shared commonality between the target and the source tasks, i.e., the center $W^*$ in \eqref{eqn:W*}, demonstrating itself in the order of standard deviation. 
For sieve approximation, by choosing the number of basis functions $p > n_{\calK}^{\frac{d}{2\kappa+d}}$, %(\log n_{\calK})^{\frac{-2d}{2\kappa+d}}$, 
this standard deviation term dominates function-approximation bias term $p^{-\kappa/d}$. Compared to the single task convergence rate $\sqrt{\frac{p}{n_0}}$, the commonality estimation error is improved into the convergence rate when the target sample size is as large as $n_{\calK}$, which shows the advantage of transfer learning.

The third term $\sqrt{\frac{p\log p}{n_0}}\wedge
\sqrt{h_r}\big(\frac{\log p}{n_0}\big)^{1/4}\wedge h_r$ is a bias term incurred by task discrepancy and mathematically characterized by $h_r$ defined in \eqref{eq:h-homo}, which demonstrates a piecewise rate based on the scale of $h_r$. %If $h_r \lesssim \sqrt{\log p/n_0}$, the task-difference bias is of the rate $h_r$.  
If $h_r \gtrsim p\sqrt{\frac{\log p}{n_0}}$, the statistical rate in \eqref{eqn:rate-homo} reduces to $O_{\PP}\Big(\frac{1}{(1-\gamma)^2}\big(\sqrt{\frac{p \log p}{n_0}}+p^{-\kappa/d}\big)\Big)$, matching the statistical rate of single-task non-parametric FQI without knowledge transfer, ignoring the logarithm term. Otherwise, if $ h_r \lesssim p\sqrt{\frac{\log p}{n_0}}$, this term becomes $\sqrt{h_r} \big(\frac{\log p}{n_0}\big)^{1/4} \wedge h_r$, which is an improvement compared to the non-transfer rate.  Therefore, our algorithm achieves a statistical error rate that is no worse than that of a non-transfer single-task FQI  and benefits from knowledge transfer as long as $h_r \lesssim p\sqrt{\frac{\log p}{n_0}}$ and $n_{\calK} \gtrsim n_0$,  demonstrating the value of transferring knowledge from similar and sufficiently large source tasks.

The above condition on task discrepancy, $h_r \lesssim p\sqrt{\frac{\log p}{n_0}}$, is mild, since the basis number $p$ is allowed to be as large as $n_{\calK}$, which allows the upper bound of $h_r$ to be of the order $\frac{n_{\calK}\sqrt{\log p}}{\sqrt{n_0}}$. Specifically, we provide a data-driven method to choose the number of basis functions $p$ in Section \ref{sec:choose-p}.
In addition, we note that the conditions for the initial estimators assumed in Theorem \ref{thm:Q-converge-homo} can be easily satisfied in practice without knowing $h_r$, for example, by setting $\hbbeta^{(k)}_0=0$ for all $k$, and the constant $c_{\delta}$ can be chosen by cross-validation.
%\end{proof}
\begin{rmk}\label{rmk:target-size}
The result in Theorem \ref{thm:Q-converge-homo} also provides an insight  into %the practice of TL for offline RL. Firstly, for a given task difference $h_r$, it spells out 
the necessary numbers of target and source samples to correct the task-difference bias and to fully enjoy the benefit of transfer learning. 
Specifically, for a fixed $h_r$ that satisfies $h_r \lesssim p\sqrt{\frac{\log p}{n_0}}$, the task-difference bias can be dominated by the commonality estimation error if practitioners collect $n_0 \gtrsim n_{\calK}^2 h_r^2 p^{-2}$ samples for the target task, ignoring the logarithm term.

This implies an artful tug-of-war between the source and the target tasks when task discrepancy exists: on the one hand, one wishes to have a large size of source data to improve the estimation accuracy of the common component shared by the source and target tasks; on the other hand, one needs to control the relative sample sizes of the source and target data such that the bias induced from the task difference can be corrected with $n_0$ target samples. 
The growth of the required $n_0$ is proportional to  $n_{\cK}^2p^{-2}$ for any fixed $h_r$, which is not restrictive in the sense that the number of $p$ is growing and $n_{\cK}^2 p^{-2}$ can be much smaller than $n_{\cK}$, as we only require $\frac{p \log^2 n_{\calK}}{n_{\calK}}=o(1)$ in Theorem \ref{thm:Q-converge-homo}.%, the growth of $p$ can be as fast as $n_{\calK}/\log^2n_{\calK}$.
%As a result, with the help of transfer learning, the target RL task with sample size $n_0$ much smaller than $n_{\cK}$ can enjoy the convergence rate of $\sqrt{p/n_{\cK}}$. 
\end{rmk}

\begin{rmk}\label{rmk:slower-rate}
In Theorem \ref{thm:Q-converge-homo}, the rate of the task-difference bias is generally larger than the commonality estimation error since the bias correction must rely on the limited target data to estimate the discrepancy between the target task and the center $W^*$ in \eqref{eqn:W*}, whereas the commonality estimation error  benefits from the aggregated sample size. Achieving a fast rate $\sqrt{\frac{p}{n_{\calK}}}$ would require either a sufficiently small task-discrepancy $h_r$ or a sufficiently large target sample size $n_0$, as discussed in Remark \ref{rmk:target-size}.  Similar rates are common in the knowledge-transfer literature (e.g., \citealp{li2022transfer-jrssb, tian2023transfer, cai2024transfer}) and are therefore not unique to reinforcement learning.
\end{rmk}

%In cases where source tasks offer no data, meaning $n_{\calK}=n_0$ and $h_r=0$, the performance of our algorithm matches the statistical convergence rate of single-task non-parametric RL estimation, with a rate of $\calO_p\Big(\frac{\gamma}{(1-\gamma)^2}\big(\sqrt{\frac{p \log^2n_0}{n_0}}+p^{-\kappa/d}\big)\Big)$. As the source sample size increases relative to the target, specifically when $n_{\calK} \gtrsim n_0$ and $h_r \lesssim \frac{p}{\sqrt{n_0\log p}}$, our algorithm's accuracy improves, demonstrating the value of transferring knowledge from similar and sufficiently large source tasks. The condition on task discrepancy, $h_r \lesssim \frac{p}{\sqrt{n_0\log p}}$, is mild, since the basis number $p$ is allowed to be as large as $n_{\calK}$, which allows the upper bound of $h_r$ to be of the order $\frac{n_{\calK}}{\sqrt{n_0}}$, ignoring the logarithm term. 

\subsection{Select the Number of Basis Functions} \label{sec:choose-p}

In this section, we provide a data-adaptive approach for selecting the number of basis function $p$, which carefully balances the trade-off between the commonality estimation error $\sqrt{\frac{p}{n_{\calK}}}$ and the function approximation bias $p^{-\kappa/d}$. We present the key ideas and summarize the theoretical properties here, while the detailed algorithm, its illustration, and complete theoretical analysis are provided in Section C of the supplementary materials.

Theoretically, the choice $p^*\asymp n_{\calK}^{d/(2\kappa+d)}$ balances the commonality estimation error and the function approximation bias and achieves the statistical rate $n_{\calK}^{-\kappa/(2\kappa+d)}\vee \Delta_{h_r}$, where $\Delta_{h_r}$ represents the task-difference bias that arises in knowledge transfer. However, this selection faces a key challenge: the smoothness order $\kappa$ is typically unknown in practice. While one could conservatively assume $\kappa=1$, this leads to suboptimal performance when the true smoothness is higher and the task-difference is small. Moreover, the task-difference bias $\Delta_{h_r}$ may depend implicitly on $p$ through the task discrepancy $h_r$ defined in \eqref{eq:h-homo}, making it more challenging to optimize the error rate in \eqref{eqn:rate-homo} as the relationship between $h_r$ and $p$ lacks an explicit form. 

To address this challenge, we develop a data-adaptive algorithm that selects the number of basis functions $p$ without requiring knowledge of $\kappa$. Inspired by Lepskii's approach \citep{lepskii1991problem}, the key idea is to identify the smallest $p$ for which the commonality estimation error dominates the approximation bias, which should be of the same order as $p^*$. Specifically, we select $p$ from a candidate set $\{p_g=2^g \mid g=0,1,\dots,g_{\max}\}$ by examining the $\ell_1$ differences $\varrho_{g,g'}$ between estimators obtained with different numbers of basis functions $p_g=2^g$ and $p_{g'}=2^{g'}$. These differences serve as estimators for the error rates and guide the construction of a set $\widehat{\calG}:=\min\big\{g: \varrho_{g, g'} \leq \frac{\widetilde{C}}{1-\gamma} \sqrt{\frac{p_{g'}}{n_{\calK}}}, \, \forall g \leq g' \big\}$. %Intuitively, $\widehat{\calG}$ identifies all candidates $p_g$ such that the commonality estimation error $\sqrt{p_{g'} / n_{\calK}}$ dominates for every larger candidate $p_{g'} \geq p_g$. 
The final number of basis functions is then set as $\widehat{p}=2^{\widehat{g}}$, where $\widehat{g}$ is the minimum element in $\widehat{\calG}$. 

As a theoretical guarantee, we prove in Theorem C.1 that our algorithm achieves a statistical rate of $n_{\calK}^{-\kappa/(2\kappa+d)} \vee \widetilde{\Delta}$ with the number of basis functions $p=\widehat{p}$ under mild conditions, where $\widetilde{\Delta}$ is an upper bound for the task-difference bias near $p^*$. %When $\widetilde{\Delta}$ is dominated by $n_{\calK}^{-\kappa/(2\kappa+d)}$, our selection method achieves the optimal statistical rate $n_{\calK}^{-\kappa/(2\kappa+d)}$ as if $\kappa$ were known. % However, when $\widetilde{\Delta}$ exceeds $n_{\calK}^{-\kappa/(2\kappa+d)}$, the error rate may be larger than the minimum achievable rate since $\widetilde{\Delta}$ serves as an upper bound. Therefore, our selection method handles the unknown smoothness $\kappa$ at the cost of a potential increase when the task-difference bias is large. 
To ensure robustness in cases where the task-difference bias is substantial near $p^*$, we incorporate an additional safeguard that guarantees our algorithm performs at least as well as non-transfer FQI. Details are provided in Section C of the supplementary materials.

\begin{rmk}\label{rmk:h-verify}
Our $p$-selection method serves as a data-adaptive way to verify the condition $h_r\sqrt{\frac{\log p}{n_0}}=o(1)$ required in Theorem \ref{thm:Q-converge-homo}. When this condition is violated, the task-difference bias $\Delta_{h_r}$ may exceed one, potentially leading to algorithmic divergence as errors accumulate through iterations. Our selection approach addresses this issue by choosing $\widehat{p} \in \widehat{\calG}$ such that the commonality estimation error $\sqrt{\frac{\widehat{p} }{ n_{\calK}}}$ dominates other terms. By restricting $g_{\max}$ to be less than $\lfloor\log_2 n_{\calK}\rfloor$, we ensure that the dominant term $\sqrt{\frac{\widehat{p}}{n_{\calK}}} = o(1)$ for all $p \in \calP$, thereby preventing divergence. In extreme cases where the task discrepancy $h_r$ is so large that Algorithm \ref{algo:fitted-q-transfer} diverges for any choice of $p$, one can resort to non-transfer FQI on the target task, as the source task becomes non-transferable.
\end{rmk}

\subsection{Theoretical Results for Transition Heterogeneous Tasks}\label{sec:res-hetero}

In this section, we generalize the results in Section \ref{sec:res-homo} to transition heterogeneous tasks, allowing the distributions of $(\bX, A)$ to differ in different tasks. 
Therefore, both the transition kernels $P^{(k)}(\bx'\,|\,\bx, a)$ and the reward functions $r^{(k)}(\bx, a)$ are different across tasks. 
The theoretical results explicitly depend on the differences between the target and the source tasks defined by \eqref{eqn:delta-r} and \eqref{eqn:delta-p}. %We first extend Assumption \ref{assum:distribution_homo} to the heterogeneous setting.

We first characterize the transferability across different tasks in the heterogeneous setting. %Recall that $\bxi(\bx, a):=\left[\bphi^{\top}(\bx)\II(a=1),\bphi^{\top}(\bx)\II(a=2),\dots,\bphi^{\top}(\bx)\II(a=m)\right]^{\top}$. 
By \eqref{eq:property} and Assumption  \ref{assum:kappa-smooth}, there exist coefficients $\big\{\bbeta^{(k)}_{r}\big\}$ and $\big\{\bbeta^{(k)}_{\rho}(\bx')\big\}$ such that 
\[\sup_{\bx, a, k} \abs{r^{(k)}\paran{\bx, a} - \bxi^{\top}(\bx, a)\bbeta^{(k)}_{r}} \leq Cp^{-\kappa/d},\]
and
\[\sup_{\bx, a, \bx', k} \abs{\rho^{(k)}\paran{\bx' \,|\, \bx, a} - \bxi^{\top}(\bx, a)\bbeta^{(k)}_{\rho}(\bx')} \leq Cp^{-\kappa/d},\]
for some constant $C$. Similar to the transition homogeneous setting, the similarity of $r^{(k)}$ and $\rho^{(k)}$ is measured through their projections on the sieve space. Specifically, define
\begin{equation}\label{eq:h-hetero}
h:=\max_{k}\left[\norm{\bbeta^{(k)}_{r}-\bbeta^{(0)}_{r}}_1+\int_{\calX}\norm{\bbeta^{(k)}_{\rho}(\bx')-\bbeta^{(0)}_{\rho}(\bx')}_1 \mathrm{d}\bx'\right],
\end{equation}
which is a generalization of $h_r$ defined in \eqref{eq:h-homo}  in the homogeneous setting. %We then connect $h$ to the difference in the optimal $Q^*$ function.

%{\red 
% \begin{lemma}\label{lem:h-hetero} Under Assumptions \ref{assum:distribution_hetero} and \ref{assum:kappa-smooth-hetero}, there exists coefficients $\big\{\bbeta_Q^{(k)}\big\}$ such that
	% \[\sup_{\bx, a, k} \abs{Q^{*(k)}\paran{\bx, a} - \bxi^{\top}(\bx, a)\bbeta^{(k)}_{Q}} \leq C'p^{-\kappa/d},\]
	% for some absolute constant $C'$, which satisfies
	% \[\max_k\big\|\bbeta^{(k)}_{Q}-\bbeta^{(0)}_{Q} \big\|_1 \lesssim h.\]
	% \end{lemma}}
	% Lemma \ref{lem:h-hetero} can be viewed as an instantiation of Lemma \ref{lem:diff-Q*} in sieve space, indicating that the discrepancy of $Q^*$ functions can be characterized by the difference in MDP tasks, i.e., rewards functions and transition kernels, which justifies our definition for $h$ in \eqref{eq:h-hetero}.}
In addition, since $Q^{*(k)}$ is approximated by $\bxi(\bx, a)^{\top}\bbeta_{\Upsilon}^{(k)}$, we also need to characterize the discrepancy between the population covariance matrices of $\bxi(\bx, a)$. Let $\overline{\bSigma}=\sum_{k=0}^{K}\frac{n_k}{n_{\calK}}\bSigma^{(k)}$, where $\bSigma^{(k)}$ is defined in condition (c) in Assumption \ref{assum:distribution}. %, and $n_{\calK}:=\sum_{k=1}^{K}n_k$ is the total sample size of the auxiliary tasks. 
Then we define
\begin{equation}
C_{\Sigma}:=1+\max_{k}\norm{\overline{\bSigma}^{-1}(\bSigma^{(k)}-\overline{\bSigma})}_1.
\label{eq:def-CSigma}
\end{equation}
The quantity $C_{\Sigma}$ is similar to the heterogeneity constant defined in \cite{li2022transfer-jrssb}. However, \cite{li2022transfer-jrssb}  characterized the differences between the covariance matrices  $\bSigma^{(k)}$ and $\bSigma^{(0)}$ directly, while we characterize the differences between $\bSigma^{(k)}$ and their weighted average for technical simplicity. 
We are ready to present the theoretical property of Algorithm \ref{algo:fitted-q-transfer} in the transition heterogeneous setting.

\begin{theorem} \label{thm:Q-converge-hetero}
Suppose Assumptions \ref{assum:distribution}, \ref{assum:kappa-smooth}, and the other conditions in Theorem \ref{thm:Q-converge-homo} hold with $C_{\Sigma}h$ replacing $h_{r}$. 
%In addition, $n_0 > \log(p) (C_{\Sigma}h)^{-2}$
Then we have
\begin{equation}\label{eqn:rate-hetero}
%\begin{aligned} 
v^{\pi^*}-v^{\widehat{\pi}_{\Upsilon}} = O_{\PP} \bigg(\frac{1}{(1-\gamma)^2} \Big[ \underbrace{p^{-\kappa/d}}_{\substack{\text{function-}\\\text{approximation} \\ \text{bias}}} + \underbrace{\sqrt{\frac{p}{n_{\calK}}}}_{\substack{\text{commonality} \\ \text{estimation error} 
}} + \underbrace{\sqrt{\frac{p\log p}{n_0}}\wedge \Delta_{C_{\Sigma}h}}_\text{task-difference bias}
\Big]  + \underbrace{\frac{\gamma^{\Upsilon}R_{\max}}{(1-\gamma)^2}}_\text{algorithmic error}\bigg),
%\end{aligned} 
\end{equation}
where $ \Delta_{C_{\Sigma}h}:=
\sqrt{C_{\Sigma}h}\big(\frac{\log p}{n_0}\big)^{1/4}\wedge C_{\Sigma}h$, and $\widehat{\pi}_{\Upsilon}$ is the output policy of Algorithm \ref{algo:fitted-q-transfer}. 
\end{theorem}
%\wenbo{RL:error propagation $\gamma$, semi-para:bias}
Compared to the convergence rate in the homogeneous setting (Theorem \ref{thm:Q-converge-homo}), Theorem \ref{thm:Q-converge-hetero} indicates that Algorithm \ref{algo:fitted-q-transfer} achieves a similar convergence rate for the transition heterogeneous tasks. Concretely, the function-approximation bias, commonality estimation error, and algorithmic error terms (i.e., the first, second, and fourth terms) are of the same rate, while the third term, which is due to the discrepancy among the tasks,  is characterized by $C_{\Sigma}h$ instead of $h_r$. 

Similar to Remark \ref{rmk:target-size} for Theorem \ref{thm:Q-converge-homo}, we have that, when $\Upsilon$ is sufficiently large, the target sample size $n_0$ needed to correct the task-difference bias is $n_{\calK}^2C_{\Sigma}^2h^2p^{-2}$, where $p$ is allowed to grow as fast as $n_{\calK}/\log^2n_{\calK}$. %Moreover, given a fixed $n_0$, the statistical convergence rate of our proposed algorithm is sharper than trivial FQI without transferring as long as $C_{\Sigma}h \lesssim \frac{p}{\sqrt{n_0\log p}}$ and $n_{\calK} \gtrsim n_0$. 
We note that Theorem \ref{thm:Q-converge-hetero} is consistent with Theorem \ref{thm:Q-converge-homo} since $C_{\Sigma}=1$ and $h=h_r$ for the transition homogeneous setting. In the heterogeneous case, where transition kernels differ across tasks, we have $h>h_r$ and $C_{\Sigma}>1$, leading to $C_{\Sigma}h>h_r$. This larger discrepancy indicates that more target samples are required to achieve the desired rate $\sqrt{\frac{p}{n_{\calK}}}$ in the transition heterogeneous setting compared to the homogeneous case. The data-adaptive method presented in Section \ref{sec:choose-p} is applicable to the transition heterogeneous setting for selecting $p$ and, as noted in Remark \ref{rmk:h-verify}, for verifying the condition $C_{\Sigma}h\sqrt{\frac{\log p}{n_{0}}}=o(1)$.

In this study, we do not specify a structure for the transition probability $P^{(k)}$ nor delve into its transfer learning aspects. However, recent research \citep{lu2021power, cheng2022provable,agarwal2023provable,chai2025transition}  highlights the benefits of leveraging $P^{(k)}$'s shared low-rank structure to enhance the estimation of the target task's $Q^*$ functions. A promising avenue for future investigation is integrating $P^{(k)}$'s low-rank structure into Transfer FQI, particularly in settings with heterogeneous transitions.

\section{Empirical Studies}\label{sec:empirical}

\subsection{Simulations}
\label{sec:simul}

In this section, we demonstrate the advantage of our proposed Algorithm \ref{algo:fitted-q-transfer} through simulation studies. We choose the state space $\calX$ to be $[-1, 1]^3$, set the action space $\calA$ to be $\{-1, +1\}$, and generate each trajectory by the equation:
% \begin{equation*}
% 	\bx_{t+1} = \frac{3}{4}\begin{pmatrix}
% 		a_{t}&&\\
% 		&-a_{t}&\\
% 		&& a_{t}
% 	\end{pmatrix}\bx_{t} + \epsilon_{x, t}, 
% \end{equation*} 
$\bx_{t+1} = 0.75 \cdot {\rm diag}(a_{t}, -a_{t}, a_{t})\cdot\bx_{t} + \epsilon_{x, t}$, 
where ${\rm diag}(\cdot)$ is a diagonal matrix, $\bx_{0} \sim \calN(0, I_3)$, $a_{t} \sim \mathrm{Bernoulli}(0.5)$, and $\epsilon_{x, t} \sim \mathcal{N}(0, I_3/4)$. We use a tanh function to map each state vector $\bx_t$ onto $[-1, 1]$. This three-dimensional setting is partly adapted from the two-dimensional setting in \cite{shi2020statistical}. We then apply a quadratic reward function
$r_t = a_{t} * 	(\bx_{t}^{\top}\bC	\bx_{t}) +  \epsilon_{r, t}$,
where $\epsilon_{r, t} \sim \calN(0, I_3/4)$ and $\bC$ is a random matrix to be specified. 

Following the settings above, we generate a target task and a source task, with index $k=0$ and $1$, respectively. The number of trajectories in task $k$ is denoted by $I^{(k)}$, and we fix the length of each trajectory to be $5$. In practice, we choose $I^{(0)}=20$ and let $I^{(1)}$ range from $10$ to $80$. For the target task, the diagonal entries of the matrix $\bC^{(0)}$ are independently drawn from $\calN(0, 1)$, while the off-diagonal entries are independently drawn from $\calN(0, 1/4)$. For the source task, we set $\bC^{(1)}=\bC^{(0)}+\bC_{\delta}$, where each entry of $\bC_{\delta}$ is independently drawn from $\calN\big(0, \sigma_C^2\big)$. The parameter $\sigma_C$ controls the level of discrepancy between the target task and the source task, whose possible values are set as $0.25$, $0.5$, $0.75$, and $1.0$ in our experiments. 

\begin{figure}[!tp]
\centering
\includegraphics[width=0.65\linewidth]{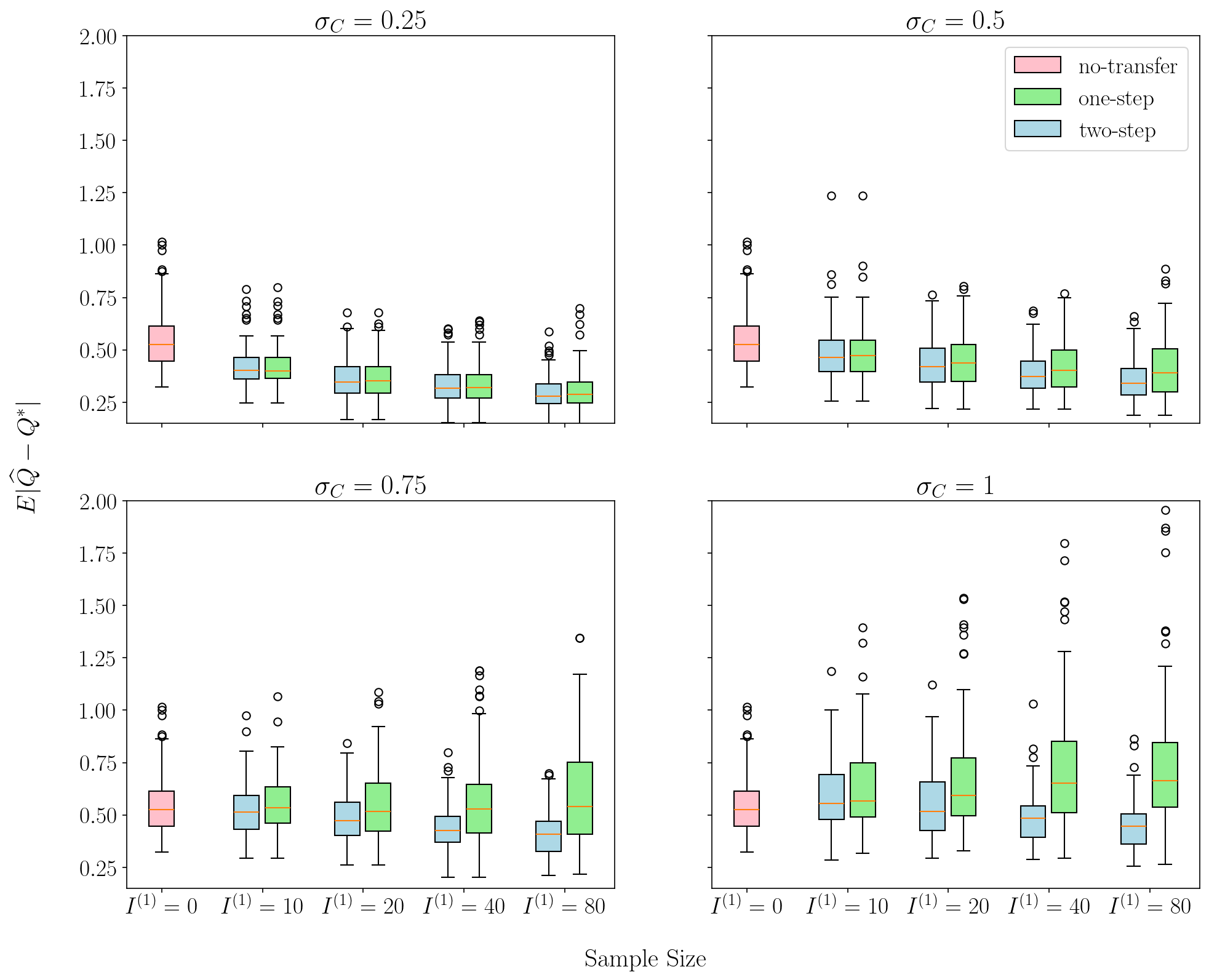}
\caption{Boxplots of the estimation errors for the $Q^*$ function with $I^{(0)}=20$ and different source sample sizes $I^{(1)}$. The parameter $\sigma_C$ on the top of each subfigure indicates the standard deviation of the difference matrix $\bC_{\delta}$.}
\label{fig:boxplot_quadratic_noise0.5}
\end{figure}

To verify the effect of our proposed two-step algorithm, we compare the following three methods: (1)``no-transfer'': Fitted Q-Iteration on the target task without transferring any information from the source task; (2) ``one-step'': Fitted Q-Iteration on the aggregated data, without correcting the difference (i.e., only do Step \rom{1} in Algorithm \ref{algo:fitted-q-transfer} for each iteration); and (3) ``two-step'': our proposed method (Algorithm \ref{algo:fitted-q-transfer}).
% \begin{itemize}
% \item[(1)]``no-transfer'': Fitted Q-Iteration on the target task without transferring any information from the source task.
% \item[(2)] ``one-step'': Fitted Q-Iteration on the aggregated data, without correcting the difference (i.e., only do Step \rom{1} in Algorithm \ref{algo:fitted-q-transfer} for each iteration). 
% \item[(3)] ``two-step'': our proposed method (Algorithm \ref{algo:fitted-q-transfer}). 
% \end{itemize}
For all algorithms, the basis function $\bphi(\bx)$ is chosen to be a set of three-dimensional B-splines on $[-1, 1]^3$ for all methods, each entry of the initial estimator $\hat\bbeta_0^{(k)}$ is independently drawn from $\calN(0, 0.01)$, and the regularization parameters $\lambda_\delta^{(k)}$ are determined through cross validation. To clearly evaluate the estimation performance of each method, we compare the expected estimation error $\EE\big|\hat Q(\bx, a) - Q^*(\bx, a)\big|$ of each method instead of the regret $v^{\pi^*}-v^{\hat{\pi}}$, where the expectation is estimated by taking an average over 200 values of $(\bx, a)$ that are independently drawn from the true distribution of $(\bx_0, a_0)$.
% Since it is hard to compute the $Q^*$ function in closed form, we also use Monte Carlo approximations to obtain the true value of $Q^*$ for any fixed $(\bx, a)$. 
The results averaged over 100 independent runs are displayed by boxplots in Figure \ref{fig:boxplot_quadratic_noise0.5}.
\begin{figure}[!t]
\centering
\begin{subfigure}[b]{0.45\textwidth}
\centering
\includegraphics[width=\textwidth]{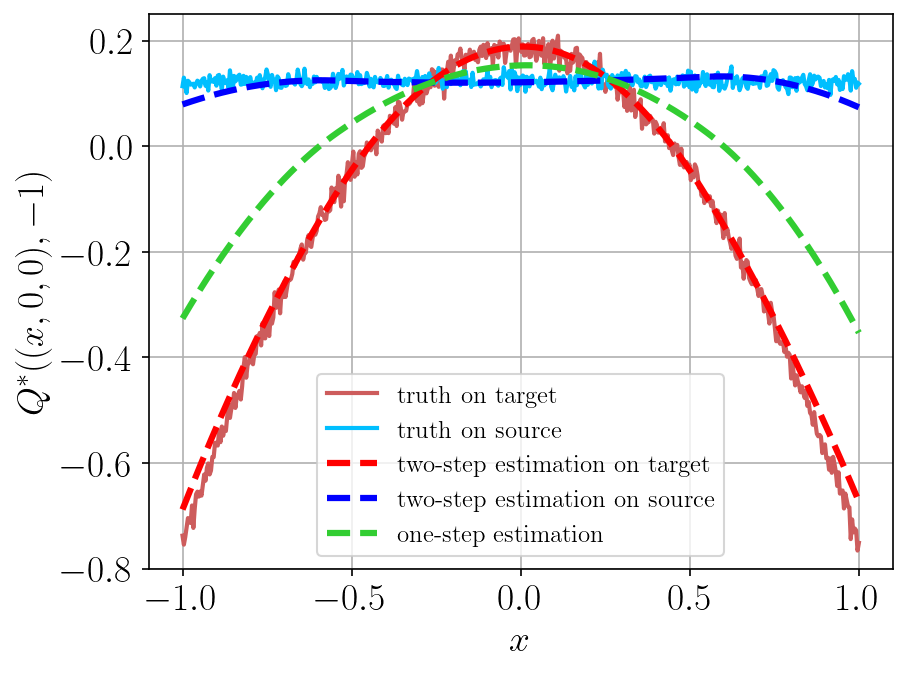}
\subcaption{$a=-1$}
\label{fig:Q1}
\end{subfigure}
\hfill
\begin{subfigure}[b]{0.45\textwidth}
\centering
\includegraphics[width=\textwidth]{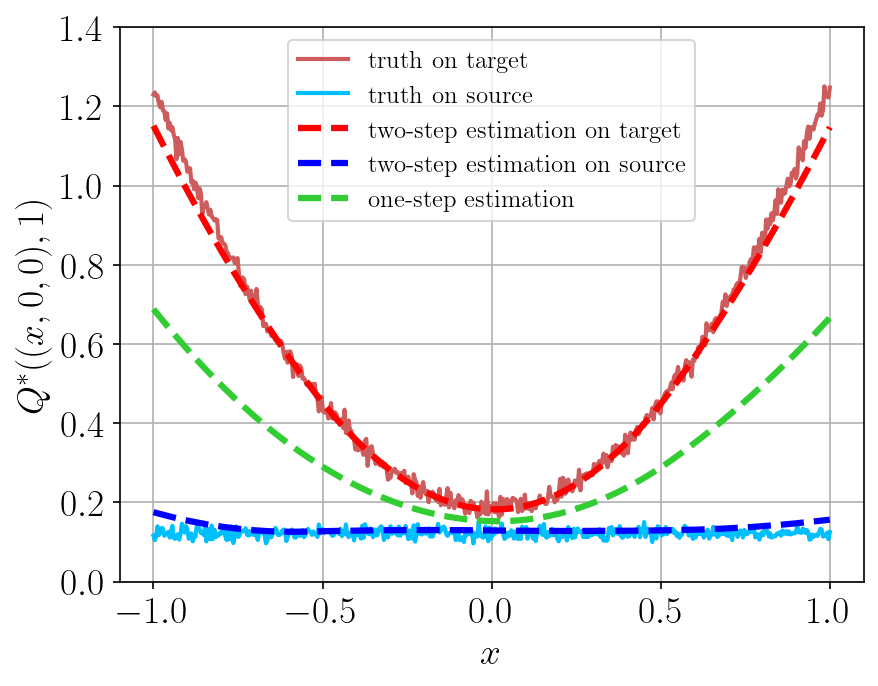}
\subcaption{$a=1$}
\label{fig:Q2}
\end{subfigure}
\caption{The estimated $Q^*$ values obtained by the ``one-step'' and ``two-step'' methods for $Q^*((x, 0, 0), -1)$ (left) and $Q^*((x, 0, 0), 1)$ (right). The legend ``truth'' represents the true value of the $Q^*$ function approximated by Monte Carlo simulation.}%, and ``two-step estimation'' represents the estimated $Q^*$ function obtained by the two-step method. %For clear presentation, the estimated $Q^*$ function of the ``no-transfer'' method is omitted since its scale is too large. 
%The matrices in the reward function are fixed to be $\bC^{(0)}=\mathrm{diag}(1, 0, 0)$ and  $\bC^{(1)}=\mathrm{diag}(0, 0, 0)$.}
\label{fig:Q-functions}
\end{figure} 

As shown in Figure \ref{fig:boxplot_quadratic_noise0.5}, when the task discrepancy is small (i.e., $\sigma_C=0.25$), the one-step and the two-step methods both significantly outperform the ``no-transfer'' FQI method, and their performance improves as the source sample size $I^{(1)}$ increases. However, the one-step method's performance deteriorates as the task discrepancy $\sigma_C$ grows larger. Particularly, when $\sigma_C \geq 0.75$, it performs even worse than the ``no-transfer'' method due to the significant bias induced by the source task. On the contrary, the proposed two-step method consistently performs well with sufficiently large source sample size, highlighting the importance of the second step in our proposed algorithm, which corrects the bias of the center estimator obtained in the first step. 

To further illustrate the learning outcomes of these methods, in Figure \ref{fig:Q-functions}, we juxtapose the estimated $Q^*$ values against the truth for $\bx=(x, 0, 0)$, with $x$ ranging between $[-1, 1]$, and $a=\pm 1$. For the target task, we set $\bC^{(0)}=\mathrm{diag}(1, 0, 0)$ and $I^{(0)}=40$, while for the source task, $\bC^{(1)}=\mathrm{diag}(0, 0, 0)$ and $I^{(1)}=40$. The comparison, as shown in Figure \ref{fig:Q-functions}, reveals that the two-step algorithm accurately estimates the true $Q^*$ values for both tasks, in contrast to the one-step method, which yields less accurate intermediate estimates far from either task.

%To further visualize what different methods learn, we compare a marginal version of the estimated $Q^*$ functions and the true $Q^*$ function in Figure \ref{fig:Q-functions}, where we plot the optimal $Q^*$ function for $\bx=(x, 0, 0)$ ($x\in [-1, 1]$) and $a=-1$. We let $\bC^{(0)}=\mathrm{diag}(1, 0, 0)$, $I^{(0)}=40$ for the target task and $\bC^{(1)}=\mathrm{diag}(0, 0, 0)$, $I^{(1)}=40$ for the source task. Figure \ref{fig:Q-functions} indicates that our two-step method performs well in estimating the true $Q^*$ value on both tasks, while the one-step estimation only obtains intermediate estimated values far from either task.  

\subsection{Real Data Analysis}
\label{sec:appl}

In this section, we apply the proposed Algorithm \ref{algo:fitted-q-transfer} on Medical Information Mart for Intensive Care (MIMIC-\rom{3}) dataset \citep{johnson2016mimic} to illustrate the benefit of knowledge transfer in $Q^*$ learning. 

MIMIC-\rom{3} is a large, publicly available database of medical records containing de-identified health-related data about patients admitted to critical care units at a large tertiary care hospital. In particular, we follow the procedure of \cite{komorowski2018artificial} to select the data of the adult sepsis patients and extract a set of features for characterizing each patient, including demographics, Elixhauser premorbid status, vital signs, laboratory values, fluids and vasopressors received. To save the computation time, we further compute the top 10 principal components of the features to be the state variable $\bX_{i, t} \in \RR^{10}$. 

The action variables of interest are the total volume of intravenous (\rom{4}) fluids and the maximum dose of vasopressors administrated over each period. Each action variable is discretized into three levels (low, medium, and high); hence, there are nine possible action combinations. For the rewards $R_{i, t}$, we follow \cite{prasad2017reinforcement} and \cite{komorowski2018artificial} to assign rewards to each state based on the health measurement and mortality of the patient. A higher reward $R_{i, t}$ indicates a better physical condition of the patient $i$ after the action $A_{i, t}$ taken at time $t$.

\begin{figure}[!tp]
\centering
\begin{subfigure}[b]{0.45\textwidth}
\centering
\includegraphics[width=\textwidth]{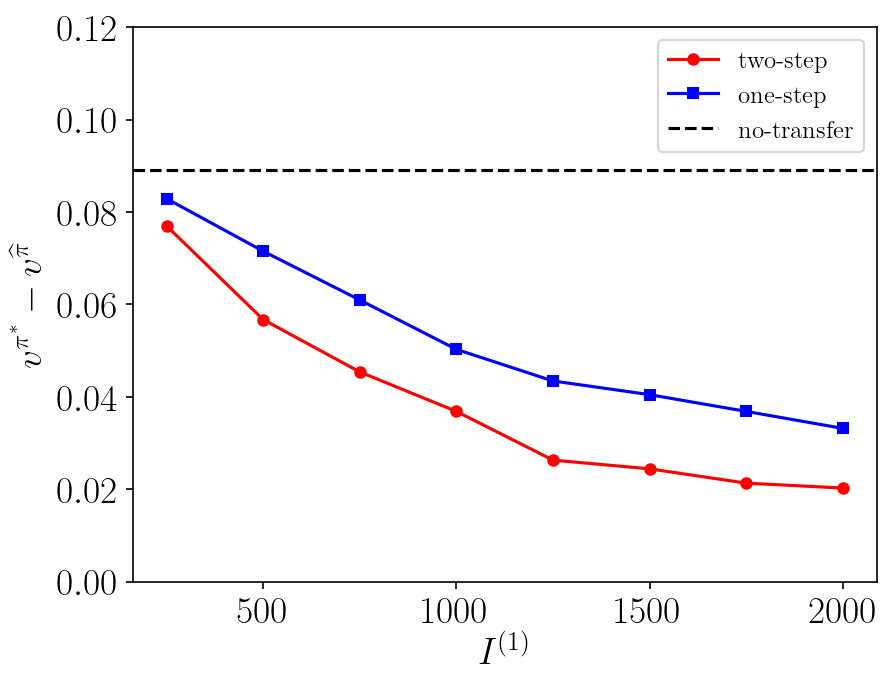}
\subcaption{$I^{(0)}=100$}
\label{fig:real_100}
\end{subfigure}
\hfill
\begin{subfigure}[b]{0.45\textwidth}
\centering
\includegraphics[width=\textwidth]{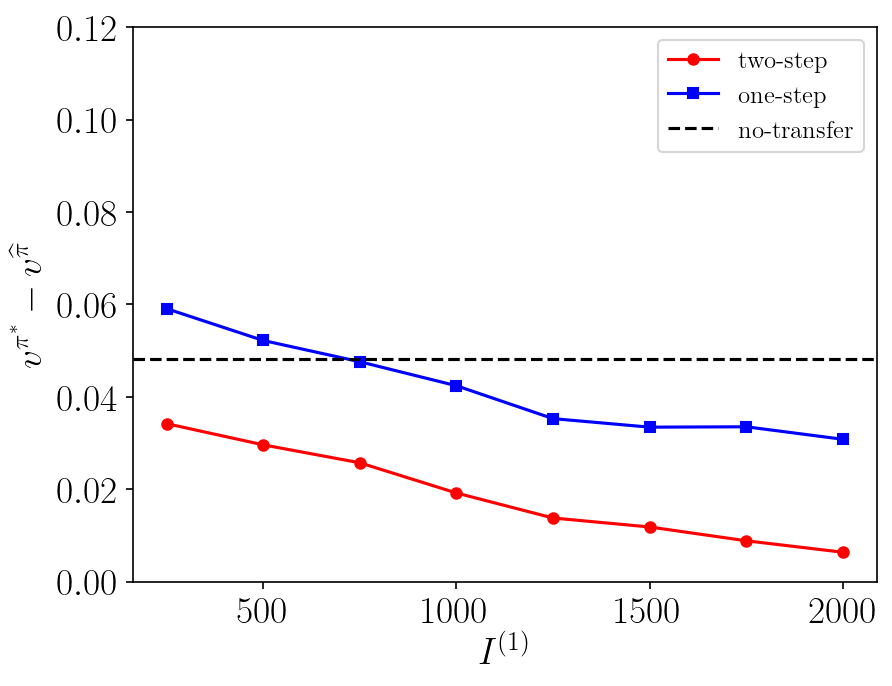}
\subcaption{$I^{(0)}=500$}
\label{fig:real_500}
\end{subfigure}
\caption{The regrets $v^{\pi^*}-v^{\widehat{\pi}}$ of the policies obtained by ``one-step'' and ``two-step'' algorithms with $I^{(0)}=100$ (left) and $I^{(0)}=500$ (right). The black dashed line shows the regret of FQI on the target task without knowledge transfer.}
\label{fig:real}
\end{figure}

The final processed dataset contains 278,598 observations from 20,943 patients, including 122,534 observations for female patients and 156,064 observations for male patients. We designate the male cohort as our target task $(k = 0)$ and the female cohort as the source task $(k = 1)$. For each task, we construct a calibrated environment by estimating the reward function, transition kernel, and initial distribution using the entire data task. Implementation details for environment calibration are provided in Section E of the supplementary materials.  

We then generate the source and target tasks from the calibrated environments and evaluate the expected returns of the three methods defined in Section \ref{sec:simul}. Specifically, we let $I^{(0)}\in\{100, 500\}$, $I^{(1)} \in [250, 2000]$, set the discount parameter $\gamma=0.6$, construct 10-dimensional B-splines as basis functions $\bphi(\bx)$, and determine the regularization parameter $\lambda_{\delta}^{(k)}$ by cross-validation.

Figure \ref{fig:real} presents the regret values $v^{\pi^*}-v^{\widehat{\pi}}$ for the three methods across different sample size configurations.  With limited target data $(I^{(0)} = 100)$, our  proposed ``two-step'' method shows steady improvement as source sample size increases, maintaining a clear advantage over both the ``no-transfer'' baseline and the ``one-step'' approach. When more target data is available ($I^{(0)} = 500$), the performance gap between ``one-step'' and ``two-step'' methods becomes more pronounced, as the 	``two-step'' method achieves near-zero regret with sufficient source data, while the ``one-step'' method consistently maintains higher regret. These results empirically validate our theoretical findings and demonstrate the practical value of our two-step transfer learning approach. 

\section{Conclusion}
\label{sec:conc}

This paper proposes a novel knowledge transfer framework for batch reinforcement learning. We introduce the Transfer Fitted Q-Iteration algorithm, an iterative procedure that learns shared structure and corrects task-specific bias within a general function approximation framework, and establish regret bounds under sieve approximation that reveal how rates depend on task similarity and sample sizes.
The results provide a unified perspective and practical guidance for when transfer is beneficial. A natural direction for future work is extending the framework to richer function classes such as deep neural networks.

%\section*{Supplementary Materials}
%
%The supplementary materials include the notations (Section A), further discussions and details for the proposed method (Sections B, C), the technical proof (Section D), and additional details in the real data analysis (Section E).

\vspace{-0.5cm}
\section*{Acknowledgment}
The authors gratefully thank the editor, the associate editor, and the anonymous referees for their helpful comments and suggestions. 

\vspace{-0.5cm}
\section*{Funding}
Elynn Chen’s research was supported in part by the National Science Foundation under Award ID 2412577. Xi Chen would like to thank the support from the National Science Foundation via the grant IIS-1845444.

\vspace{-0.5cm}
\section*{Disclosure Statement}
The authors report there are no competing interests to declare.

%%---------------------------------------------------
%%
%% Bibliography
%%
\spacingset{1.18}
\bibliographystyle{chicago}
\bibliography{rl-appl,rl-ml,rl-ms,rl-stat,tl}

%%---------------------------------------------------
%%
%% Appendix in a separate file
%%
\newpage
\setcounter{page}{1}
\appendix
\spacingset{1.8}
\begin{center}
{\large\bf Supplementary Materials of \\
    ``\TITLE''}
\end{center}

%\section{Proofs} \label{sec-proof}
%\listoffigures
%\listoftables

\section{Notations}

Let lowercase letter $x$, boldface letter $\bx$, capital letter $X$, boldface capital letter $\bX$, and calligraphic letter $\calX$ represent scalar, vector, random variable, random vector (or matrix), and set (or space), respectively.
We use the notation $[N]$ to refer to the positive integer set $\braces{1, \ldots, N}$ for $N \in \ZZ_+$.
For sequences $\{a_n\}$ and $\{b_n\}$, we write $a_n \lesssim b_n$ if $a_n \leq Cb_n$ for some constant $C$ that does not depend on $n$, and write $a \asymp b$ if $a_n \lesssim b_n$ and $b_n \lesssim a_n$. Moreover, for random variable sequences $\{X_n\}$ and $\{Y_n\}$, we let $X_n=\Op{Y_n}$ if $X_n / Y_n$ is bounded in probability.
We let $C$, $c$, $C_0$, $c_0$, $\ldots$ denote generic constants, where the uppercase and lowercase letters represent large and small constants, respectively. The actual values of these generic constants may vary from time to time.

As a convention, in this paper, we use upper case letters, such as $I$ and $T$, as the sizes of dimensions (fixed or growing). 
We use the corresponding lowercase letters, e.g., $i$ and $t$, as the running indices.

%For any matrix $\bX$,  we use $\bx_{i \cdot}$, $\bx_{\cdot j}$, and $x_{ij}$ to refer to its $i$-th row, $j$-th column, and $ij$-th entry, respectively. 
%All vectors are column vectors and row vectors are written as $\bx^\top$ for any vector $\bx$. 
%Notation $\bbone$ represents a vector or a matrix of all ones of proper size. 
%The set of $N\times K$ orthonormal matrices is defined as $\OO^{N \times K}$

%Let $\sigma_i$ be the $i$-th largest square root of eigenvalues of $\bX^\top \bX$, which are called the singular values of $\bX$. $\sigma_1$ is the largest in absolute value among $\sigma_i$.
We use $\lambda_{\min}(\bX)$ and $\lambda_{\max}(\bX)$ to denote its smallest and largest eigenvalues of matrix $\bX$. 
We use the following matrix norms: maximum norm $\norm{\bX}_{\max} \defeq \underset{ij}{\max} \; \lvert x_{ij} \rvert$, $\ell_1$-norm $\norm{\bX}_1 \defeq \underset{j}{\max} \sum_{i} \lvert x_{ij} \rvert$, $\ell_{\infty}$-norm $\norm{\bX}_{\infty} \defeq \underset{i}{\max} \; \sum_{j} \lvert x_{ij} \rvert$, and $\ell_2$-norm $\norm{\bX}_2 \defeq \lambda_{\max}(\bX)$. %$(2,\ell_\infty)$-norm $\norm{\bX}_{(2,\ell_\infty)} \defeq \underset{\norm{\ba}_2=1}{\max} \norm{\bX\ba}_\infty = \underset{i}{\max} \norm{\bx_{i \cdot}}_2$.
%When $\bX$ is a square matrix, we denote by $\Tr \paran{\bX}$, $\lambda_{max} \paran{\bY}$, and $\lambda_{min} \paran{\bY}$ the trace, maximum and minimum singular value of $\bX$, respectively.

%In addition, define projection matrices $\bP_X^\perp = \bI - \bP_X$ and $\bP_X = \bX \paran{\bX^\top\bX}^{-1} \bX^\top$; when $\bX^\top\bX$ is not inversible, then $\paran{\bX^\top\bX}^{-1}$ is replaced by the Moore-Penrose generalized inverse of $\bX^\top\bX$.

%\section{Additional Theoretical Results}

\section{Distinction of Transferred FQI from Transferred SL} \label{sec:diff-TRL-TSL}

In the realm of RL, FQI represents a pivotal algorithmic approach for estimating the optimal action-value function $Q^*$. When considering transferred FQI within the context of transfer learning, it becomes evident that several key distinctions arise compared to algorithms designed for transferred supervised learning (SL).

Firstly, while transferred SL often relies on non-iterative regression-based estimation methods, FQI adopts a fundamentally different approach. Rather than relying on regression, FQI addresses the problem through an iterative fixed-point solving process. This iterative nature not only distinguishes FQI from conventional regression-based methods but also introduces novel challenges and opportunities within the transfer learning paradigm. Consequently, our transferred FQI algorithm and its associated theoretical properties introduce new contributions to the landscape of transfer learning, particularly within the dynamic domain of RL.

Moreover, in transferred SL, the primary focus typically revolves around the target task, without the need to construct better estimators for the source tasks. However, the nature of RL introduces additional complexities, as the response of the optimal action-value function is not directly observable. Consequently, in transferred RL scenarios, the construction of pseudo responses for the source tasks becomes indispensable. Despite the primary emphasis on the target task, our algorithm adeptly handles the construction of pseudo responses for the source tasks as well. This dual focus not only underscores the versatility of our approach but also enables the derivation of estimators for the $Q^*$ function of the source tasks as byproducts. This holistic perspective enhances the scope and applicability of the proposed transferred FQI algorithm, extending its utility beyond traditional transfer learning methodologies.

\section{Select the Number of Basis Functions}\label{sec:supp-p-select}

In this section, we develop a data-adaptive algorithm for selecting the number of basis functions $p$ in Algorithm \ref{algo:select-p-1}. The selection procedure primarily balances the trade-off between the commonality estimation error and the function approximation bias, while accounting for the additional task-difference bias that arises in the knowledge transfer setting. Throughout this section, we denote the task-difference bias term $\Delta_{h_r}=\sqrt{h_r}\left(\frac{\log p}{n_0}\right)^{1/4}\wedge h_r$ as $\Delta_p$ since the task discrepancy $h_r$ defined in \eqref{eq:h-homo} depends on $p$, the dimension of $\bbeta_r^{(k)}$.

To provide insight into Algorithm \ref{algo:select-p-1}, we first examine the fundamental trade-off in selecting $p$. The commonality estimation error $\sqrt{p/n_{\calK}}$ decreases with $p$, while the function approximation bias $p^{-\kappa/d}$ increases with $p$. Balancing these two terms leads to the choice $p^*\asymp n_{\calK}^{d/(2\kappa+d)}$, which achieves the statistical rate $n^{-\kappa/(2\kappa+d)}\vee \Delta_{p^*}$. 

The implementation of this optimal choice faces a key challenge: the smoothness order $\kappa$ is typically unknown in practice. A conservative approach to handle unknown $\kappa$ is to assume $\kappa=1$, which leads to the choice $p'\asymp n_{\calK}^{d/(2+d)}$ and yields an error rate of $n_{\calK}^{-1/(2+d)} \vee \Delta_{p'}$. However, when the true smoothness $\kappa$ is larger than 1 and the task-difference bias is small,  this approach fails to achieve the optimal rate $n_{\calK}^{-\kappa/(2\kappa+d)}$. We therefore propose an alternative method that attains this optimal rate without requiring prior knowledge of $\kappa$.

\begin{algorithm}[!t]
	\caption{Select the number of basis functions $p$ in {\sc TransFQI}}\label{algo:select-p-1}
	\DontPrintSemicolon % Some LaTeX compilers require you to use \dontprintsemicolon instead
	
	\KwIn{ The same input in Algorithm \ref{algo:fitted-q-transfer}, a candidate set $\calP=\{p_g=2^{g}, g=0,1,2,\dots,g_{\max}\}$, and a constant $\widetilde{C}>0$.
	}
	
	\vspace{1ex}
	
	\KwOut{An estimator $\hat{Q}_{\Upsilon}$ and the corresponding greedy policy $\hat\pi_\Upsilon=\pi^{\hat{Q}_{\Upsilon}}$.}
	
	\vspace{1ex}
	
	\For{$g=g_{\max}, g_{\max}-1,\dots, 0$}{
		
		Run Algorithm \ref{algo:fitted-q-transfer} with $p=2^{g}$. Record the estimator obtained as $\widehat{Q}^{(0)}_{\Upsilon, g}$;
		
		\vspace{1ex}
		
		Compute $\varrho_{g, g'}=\int_{\calX}\sum_{a\in \calA} \abs{\widehat{Q}^{(0)}_{\Upsilon,g}(\bx, a)-\widehat{Q}^{(0)}_{\Upsilon, g'}(\bx, a)} \mathrm{d}\bx$ for all $g <g' \leq g_{\max}$.
	}
	Compute $\widetilde{\calG}=\Big\{g: \varrho_{g-1, g} \leq \frac{\widetilde{C}}{1-\gamma} \sqrt{\frac{p_{g}}{n_{\calK}}}\Big\}$.
	
	\If{$\widetilde{\calG}\neq\emptyset$}{
		Let $\widehat{\calG}=\min\Big\{g: \varrho_{g, g'} \leq \frac{\widetilde{C}}{1-\gamma} \sqrt{\frac{p_{g'}}{n_{\calK}}}, \, \forall g \leq g' \leq \widetilde{g}_{\max}\Big\}$, where $\widetilde{g}_{\max}=\max \widetilde{\calG}$.
		
		Select $\widehat{g}=\min \widehat{\calG}$ and  $\hat{Q}_{\Upsilon}=\widehat{Q}_{\Upsilon, \widehat{g}}^{(0)}$.
	}
	\Else{Let $\widetilde{\calG}_0=\Big\{g: \varrho_{g-1, g} \leq \frac{\widetilde{C}}{1-\gamma} n_0^{-2/(d+2)}\Big\}$. 
		
		Select $\widehat{g}=\argmin_{g \in \widetilde{\calG}_0}\varrho_{g-1, g}$ and $\hat{Q}_{\Upsilon}=\widehat{Q}_{\Upsilon, \widehat{g}}^{(0)}$.}
\end{algorithm}

We begin by constructing a candidate set $\calP=\{p_g=2^{g}, g=0,1,2,\dots,g_{\max}\}$ for the number of basis functions, where $g_{\max}:=\lfloor\log_2 n_{\calK}-\frac{2}{2+d}\log_2 n_0\rfloor$. This choice of $g_{\max}$ ensures that $\sqrt{p/n_{\calK}} \lesssim n_0^{-1/(2+d)}$ for all $p \in \calP$, thereby preventing the error rate of our transferred algorithm from exceeding that of non-transfer FQI on the target task.

Inspired by Lepskii's approach \citep{lepskii1991problem},  Algorithm \ref{algo:select-p-1} handles both the unknown smoothness $\kappa$ and the dependence of $\Delta_{p}$ on $p$. The core idea is to identify the smallest $p$ for which the commonality estimation error $\sqrt{p/n_{\calK}}$ dominates the approximation bias $p^{-\kappa/d}$, corresponding to the optimal choice $p^*$. To implement this idea, we first construct a set $\widetilde\calG$ containing candidates from $\calP$ where $\sqrt{p/n_{\calK}}$ is the dominant term. For any $g < g'$, we compute the $\ell_1$ difference $\varrho_{g, g'}$ between estimators obtained with $p=p_g$ and $p=p_{g'}$. This difference satisfies
\[\varrho_{g, g'}=O_{\PP}\bigg[\frac{1}{1-\gamma} \Big(\sqrt{\frac{p_{g'}}{n_{\calK}}}+p_{g}^{-\kappa/d}+(\Delta_{p_g} \vee \Delta_{p_{g'}})\Big)\bigg].\] 
Therefore, the quantity $\varrho_{g-1, g}$ serves as an estimator for the error rate at $p_g$, and $g \in \widetilde\calG$ with high probability when $\sqrt{p_{g}/n_{\calK}}$ dominates both $p_{g-1}^{-\kappa/d}$ and $\Delta_{p_g} \vee \Delta_{p_{g-1}}$. 

To ensure robustness, we further refine $\widetilde\calG$ to $\widehat{\calG}$ by removing isolated points, thus guaranteeing that $\sqrt{p_{g}/n_{\calK}}$ remains the dominant term across an interval. Finally, the number of basis functions is set as $p_{\widehat{g}}$, where $\widehat
g$ is the minimum integer in $\widehat\calG$. As a theoretical guarantee. Theorem \ref{thm:p-select} establishes that this selection procedure achieves an error rate equal to the maximum of the optimal rate $n^{-\kappa/(2\kappa+d)}$ and a task-difference bias term, which parallels the theoretical behavior at $p^*$.

%We provide the theoretical guarantees for the performance of Algorithm  \ref{algo:select-p-1}.  
Define $\widetilde{\calP}:=\left\{p \in \calP: (\Delta_p \vee p^{-\kappa/d}) \leq \sqrt{\frac{p}{n_{\calK}}} \right\}$ as the set where the commonality estimation error is the dominant term. %and $\overline{g} = \arg\!\max_g p_g \in \widetilde\calP$ if $\widetilde\calP \neq \emptyset$. 
\begin{theorem}\label{thm:p-select}
	Assume the assumptions in Theorem \ref{thm:Q-converge-homo} hold, the set $\widetilde{\calG}$ is non-empty, and $p_{\widetilde{g}_{\max}} \in \widetilde{\calP}$. Let $\underline{g}:=\min\{g \mid p_{g'} \in \calP, \, \forall g' = g, g+1, \dots,  \widetilde{g}_{\max}\}$. With sufficiently large constants $\Upsilon$ and $\widetilde{C}$, we have
	\begin{equation}\label{eq:p-select}
		v^{*}-v^{\widehat{\pi}_{\Upsilon}}=O_{\PP} \bigg[\frac{1}{(1-\gamma)^2} \Big(n_{\calK}^{-\frac{\kappa}{2\kappa+d}} \vee \Delta_{p_{\underline{g}-1}}\Big)\bigg].
	\end{equation}
\end{theorem}

Theorem \ref{thm:p-select} ensures that, when $\widetilde{\calG}$ is non-empty, the regret of the estimated policy achieves the maximum of the optimal rate $n_{\calK}^{-\frac{\kappa}{2\kappa+d}}$ and a task-difference bias term $\Delta_{p_{\underline{g}-1}}$. The latter term serves as an upper bound for the task-difference bias $\Delta_{p^*}$ at $p^*=\lceil n_{\calK}^{d/(2\kappa+d)} \rceil$ if $\Delta_p$ is monotone non-decreasing for  $p \geq p^*$. Therefore, our algorithm successfully achieves the optimal statistical rate without requiring knowledge of $\kappa$, at the cost of a modest increase in the task-difference bias. The proof of Theorem \ref{thm:p-select} is provided in Section \ref{sec:proof}.

However, when the task-difference bias is substantial, it is possible that the set $\widetilde\calG$ is empty, indicating that the commonality estimation error never dominates the task-difference bias. To handle this scenario, we construct a larger set $\widetilde{\calG}_0$ that contains all the $p_g \in \calP$ for which $\rho_{g-1, g}$ is of order $O(n_{0}^{-2/(d+2)})$, matching the upper bound for error rate of non-transfer FQI. The non-emptiness of $\widetilde{\calG}_0$ is guaranteed by Theorem \ref{assum:kappa-smooth}, which ensures an error rate of $O_{\PP}\left(n_0^{-\kappa/(2\kappa+d)}\log n_0\right)$ when $g = \lceil \frac{d}{d+2\kappa}\log_2(n_0) \rceil$. We then select $\widehat{g}$ as the element in $\widetilde{\calG}_0$ that minimizes $\varrho_{g-1, g}$. This approach ensures that Algorithm \ref{algo:select-p-1} performs at least as well as non-transfer FQI, with the potential to achieve faster convergence when $\Delta_{p} \lesssim n_{0}^{-2/(d+2)}$ for some $p$.

\section{Technical Proof for Theoretical Results}
\label{sec:proof}
%\subsection{Proof for Technical Lemmas}
\paragraph{Proof for Lemma \ref{lem:diff-Q*}.}

\begin{proof}
	By the Bellman optimality equation, we have
	\[Q^{*(k)}\paran{\bx, a}=r^{(k)}\paran{\bx, a}+\gamma \int_{\calX}\brackets{\underset{a' \in \calA}{\max}  Q^{*(k)}\paran{\bx', a'}}  \rho^{(k)}\paran{\bx'\mid \bx, a } \mathrm{d} \bx',\]
	which implies that
	\begin{align*}
		\delta_{Q}^{(k)}\paran{\bx, a}&=\delta_r^{(k)}\paran{\bx, a}+\gamma \int_{\calX}\brackets{\underset{a' \in \calA}{\max}\;  Q^{*(k)}\paran{\bx', a'}}  \delta_\rho^{(k)}\paran{\bx'\mid \bx, a } \mathrm{d} \bx'\\
		&\quad + \gamma \int_{\calX}\brackets{\underset{a' \in \calA}{\max}\;  Q^{*(k)}\paran{\bx', a'}  - \underset{a' \in \calA}{\max}\;  Q^{*(0)}\paran{\bx', a'}}  \rho^{(0)}\paran{\bx'\mid \bx, a } \mathrm{d} \bx'\\
		&\leq \delta_r^{(k)}\paran{\bx, a}+\frac{\gamma R_{\max}}{1-\gamma}\int_{\calX} \delta_\rho^{(k)}\paran{\bx'\mid \bx, a } \mathrm{d} \bx' \\
		&\quad + \gamma \int _{\calX} \underset{a' \in \calA}{\max}\; \delta_Q^{(k)}(\bx', a') \rho^{(0)}\paran{\bx'\mid \bx, a }\mathrm{d} \bx',
	\end{align*}
	where we use 
	\begin{equation}\label{eq:Vmax}
		\sup_{\bx, a} Q^{*(k)}(\bx, a) = \sup_{\bx, a, \pi}\EE_{\calH \sim \calP_\pi(\calH)}\brackets{ \sum_{t= 0}^T \gamma^t R_{i,t} \bigg| \bX_{i,0} = \bx, A_{i,0} = a} \leq \sum_{t= 0}^T \gamma^t R_{\max} =\frac{R_{\max}}{1-\gamma}.
	\end{equation}
	Taking supreme over $\bx \in \calX$ and $a \in \calA$ on both sides leads to
	\[\sup_{\bx, a}\delta_{Q}^{(k)}\paran{\bx, a} \leq  \sup_{\bx, a}\delta_r^{(k)}\paran{\bx, a}+\frac{\gamma R_{\max}}{1-\gamma}\int_{\calX} \sup_{\bx, a}\delta_\rho^{(k)}\paran{\bx'\mid \bx, a } \mathrm{d} \bx' + \gamma \sup_{\bx, a}\delta_Q^{(k)}(\bx, a),\]
	hence
	\[\sup_{\bx, a}\delta_{Q}^{(k)}\paran{\bx, a} \leq \frac{1}{1-\gamma}\sup_{\bx, a}\delta_r^{(k)}\paran{\bx, a}+\frac{\gamma R_{\max}}{(1-\gamma)^2}\int_{\calX} \sup_{\bx, a}\delta_\rho^{(k)}\paran{\bx'\mid \bx, a } \mathrm{d} \bx'.\]
\end{proof}
\paragraph{Proof of Lemma \ref{lem:Q-kappa-smooth}.}
Note that Assumption \ref{assum:kappa-smooth} assumes that the constants $\kappa$, $c$, and $R_{\max}$ are uniform for all $k$, then the results directly follow from Lemma 1 in \cite{shi2020statistical} and Equation \eqref{eq:Vmax}.

% \paragraph{Proof for Lemma \ref{lem:h-hetero}.}
% By the Bellman optimality equation, we have
% \[Q^{*(k)}\paran{\bx, a}=r^{(k)}\paran{\bx, a}+\gamma \int_{\calX}\brackets{\underset{a' \in \calA}{\max}  Q^{*(k)}\paran{\bx', a'}}  \rho^{(k)}\paran{\bx'\mid \bx, a } \mathrm{d} \bx'.\]
% Define 
% \[\bbeta_Q^{(k)} := \bbeta_r^{(k)} + \gamma \int_{\calX} \brackets{\underset{a' \in \calA}{\max}  Q^{*(k)}\paran{\bx', a'}} \bbeta^{(k)}_{\rho}(\bx') \mathrm{d} \bx'.\]
% By
% \[\sup_{\bx, a, k} \abs{r^{(k)}\paran{\bx, a} - \bxi^{\top}(\bx, a)\bbeta^{(k)}_{r}} \leq Cp^{-\kappa/d},\]
% and
% \[\sup_{\bx, a, \bx', k} \abs{\rho^{(k)}\paran{\bx' \,|\, \bx, a} - \bxi^{\top}(\bx, a)\bbeta^{(k)}_{\rho}(\bx')} \leq Cp^{-\kappa/d},\]
% it holds that
% \[\abs{Q^{*(k)}(\bx, a) - \bxi^{\top}(\bx, a) \bbeta_{Q}^{(k)}} \leq C\paran{1+\frac{\gamma R_{\max}\mu(\calX)}{1-\gamma}}p^{-\kappa / d},\]
% where $\mu(\calX)$ denotes the Lebesgue measure of the compact space $\calX$. Moreover,
% \[\norm{\bbeta_{Q}^{(k)}-\bbeta_{Q}^{(0)}}_1\]

\smallskip\noindent
\textbf{Proof Sketch for Theorem \ref{thm:Q-converge-homo}.}  
We first provide a proof sketch for Theorem \ref{thm:Q-converge-homo}, followed by the complete proof.
To establish the error rate in \eqref{eqn:rate-homo}, we begin with an error decomposition:
\begin{equation}\label{eq:Emub-homo-sketch}
	\begin{aligned}
		\abs{Q^{*}(\bx, a)-\hat Q^{(0)}_{{\Upsilon}}(\bx, a)}
		&\leq \sum_{\tau=0}^{\Upsilon-2}\gamma^{\Upsilon-1-\tau}\sup_{\omega_1,\dots,\omega_{\Upsilon-1-\tau}}P^{\omega_{\Upsilon-1-\tau}}\cdots P^{\omega_1}\abs{\calT \hat Q^{(0)}_{\tau}(\bx, a)-\hat Q^{(0)}_{\tau+1}(\bx, a)} \\
		&\quad 
		+  \abs{\calT \hat Q^{(0)}_{\Upsilon-1}(\bx, a)-\hat Q^{(0)}_{\Upsilon}(\bx, a)} +\frac{2\gamma^{\Upsilon}R_{\max}}{1-\gamma},
	\end{aligned}
\end{equation}
where $Q^*$ denotes the optimal action-value function, $\calT$ is the Bellman optimality operator, and $ P^{\omega}f(\bx, a):=\EE\left[f(\bx^{\prime}, a^{\prime}) \,|\, \bx^{\prime} \sim P(\cdot \,|\, \bx, a), a^{\prime} \sim \omega(\cdot \,|\, \bx^{\prime}) \right]$.  This decomposition relates the estimation error to the propagation of errors $\big| \calT \hat Q^{(0)}_{\tau}-\hat Q^{(0)}_{\tau+1}\big|$ across iterations.

Under Assumption \ref{assum:distribution}, we have
$
Y^{(k), \tau}_{i,t} = \calT\hat Q_{\tau-1}^{(k)}\big(\bX^{(k)}_{i,t}, A^{(k)}_{i,t}\big) + e^{(k), \tau}_{i,t} 
$,
where $e^{(k), \tau}_{i,t}$ is sub-Gaussian. Assumption \ref{assum:kappa-smooth} further guarantees that, for all $k$ and $\tau$,  there exists a set of coefficients $\{\bbeta_{\tau}^{(k)}\}$ such that  $\sup\big|\calT\hat Q_{\tau-1}^{(k)}\paran{\bx, a}-\bxi^{\top}(\bx, a)\bbeta_{\tau}^{(k)}\big|\lesssim p^{-\kappa/d}$ and $\sup \big\|\bbeta_{\tau}^{(k)}-\bbeta_{\tau}^{(0)}\big\|_1\leq h_r$. This allows us to treat $Y_{i, t}^{(k), \tau}$ as the response variable and $\bxi\big(\bX_{i, t}^{(k)}, A_{i, t}^{(k)}\big)$ as predictors, computing the regression coefficient $\widehat\bbeta_{\tau}^{(k)}$ as a consistent estimator for $\bbeta_{\tau}^{(k)}$.
This analysis proceeds in two steps: bounding the error of the aggregated estimator $\widehat{\bw}_{\tau}$ (Lemma \ref{lem:step-1-error}) and the error of the bias correction estimator $\widehat{\bdelta}_{\tau}^{(k)}$ (Lemma \ref{lem:step-2-error}). 

With the bound on $\big\|\bbeta_{\tau}^{(0)}-\widehat{\bbeta}_{\tau}^{(0)}\big\|_2$ established, we can bound the error propagation term $\big|\calT \hat Q^{(0)}_{\tau}-\hat Q^{(0)}_{\tau+1}\big| \approx \bxi^{\top}\big(\bX_{i, t}^{(k)}, A_{i, t}^{(k)}\big)\big(\bbeta_{\tau}^{(0)}-\widehat{\bbeta}_{\tau}^{(0)}\big)$ and further establish the error bound on $\big|Q^{*}-\hat Q^{(0)}_{{\Upsilon}}\big|$ through \eqref{eq:Emub-homo-sketch}. The final bound on $v^{\pi^*}-v^{\widehat{\pi}_{\Upsilon}}$ follows from the bound on $\big|Q^{*}-\hat Q^{(0)}_{{\Upsilon}}\big|$ via Lemma 13 in \cite{chen2019information}.

\paragraph{Proof of Theorem \ref{thm:Q-converge-homo}.}

\begin{proof}
	We first restate the assumptions on the distribution of the data in task $k$, i.e., $\calS^{(k)}=\left\{\big(\bX_{i, t}^{(k)}, A_{i, t}^{(k)}, R_{i, t}^{(k)}, \bX_{i, t+1}^{(k)}\big)\right\}$ for $i \in [I^{(k)}]$, $t\in \{0\} \cup [T]$, under the homogeneous transition setting.
	
	The initial state $\bX_{1,0}^{(k)}$ has a distribution density $\nu$ almost everywhere. For any $0\leq t \leq T$, the probability distribution of the action $A^{(k)}_{1,t}$ conditional on $\bX^{(k)}_{1,t}=\bx$ is given by an underlying behavior policy $b(\cdot \,|\, \bx)$. Given $\bX^{(k)}_{1,t}=\bx, A^{(k)}_{1,t}=a$, the distribution of the next state $\bX_{1,t+1}^{(k)}$ is determined by a transition kernel $P\big(\cdot \,|\, \bx, a\big)$.  Suppose that the density $\rho(\bx' \,|\, \bx, a)=\frac{\mathrm{d}}{\mathrm{d} \bx'}P\big(\bx' \,|\, \bx, a\big)$ exists and is continuous for almost every $(\bx, a)$. Also, a reward $R_{i,t}^{(k)}=r^{(k)}\left(\bX^{(k)}_{i,t}, A^{(k)}_{i,t}\right) + \eta^{(k)}_{i,t}$ is observed for any $t$. The tuple set 
	\[\Big\{\big(\bX^{(k)}_{1,t}, A^{(k)}_{1,t}, R_{1,t}^{(k)}, \bX^{(k)}_{1,t+1}\big)\Big\}_{t=0}^{T-1}\] 
	forms a trajectory, and there are multiple i.i.d. trajectories $\Big\{\big(\bX^{(k)}_{i,t}, A^{(k)}_{i,t}, R_{i,t}^{(k)}, \bX^{(k)}_{i,t+1}\big)\Big\}$ in one task. We further split $\calS^{(k)}$ into $\tau$ subsets, $\left\{\calS_{\tau}^{(k)}\right\}_{\tau \in [\Upsilon]}$, each with sample size $n_{k}$, and use subset $\calS_{\tau}^{(k)}$ in iteration $\tau$, which ensures that the data used in different iterations are independent. %For simplicity, we put the tuples of all of the trajectories together and let $n_k=I^{(k)}T$ be the sample size of task $k$. 
	Assumption \ref{assum:distribution} ensures that, for any $i$, $\left\{\left(\bX^{(k)}_{i,t}, A^{(k)}_{i,t} \right)\right\}_{t =0}^{T-1}$ is a Markov chain on $\calX \times \calA$ with the transition kernel $P(\bx^{\prime}\,|\, \bx, a)b(a^{\prime}\,|\,\bx^{\prime})$, and  $\left\{\bX^{(k)}_{i,t}\right\}_{t =0}^{T-1}$ is also a Markov chain with the transition kernel $\sum_{a \in \calA} b(a \,|\,\bx)P(\bx^{\prime}\,|\, \bx, a)$. The latter chain has an invariant distribution $\mu(\bx)$, which implies that the former chain has an invariant distribution $\mu(\bx) b(a \,|\, \bx)$. %Furthermore, since $\nu^{(0)}=\mu^{(0)}$, the target task follows the invariant distribution, i.e.,  $\left(\bX^{(0)}_{i,t}, A^{(0)}_{i,t} \right) \sim \mu^{(0)}(\bx) b^{(0)}(a \mid \bx)$ for all $i, t$. %Also notice that the continuity of %$\calX$ is bounded in $L_2$ norm,$\mu^{(0)}(\cdot)$ implies that it is bounded away from 0 and $\infty$ on $\supp(\mu^{(k)}(\bx))$, since $\supp(\mu^{(k)}(\bx))$ is bounded. %, and $\mu^{(0)}(\cdot)$ is uniformly bounded away from zero and infinity. 
	%Moreover, $\inf_{a, \bx} b^{(0)}(a \,|\, \bx) \geq \underline{b} > 0$. 
	
	Define the Bellman optimality operator $\calT^{(k)}$  on the $k$-th task as
	\begin{equation}
		\calT^{(k)} Q \paran{\bx, a} = r^{(k)}\paran{\bx, a} + \gamma\cdot\EE\brackets{ \underset{a'\in\calA}{\max}\,Q\paran{\bx', a'} \,\big|\, \bx'\sim P\paran{\cdot\,|\,\bx, a} },
	\end{equation}
	for any function $Q(x, a): \calX \times \calA \to \RR$.
	%We have $\calT^{(k)} Q^{*, (k)} = Q^{*, (k)}$.
	For any function $f(\bx, a)$ defined on $\calX \times \calA$ and a policy function $\omega(\cdot \mid \bx)$, define the operator $P^{\omega}$ by
	\[P^{\omega}f(\bx, a):=\EE\left[f(\bx^{\prime}, a^{\prime}) \,|\, \bx^{\prime} \sim P(\cdot \,|\, \bx, a), a^{\prime} \sim \omega(\cdot \,|\, \bx^{\prime}) \right].\]
	
	%\begin{proof}
	
	Let $Q^{*, (k)}$ be the optimal action-value function on the $k$-th dataset, and for simplicity, let $Q^*=Q^{*, (0)}$ be the target $Q^*$ function. The iterative algorithm gives a series of estimators $\hat Q^{(0)}_\tau$ for $Q^*$ and finally an estimated optimal policy $\hat Q^{(0)}_{\Upsilon}$. By Lemma C.2 in \cite{fan2020theoretical}, %By the proof of Theorem 6.1 in \cite{fan2020theoretical}, % Equation (C.27) and the equation under (C.27) 
	it holds that
	\begin{equation}\label{eq:Emub-homo-0}
		\begin{aligned}
			&\quad\abs{Q^{*}-\hat Q^{(0)}_{{\Upsilon}}}\\
			&\leq \sum_{\tau=0}^{\Upsilon-2}\gamma^{\Upsilon-1-\tau}\sup_{\omega_1,\dots,\omega_{\Upsilon-1-\tau}}P^{\omega_{\Upsilon-1-\tau}}P^{\omega_{\Upsilon-\tau-2}}\cdots P^{\omega_1}\abs{\calT^{(0)} \hat Q^{(0)}_{\tau}-\hat Q^{(0)}_{\tau+1}} \\
			&\quad 
			+  \abs{\calT^{(0)} \hat Q^{(0)}_{\Upsilon-1}-\hat Q^{(0)}_{\Upsilon}} +\gamma^{\Upsilon}\sup_{\omega_1,\dots,\omega_{\Upsilon}}P^{\omega_{\Upsilon}}P^{\omega_{\Upsilon-1}}\cdots P^{\omega_1}\abs{Q^*-\hat Q^{(0)}_0},
		\end{aligned}
	\end{equation}
	%\begin{equation}\label{eq:Emub-homo}
	%	\begin{aligned}
		%		&\quad\EE_{\mu, b}\abs{Q^{*}-Q^{\pi_{\Upsilon}}}\\
		%		&\lesssim \sum_{\tau=0}^{\Upsilon-1}\sum_{j=0}^{\infty}\gamma^{\Upsilon-\tau+j}\sup_{\omega_1,\dots,\omega_{\Upsilon-\tau+j}}\EE_{\mu, b}\left[(P^{\omega_{\Upsilon-\tau+j}}P^{\omega_{\Upsilon-\tau+j-1}}\cdots P^{\omega_1})\abs{\calT^{(0)} \hat Q^{(0)}_{\tau}-\hat Q^{(0)}_{\tau+1}}\right] \\
		%		&\quad +\frac{\gamma^{\Upsilon+1}}{(1-\gamma)^2}R_{\max},
		%	\end{aligned}
	%\end{equation}
	where the supreme is taken over all possible policies $\omega_1,\dots,\omega_{\Upsilon}$. Let $V_{\max}=R_{\max}/(1-\gamma)$, which is an upper bound for the optimal $Q^*$ function since \[Q^*(\bx, a)=\sum_{t=0}^{\infty}\gamma^{t}\EE[r^{(0)}(\bX_{i, t}, A_{i, t}) \mid \bX_{i, 0}=x, A_{i, t}=a] \leq \sum_{t=0}^{\infty}\gamma^{t}R_{\max} = V_{\max}.\] The choice of initial estimators ensures that  $\max_{\bx, a} Q^{(0)}_{0}(\bx, a) \leq V_{\max}$, which implies
	
	\begin{equation}\label{eq:Emub-homo}
		\begin{aligned}
			&\quad\abs{Q^{*}-\hat Q^{(0)}_{{\Upsilon}}}\\
			&\leq \sum_{\tau=0}^{\Upsilon-2}\gamma^{\Upsilon-1-\tau}\sup_{\omega_1,\dots,\omega_{\Upsilon-1-\tau}}P^{\omega_{\Upsilon-1-\tau}}P^{\omega_{\Upsilon-\tau-2}}\cdots P^{\omega_1}\abs{\calT^{(0)} \hat Q^{(0)}_{\tau}-\hat Q^{(0)}_{\tau+1}} \\
			&\quad 
			+  \abs{\calT^{(0)} \hat Q^{(0)}_{\Upsilon-1}-\hat Q^{(0)}_{\Upsilon}} +\frac{2\gamma^{\Upsilon}R_{\max}}{1-\gamma}.
		\end{aligned}
	\end{equation}
	
	%, and $\EE_{\mu^{(0)}, b^{(0)}}$ denotes the expectation with respect to the invariant distribution $\mu(\bx) b(a \,|\, \bx)$. 
	
	Therefore, we first focus on bounding $\abs{\calT^{(0)} \hat Q^{(0)}_{\tau-1} - \hat Q^{(0)}_\tau}(\bx, a)$ for a fixed $ \tau  \in [\Upsilon]$, where $\hat Q^{(k)}_{\tau}(\bx, a)=\bxi(\bx, a)^{\top}\hbbeta^{(k)}_{\tau}$. We first assume that $\norm{\hat Q_{\tau-1}^{(0)}}_{\infty} \lesssim V_{\max}$ and $\sup_{k}\norm{\hat\bbeta_{\tau-1}^{(k)}-\hat\bbeta_{\tau-1}^{(0)}}_{1} \lesssim h$, which are satisfied for $\tau=1$ by assumption.
	In Algorithm \ref{algo:fitted-q-transfer}, for each $k$,
	we compute 
	\[Y^{(k), \tau}_{i,t} = R^{(k)}_{i,t} + \gamma\cdot \underset{a\in\calA}{\max}\, \hat Q^{(k)}_{\tau-1}\paran{\bX^{(k)}_{i, t+1}, a}.\]
	By definition, we have 
	\[\E{Y^{(k), \tau}_{i,t} \,\big|\, \bX^{(k)}_{i,t}, A^{(k)}_{i,t}} = \paran{\calT^{(k)} \hat Q^{(k)}_{\tau-1}}\paran{\bX^{(k)}_{i,t}, A^{(k)}_{i,t}},\] for any $\paran{\bX^{(k)}_{i,t}, A^{(k)}_{i,t}} \in \calX \times \calA$.
	Thus, $\calT^{(k)} \hat Q^{(k)}_{\tau-1}$ can be viewed as the underlying truth of the regression problem defined in \eqref{eqn:fqi-update}, where the covariates and responses are $\paran{\bX^{(k)}_{i,t}, A^{(k)}_{i,t}}$ and $Y^{(k), \tau}_{i,t}$, respectively. Therefore, we rewrite $Y^{(k), \tau}_{i,t}$ as
	\begin{equation}\label{eq:Y=T+E}
		Y^{(k), \tau}_{i,t} = \calT^{(k)}\hat Q_{\tau-1}^{(k)}\paran{\bX^{(k)}_{i,t}, A^{(k)}_{i,t}} + e^{(k), \tau}_{i,t}, 
	\end{equation}
	where $\calT^{(k)}\hat Q_{\tau-1}^{(k)}:\calX\times\calA\mapsto\RR$ is an unknown regression function to be estimated. Furthermore, the reward \[R^{(k)}_{i,t}=r^{(k)}\paran{\bX_{i,t}^{(k)}, A_{i,t}^{(k)}}+\eta^{(k)}_{i,t},\] 
	and 
	\[\paran{\calT^{(k)} \hat Q^{(k)}_{\tau-1}}\paran{\bX^{(k)}_{i,t}, A_{i,t}^{(k)}}=r^{(k)}\paran{\bX_{i,t}^{(k)}, A_{i,t}^{(k)}}+\gamma \E{\underset{a \in \calA}{\max} \hat Q^{(k)}_{\tau-1}\paran{\bx', a} \big| \bx' \sim P\paran{\cdot \mid \bX_{i,t}^{(k)}, A_{i,t}^{(k)}}  }.\]
	Hence, the regression noise term $e^{(k), \tau}_{i,t}$ is zero-mean sub-Gaussian, since
	\[
	e^{(k), \tau}_{i,t} = \eta^{(k)}_{i, t} + \gamma \paran{1-\mathbb E}\left[ \underset{a \in \calA}{\max} \hat Q^{(k)}_{\tau-1}\paran{\bx', a} \big| \bx' \sim P\paran{\cdot \mid \bX_{i,t}^{(k)}, A_{i,t}^{(k)}}  \right],\]
	where $\eta^{(k)}_{i, t}$ is zero-mean sub-Gaussian, and the second term is bounded. 
	
	%Let $\calF_p$ be a finite dictionary of functions $\phi_j:\calX\times\calA\mapsto\RR$, $j\in[p]$. 
	%By Lemma \ref{thm:bellman-opt-close} and the Assumption that $\calF_p$ is a collection of basis functions (e.g., wavelets, splines with fixed knots, step functions), the unknown function $f$ can be well approximated by a member of the span of $\calF_p$. 
	%In this paper, we have in mind the situation where $p >> n$, and $f$ can be estimated reasonably only because it can be approximated by a linear combination of a small number of members of $\calF_p$, or, in other words, it has a sparse approximation in the span of $\calF_p$. \wenbo{How to state this as an assumption?}
	In each step $\tau$, we use $\hat Q^{(k)}_{\tau}(\bx, a)=\bxi^{\top}(\bx, a)\hbbeta_\tau^{(k)}$ to estimate $\paran{\calT^{(k)} \hat Q^{(k)}_{\tau-1}}\paran{\bx, a}$, where 
	\[\bxi(\bx, a):=\left[\bphi^{\top}(\bx)\II(a=1),\bphi^{\top}(\bx)\II(a=2),\dots,\bphi^{\top}(\bx)\II(a=m)\right]^{\top},\]
	and $\bphi(\cdot)=\paran{\phi_1(\cdot),\dots, \phi_p(\cdot)}^{\top}$ is a set of sieve basis functions.  
	By definition,
	\[\paran{\calT^{(k)} \hat Q^{(k)}_{\tau-1}}\paran{\bx, a}=r^{(k)}\paran{\bx, a}+\gamma \int_{\calX}\brackets{\underset{a' \in \calA}{\max} \hat Q^{(k)}_{\tau-1}\paran{\bx', a'}}  \rho\paran{\bx'\mid \bx, a } \mathrm{d} \bx'.\]
	The estimation is guaranteed by the following property: 
	For any function $f(\bx, a)$ that satisfies $f(\cdot, a) \in \Lambda(\kappa, c)$ for all $a\in\calA$,  there exists a set of vectors $\braces{\bbeta_a\in\RR^p}_{a \in \calA}$ that 
	\begin{equation*}
		\underset{\bx\in\calX, a\in \calA}{\sup} \abs{  f\paran{\bx, a} - \bbeta_a^\top\bphi(\bx)} 
		\le C p^{-\kappa/d},
	\end{equation*}
	for some positive constant $C$. Let $\bbeta=\paran{\bbeta_1^{\top},\dots,\bbeta_m^{\top}}^{\top}$, and then we have
	\[	\underset{\bx\in\calX, a\in \calA}{\sup} \abs{  f\paran{\bx, a} - \bbeta^\top\bxi(\bx, a)} 
	\le C p^{-\kappa/d}.\]
	By Assumption \ref{assum:kappa-smooth} and the definition of $h_r$, there exist $\left\{\bbeta^{(k)}_{r}\right\}_{k \in \calK}$, $\left\{\bbeta_{\rho}(\bx')\right\}_{\bx' \in \calX}$ 
	such that 
	\[\sup_{\bx, a, k} \abs{r^{(k)}\paran{\bx, a} - \bxi^{\top}(\bx, a)\bbeta^{(k)}_{r}} \leq Cp^{-\kappa/d},\]
	\[\sup_{\bx, a, \bx'} \abs{\rho\paran{\bx' \,|\, \bx, a} - \bxi^{\top}(\bx, a)\bbeta_{\rho}(\bx')} \leq Cp^{-\kappa/d},\]
	and 
	\[%\int_{\calX}\norm{\bbeta^{(k)}_{\rho}(\bx')-\bbeta^{(0)}_{\rho}(\bx')}_1 \mathrm{d}\bx' \leq h, \quad  
	\norm{\bbeta^{(k)}_{r}-\bbeta^{(0)}_{r}}_1 \leq h, \]%\[\quad \norm{\bbeta_{\rho}^{(0)}(\bx')}_1 \leq C_{\bbeta}.\]
	where we replace $h_r$ with $h$ for simplicity.
	Define 
	\[\bbeta_{\tau}^{(k)}:=\bbeta^{(k)}_{r}+\gamma \int \brackets{\max_{a' \in \calA}\hat Q^{(k)}_{\tau-1}\paran{\bx', a'}}\bbeta_{\rho}(\bx')   \mathrm{d} \bx'.\]
	Since $\max\limits_{a' \in \calA}\hat Q^{(k)}_{\tau-1}\paran{\bx', a'} \lesssim V_{\max}$, it follows that 
	\begin{equation}
		\label{eq:bias_f-beta-homo}
		\sup_{\bx, a, k}\abs{\paran{\calT^{(k)} \hat Q^{(k)}_{\tau-1}}\paran{\bx, a}-\bxi^{\top}(\bx, a)\bbeta_{\tau}^{(k)}} \leq C^{\prime}p^{-\kappa/d},
	\end{equation}
	for some absolute constant $C'$. 
	
	Moreover, note that
	\begin{equation}
		\label{eq:betak-beta0-homo}
		\begin{aligned}
			&\quad \bbeta_{\tau}^{(k)}-\bbeta_{\tau}^{(0)}\\
			=&\quad \bbeta_r^{(k)}-\bbeta_r^{(0)}+ \gamma \int_{\cX} \brackets{\max_{a' \in \calA}\hat Q^{(k)}_{\tau-1}\paran{\bx', a'}- \max_{a' \in \calA}\hat Q^{(0)}_{\tau-1}\paran{\bx', a'}}\bbeta_{\rho}(\bx')   \mathrm{d} \bx'.
		\end{aligned} 
	\end{equation}
	%The $\ell_1$-norm of the first two terms on the RHS of \eqref{eq:betak-beta0-homo} can be bounded by a constant times $h$. For the third term, we first have 
	Note that
	\begin{align*}
		&\quad \int_{\calX}\abs{\underset{a' \in \calA}{\max} \hat Q^{(k)}_{\tau-1}\paran{\bx', a'}-\underset{a' \in \calA}{\max} \hat Q^{(0)}_{\tau-1}\paran{\bx', a'} } \mathrm{d} \bx'\\
		&\leq \int_{\calX}  \abs{\max_{a'} \bxi^{\top}(\bx', a')\paran{\hat\bbeta_{\tau-1}^{(k)}-\hat\bbeta_{\tau-1}^{(0)}}} \mathrm{d} \bx'\\
		&\leq \sqrt{\paran{\hat\bbeta_{\tau-1}^{(k)}-\hat\bbeta_{\tau-1}^{(0)}}^{\top}\int_{\mathcal{X}}\brackets{\bPhi(\bx')\bPhi^{\top}(\bx')}\mathrm{d}\bx'\paran{\hat\bbeta_{\tau-1}^{(k)}-\hat\bbeta_{\tau-1}^{(0)}}}\\
		& \lesssim \norm{\hat\bbeta_{\tau-1}^{(k)}-\hat\bbeta_{\tau-1}^{(0)}}_2,
	\end{align*}
	where $\bPhi(\bx):=\left[\bphi^{\top}(\bx),\bphi^{\top}(\bx),\dots,\bphi^{\top}(\bx)\right]^{\top}$ is the concatenation of $|\calA|$ copies of the basis functions $\bphi(\bx)$. The last inequality follows from Lemma 2 in \cite{shi2020statistical}, which shows that
	$\lam_{\max}\brackets{\int_{\cal X} \bphi(\bx)\bphi^{\top}(\bx) \mathrm{d} x} < \infty$.
	%and $\rho^{(0)}$ is uniformly bounded away from infinity.
	%The assumption that $\mu^{(0)}$ is continuous and bounded away from 0 implies that $\mathrm{supp}(\mu^{(0)})$ is compact. Since $\mu^{(0)}$ is the distribution of all $\bX_{i, t}^{(0)}$, we can assume without loss of generality that $\calX=\mathrm{supp}(\mu^{(0)})$. Then $\rho^{(0)}$ is also uniformly bounded away from zero. Adding that $\norm{\bbeta_P^{(0)}(\bx')}_1$ is bounded, 
	%Then we obtain that the $\ell_1$-norm of the third term of \eqref{eq:betak-beta0-homo} is upper bounded by a constant times $\gamma\norm{\hat\bbeta_{\tau-1}^{(k)}-\hat\bbeta_{\tau-1}^{(0)}}_2$.
	Therefore, 
	\[\norm{\bbeta_{\tau}^{(k)}-\bbeta_{\tau}^{(0)}}_1 \lesssim h + \gamma \norm{\hat\bbeta_{\tau-1}^{(k)}-\hat\bbeta_{\tau-1}^{(0)}}_2 \lesssim h, \quad\forall k.\] 
	%Note that 
	%\begin{equation}
	%	\begin{aligned}
		%		&\quad \left(\bbeta^{(k)}-\bbeta^{(0)}\right)^{\top}\E{\bxi(\bx, a)\bxi^{\top}(\bx, a)}	\left(\bbeta^{(k)}-\bbeta^{(0)}\right)\\&=\E{\paran{f_{\bbeta^{(k)}}(\bx, a)-f_{\bbeta^{(0)}}(\bx, a)}^2} \\
		%		&\lesssim p^{-2\kappa/d} + \E{\left(f^{(k)}(\bx, a)-f^{(0)}(\bx, a)\right)^2}\\
		%		&\lesssim p^{-2\kappa/d} + 
		%	\end{aligned}
	%\end{equation}
	%Recall that
	%\[f^{(k)}(\bx, a):=\paran{\calT^{(k)} \hat Q^{(k)}_{\tau-1}}\paran{\bx, a}=r^{(k)}(\bx, a)+\gamma\int_{\cal X}\left[\max_{a' \in \calA} \bxi^{\top}(\bx', a') \hat\bbeta_{\tau-1}^{(k)}\right] P^{(k)}(\bx' \,|\, \bx, a)\mathrm{d} \bx'.\]
	%\begin{align*}
	%	&\quad \E{\left(f^{(k)}(\bx, a)-f^{(0)}(\bx, a)\right)^2} \\
	%	&\lesssim \E{\paran{r^{(k)}(\bx, a)-r^{(0)}(\bx, a)}^2} + \gamma\E{\int \max_{a'} \abs{\bxi^{\top}(\bx', a')\left(\hat\bbeta_{\tau-1}^{(k)}-\hat\bbeta_{\tau-1}^{(0)}\right)}P^{(k)}(\bx' \,|\, \bx, a)\mathrm{d} \bx'}\\
	%	&\quad + 
	%\end{align*}
	Without loss of generality, we write that $\bbeta_{\tau}^{(k)}=\bbeta_{\tau}+\bdelta_{\tau}^{(k)}$ with $\norm{\bdelta_{\tau}^{(k)}}_1 \lesssim h$ for $k=0,1,\dots, K$. %, where 
	%\[\widetilde h :=  h \vee \gamma \sup_k\norm{\hat\bbeta_{\tau-1}^{(k)}-\hat\bbeta_{\tau-1}^{(0)}}_2.\] 
	%Let $\hat f_{\bbeta} = \hat \bbeta^{\top}\left[\bphi(\bx)I(a=0),\bphi(\bx)I(a=1),\dots,\bphi(\bx)I(a=m)\right]$ be the estimator given by Algorithm \ref{algo:fitted-q-transfer}, which is refreshed as follows.
	
	Till now, we have already established the formal definition and properties of the parameter of interest $\bbeta_{\tau}^{(k)}$.
	In the sequel, we will provide an upper bound for the $\ell_2$ error $\norm{\hat\bbeta_{\tau}^{(k)}-\bbeta_{\tau}^{(k)}}_2$, which is closely related to $\abs{\calT^{(0)} \hat Q^{(0)}_{\tau-1} - \hat Q^{(0)}_\tau}$. We first simplify some notations for a clear presentation.
	
	Let $f_{\tau}^{(k)}=\calT^{(k)}\hat Q_{\tau-1}^{(k)}$.
	Define the $k$-th sample matrix $\bZ_\tau^{(k)}\in \RR^{n_k \times mp}$ such that the rows of $\bZ_{\tau}^{(k)}$ are $ \bxi^{\top}\big(\bX_{i,t}^{(k)}, A_{i,t}^{(k)}\big)$, for $(i, t) \in \calS_{\tau}^{(k)}$. Likewise, define $\by_{\tau}^{(k)}$ to be the concatenation of $Y_{i, t}^{(k),\tau}$ for all $(i, t) \in \calS_{\tau}^{(k)}$, $[\bff_{\tau}^{(k)}$ to be the concatenation of $f_{\tau}^{(k)}\big(\bX_{i,t}^{(k)}, \bA_{i,t}^{(k)}\big)$ for all $(i, t) \in \calS_{\tau}^{(k)}$, and $\be_{\tau}^{(k)}$ to be the concatenation of $e_{i, t}^{(k), \tau}$ for all $(i, t) \in \calS_{\tau}^{(k)}$.
	%\[\by_{\tau}^{(k)}: = \brackets{Y_{i_1,0}^{(k), \tau}, \dots, Y_{i_1,T-1}^{(k), \tau}, Y_{i_2,0}^{(k), \tau}, \dots, Y^{(k), \tau}_{i_{n_k/T},T-1}}^\top, (i_1,\dots,i_{n_k/T} \in \calS_{\tau}^{(k)})\] 
	%\[\bff_{\tau}^{(k)}:=\brackets{f_{\tau}^{(k)}\big(\bX_{i_1,0}^{(k)}, \bA_{i_1,0}^{(k)}\big), \dots, f_{\tau}^{(k)}\big(\bX_{i_{n_k/T},T-1}^{(k)}, \bA_{i_{n_k/T},T-1}^{(k)}\big)}^{\top},\] 
	%and $\be_{\tau}^{(k)} := \brackets{e_{i_1, 0}^{(k), \tau}, \dots, e_{i_{n_k/T}, T-1}^{(k), \tau}}^\top$.
	Using these notations and \eqref{eq:Y=T+E}, we have 
	\[\by_{\tau}^{(k)}=\bZ_{\tau}^{(k)}\bbeta_{\tau}^{(k)}+\brackets{\bff_{\tau}^{(k)}-\bZ_{\tau}^{(k)}\bbeta_{\tau}^{(k)}}+\be_{\tau}^{(k)},\]
	where $\be_{\tau}^{(k)}$ is sub-Gaussian, and by \eqref{eq:bias_f-beta-homo}, 
	\[\norm{\bff_{\tau}^{(k)}-\bZ_{\tau}^{(k)}\bbeta_{\tau}^{(k)}}_{\infty} \leq C^{\prime}p^{-\kappa/d}.\]
	Moreover, define $n_{\calK}=\sum_{0\le k\le K}n_k$, $\hat\bSigma_{\tau}^{(k)} := \paran{\bZ_{\tau}^{(k)}}^\top\bZ_{\tau}^{(k)}/n_k$,
	$\hat\bSigma_{\tau} := \sum_{0\le k \le K} \alpha_k \hat\bSigma_{\tau}^{(k)}$ with $\alpha_k = n_k / n_{\calK}$. By Assumption \ref{assum:distribution},
	\[\bSigma = \frac{1}{T}\EE \left[\sum_{t=0}^{T-1}\bxi\big(\bX_{i,t}^{(k)}, A_{i,t}^{(k)}\big)\bxi^{\top}\big(\bX_{i,t}^{(k)}, A_{i,t}^{(k)}\big)\right],\]
	and thus 
	\[\bSigma=\E{\hat\bSigma_{\tau}^{(k)}}=\E{\hat\bSigma_{\tau}}.\]%, and 
	%\[C_{\Sigma}:=1+\max_{k}\norm{\overline\bSigma^{-1}\paran{\overline\bSigma^{(k)}-\overline\bSigma}}_1.\]
	
	%By definition, 
	%\[ \overline\bSigma = \frac{1}{T}\EE \left[\sum_{t=0}^{T-1}\bxi\big(\bX_{i,t}^{(k)}, A_{i,t}^{(k)}\big)\bxi^{\top}\big(\bX_{i,t}^{(k)}, A_{i,t}^{(k)}\big)\right].\]% = \int_{\calX} \sum_{a \in \calA} \bxi\big(\bx, a\big)\bxi^{\top}\big(\bx, a\big)\mu^{(k)}(\bx)b^{(k)}(a \mid \bx) \mathrm{d} \bx.\]
	By Assumption \ref{assum:distribution}, there exists a constant $c_{\Sigma} \geq 1$ such that %$c_{\Sigma}^{-1}<\lambda_{\min}\paran{ \overline\bSigma^{(k)}}<\lambda_{\max}\paran{\overline\bSigma^{(k)}} < c_{\Sigma}$ for all $k$ and thus,
	$c_{\Sigma}^{-1}<\lambda_{\min}\paran{\bSigma}<\lambda_{\max}\paran{\bSigma} < c_{\Sigma}$. By the proof of Lemma 4 in \cite{shi2020statistical}, there exists a constant $c'_{\Sigma} \geq 1$ such that $c_{\Sigma}^{'-1}<\lambda_{\min}\paran{\hat \bSigma^{(k)}}<\lambda_{\max}\paran{\hat \bSigma^{(k)}} < c'_{\Sigma}$ for all $k \in \calK$ and $c_{\Sigma}^{'-1}<\lambda_{\min}\paran{\hat \bSigma}<\lambda_{\max}\paran{\hat \bSigma} < c'_{\Sigma}$, with probability approaching one. Without less of generality, we suppose $c_{\Sigma}=c'_{\Sigma}$. Therefore, we have that $\norm{\hat \bSigma_{\tau}^{-1}}_2=O_{\PP}(1)$ and $\sup_{k} \norm{\paran{\hat \bSigma_{\tau}^{(k)}}^{-1}}_2=O_{\PP}(1)$. 
	
	We then establish the convergence rate for aggregated estimator in Step \rom{1} of Algorithm \ref{algo:fitted-q-transfer}. Let
	\begin{equation}
		\label{eq:def:w}
		\bw_{\tau}: = \bbeta_{\tau} + \bdelta_{\tau}, 
	\end{equation}
	where 
	\begin{equation}
		\label{eq:def:bdelta-homo}
		\bdelta_{\tau}: = \sum_{0\le k\le K} \alpha_k\bdelta_{\tau}^{(k)}.  
	\end{equation}
	We will show that
	\begin{lemma}\label{lem:step-1-error}
		Under the assumptions of Theorem \ref{thm:Q-converge-homo}, 
		\begin{equation}	\norm{\bw_{\tau}-\hat\bw_{\tau}}_2 =O_{\PP}\paran{\sqrt{\frac{p}{n_{\calK}}}+%h\sqrt{\frac{p\log^2(n_{\calK})}{n_{\calK}}}+
				p^{-\kappa/d}}.
			\label{eq:step-1-error}
		\end{equation} 
	\end{lemma}
	
	\begin{proof}[Proof of Lemma \ref{lem:step-1-error}]
		For simplicity, we omit the subscript $\tau$ in the proof of Lemma \ref{lem:step-1-error}, that is,  $\bbeta^{(k)}=\bbeta+\bdelta^{(k)}$, $\bdelta = \sum_{0\le k\le K} \alpha_k\bdelta^{(k)}$, $\bw = \bbeta + \bdelta$, and \[\by^{(k)}=\bZ^{(k)}\bbeta^{(k)}+\brackets{\bff^{(k)}-\bZ^{(k)}\bbeta^{(k)}}+\be^{(k)}.\] Then the error of the estimator $\hat \bw$ in Step I can be decomposed as
		\begin{equation}
			\begin{aligned}
				\hat \bw - \bw &= \hat \bSigma^{-1}\paran{\frac{1}{n_{\calK}}\sum_{0\le k \le K}\paran{\bZ^{(k)}}^{\top}\by^{(k)}}-\bbeta-\bdelta\\
				&= \underbrace{\paran{\hat\bSigma^{-1}\sum_{0\le k \le K} \alpha_k \hat \bSigma^{(k)}\bdelta^{(k)}-\bdelta}}_{\bE_1} + \underbrace{\hat \bSigma^{-1}\paran{\frac{1}{n_{\calK}}\sum_{0\le k \le K}\paran{\bZ^{(k)}}^{\top}\brackets{\bff^{(k)}-\bZ^{(k)}\bbeta^{(k)}}}}_{\bE_2}\\
				&\quad +\underbrace{\hat \bSigma^{-1}\paran{\frac{1}{n_{\calK}}\sum_{0\le k \le K}\paran{\bZ^{(k)}}^{\top}\be^{(k)}}}_{\bE_3}.
			\end{aligned}
		\end{equation}
		Then we separately bound $\bE_1, \bE_2$ and $\bE_3$. 
		
		\textbf{Bound on $\bE_1$.} Note that
		\[\bE_1=\hat\bSigma^{-1}\Big(\sum_{0\le k \le K} \alpha_k \paran{\hat \bSigma^{(k)}-\hat\bSigma}\bdelta^{(k)}\Big).\]
		By the proof of Lemma 3 and Lemma 4 in \cite{shi2020statistical}, we have
		\[\norm{ \bSigma - \hat\bSigma}_2=O_{\PP}\paran{\sqrt{\frac{p \log^2 (n_{\calK})}{n_{\calK}}}}, \norm{ \bSigma - \hat\bSigma^{(k)}}_2=O_{\PP}\paran{ \sqrt{\frac{p\log^2 (n_{k})}{n_{k}}}}.\]
		Therefore,
		\begin{equation}
			\begin{aligned}
				\norm{\bE_1}_2 & \leq \norm{\hat\bSigma^{-1}}_2 \brackets{ \sum_{0\le k \le K} \alpha_k\norm{ \paran{\hat \bSigma^{(k)}-\hat\bSigma}}_2 \norm{\bdelta^{(k)}}_2}\\
				%&\lesssim \sqrt{\frac{p\log^2 (n_{\calK})}{n_{\calK}}}\paran{1+\sum_{0\le k \le K}\sqrt{\frac{n_k}{n_{\calK}}}}\\
				&= O_{\PP}\left(h\sqrt{\frac{p\log^2 (n_{\calK})}{n_{\calK}}}\right).
			\end{aligned}
		\end{equation}
		%\wenbo{can we improve by infinity norm?}
		% using the fact that $\norm{\bdelta^{(k)}}_2 \leq  \norm{\bdelta^{(k)}}_1 \leq h$ and $\norm{\bdelta}_2 \leq  \norm{\bSigma^{-1}}_2 \sum_{0\le k \le K} \alpha_k \norm{\bSigma^{(k)}}_2\norm{\bdelta^{(k)}}_2\lesssim h$.

		\textbf{Bound on $\bE_2$.} Since $\sup_{k} \norm{\bff^{(k)}-\bZ^{(k)}\bbeta^{(k)}}_{\infty} \leq C'p^{-\kappa/d}$, we have that for any vector $\bv \in \RR^{mp}$, 
		\begin{align*}
			&\quad \frac{1}{n_{\calK}}\bv^{\top}\sum_{0\le k \le K}\paran{\bZ^{(k)}}^{\top}\brackets{\bff^{(k)}-\bZ^{(k)}\bbeta^{(k)}}\\
			&=\frac{1}{n_{\calK}}\sum_{0\le k \le K}\paran{\bZ^{(k)}\bv}^{\top}\brackets{\bff^{(k)}-\bZ^{(k)}\bbeta^{(k)}}\\
			&\leq  \frac{C'p^{-\kappa/d}}{n_{\calK}} \sum_{0\le k \le K} \norm{\bZ^{(k)}\bv}_1\\
			&\leq C'p^{-\kappa/d}\sqrt{\bv^{\top}\paran{\frac{1}{n_{\calK}}\sum_{0\le k \le K} \paran{\bZ^{(k)}}^{\top}\bZ^{(k)}}\bv}\\
			& \leq C'p^{-\kappa/d}\sqrt{\norm{\hat\bSigma}_2}\norm{\bv}_2,
		\end{align*}
		where the second-to-the-last inequality follows from the Cauchy-Schwartz inequality. Therefore, we have that 
		\[\norm{\bE_2}_2  \lesssim p^{-\kappa/d}.\]
		
		\textbf{Bound on $\bE_3$.} We have
		\begin{align*}
			&\quad\EE\brackets{\norm{\sum_{0\le k \le K}\paran{\bZ^{(k)}}^{\top}\be^{(k)}}^2_2}\\
			&=\EE\brackets{\paran{\sum_{0\le k \le K}\sum_{i,t}\bxi\paran{\bX_{i,t}^{(k)}, A_{i,t}^{(k)}}e_{i,t}^{(k)}}^{\top}\paran{\sum_{0\le k \le K}\sum_{i, t}\bxi\paran{\bX_{i,t}^{(k)}, A_{i,t}^{(k)}}e_{i,t}^{(k)}}}\\
			&=\EE\brackets{\sum_{0\le k \le K}\sum_{i=1}^{n_k}\bxi^{\top}\big(\bX_{i,t}^{(k)}, A_{i,t}^{(k)}\big)\bxi\big(\bX_{i,t}^{(k)}, A_{i,t}^{(k)}\big) \big(e_{i,t}^{(k)}\big)^2}\\
			&\leq n_{\calK} \sup_{\bx} \norm{\bphi(\bx)}_2^2 \sup_{k} \EE\brackets{\big(e_{i,t}^{(k)}\big)^2},
		\end{align*}
		where the second equality follows from the independence between $e_{i,t}^{(k)}$ with different $(i, t)$ or different $k$. Lemma 2 in \cite{shi2020statistical} shows that $\sup_{\bx} \norm{\bphi(\bx)}_2^2 \lesssim p$. Since $e_{i, t}^{(k)}$ is sub-Gaussian with a uniform sub-Gaussian parameter, we obtain that \[\EE\brackets{\norm{\sum_{0\le k \le K}\paran{\bZ^{(k)}}^{\top}\be^{(k)}}^2_2} \lesssim n_{\calK}p.\]
		Then by Markov's inequality, we have 
		\[\norm{\bE_3}_2 =  O_{\PP}\paran{\sqrt{\frac{p}{n_{\calK}}}}.\]
		
		Combining the bounds on $\bE_1, \bE_2$ and $\bE_3$ leads to \eqref{eq:step-1-error}.
	\end{proof}

	Now we turn to the analysis of Step \rom{2} in Algorithm \ref{algo:fitted-q-transfer}. The quantity $\bdelta_{\tau}$ defined in \eqref{eq:def:bdelta-homo} is a weighted sum of the difference vectors $\bdelta^{(k)}_{\tau}$ between the tasks. Let $\bv^{(k)}_{\tau}=\bdelta^{(k)}_{\tau}-\bdelta_{\tau}$. We have the following result for the estimation error of $\hat\bdelta^{(k)}_{\tau}$:
	\begin{lemma}\label{lem:step-2-error}
		Under the assumptions of Theorem \ref{thm:Q-converge-homo}, it holds that
		\[\norm{\hat\bdelta^{(k)}_{\tau}-\bv^{(k)}_{\tau}}_2=O_{\PP}\paran{\sqrt{\frac{p}{n_{\calK}}}+p^{-\kappa/d}+ \sqrt{\frac{p\log p}{n_k}} \wedge \sqrt{h}\left(\frac{\log p}{n_k}\right)^{1/4} \wedge h}.\]  
	\end{lemma}
	
	\begin{proof}[Proof of Lemma \ref{lem:step-2-error}]
		For simplicity, we omit the subscript $\tau$ in the proof of Lemma \ref{lem:step-2-error}.
		By the definition of $\bdelta$, it is straightforward to show that $\norm{\bdelta}_1 \lesssim h$.  By the definition of $\hat \bdelta^{(k)}$ in Step \rom{2},  it holds that
		
		\[\frac{1}{2n_{k}}\norm{\by^{(k)}-\bZ^{(k)}\paran{\hat{\bw}+\hat\bdelta^{(k)}}}_2^2+\lambda_{\delta}\norm{\hat \bdelta^{(k)}}_1 \leq \frac{1}{2n_{k}}\norm{\by^{(k)}-\bZ^{(k)}\paran{\hat{\bw}+\bv^{(k)}}}_2^2+\lambda_{\delta}\norm{\bv^{(k)}}_1.\]
		With some algebra, the above equation is equivalent to 
		\begin{align*}
			\frac{1}{2}\big(\hat\bdelta^{(k)}-\bv^{(k)}\big)^{\top}\hat \bSigma^{(k)}\big(\hat\bdelta^{(k)}-\bv^{(k)}\big) &\leq \lambda_{\delta}\left(\norm{\bv^{(k)}}_1-\norm{\hat\bdelta^{(k)}}_1\right)\\
			&\quad+\frac{1}{n_k}\left(\be^{(k)}\right)^{\top}\bZ^{(k)}\big(\hat\bdelta^{(k)}-\bv^{(k)}\big)-\big(\hat\bw-\bw\big)^{\top}\hat\bSigma^{(k)}\big(\hat\bdelta^{(k)}-\bv^{(k)}\big)\\
			&\quad + \frac{1}{n_k}\left[\bff^{(k)}-\bZ^{(k)}\bbeta^{(k)}\right]^{\top}\bZ^{(k)}\big(\hat\bdelta^{(k)}-\bv^{(k)}\big).
		\end{align*}
		Since $\be^{(k)}$ is sub-Gaussian, it is standard to show that 
		\[\norm{\frac{1}{n_k}(\be^{(k)})^{\top}\bZ^{(k)}}_{\infty} = O_{\PP}\left(\sqrt{\frac{\log p}{n_k}}\right).\]
		By taking $\lambda_{\delta}=c_{\lambda}\sqrt{\frac{\log p}{n_k}}$ for sufficiently large $c_{\lambda}$, we have \[\frac{1}{n_k}(\be^{(k)})^{\top}\bZ^{(k)}\big(\hat\bdelta^{(k)}-\bv^{(k)}\big) \leq \frac{\lambda_{\delta}}{2}\norm{\hat\bdelta^{(k)}-\bv^{(k)}}_1,\] with probability tending to one. 
		Furthermore, since $c_{\Sigma}^{-1}\leq \lambda_{\min}\paran{\hat\bSigma^{(k)}}\leq \lambda_{\max}\paran{\hat\bSigma^{(k)}} \leq c_{\Sigma}$, we have
		\[	\frac{1}{2}\big(\hat\bdelta^{(k)}-\bv^{(k)}\big)^{\top}\hat \bSigma^{(k)}\big(\hat\bdelta^{(k)}-\bv^{(k)}\big) \geq \frac{1}{2c_{\Sigma}}\norm{\hat\bdelta^{(k)}-\bv^{(k)}}_2^2,\]
		and \[-\big(\hat\bw-\bw\big)^{\top}\hat\bSigma^{(k)}\big(\hat\bdelta^{(k)}-\bv^{(k)}\big) \leq c_{\Sigma}\norm{\hat\bw-\bw}_2\norm{\hat\bdelta^{(k)}-\bv^{(k)}}_2 \leq 4c^2_{\Sigma}\norm{\hat\bw-\bw}^2_2+\frac{1}{4c_{\Sigma}}\norm{\hat\bdelta^{(k)}-\bv^{(k)}}^2_2.\]
		Finally, by the proof of the bound on $\bE_2$ in the proof of Lemma \ref{lem:step-1-error}, we have that 
		\[ \frac{1}{n_k}\left[\bff^{(k)}-\bZ^{(k)}\bbeta^{(k)}\right]^{\top}\bZ^{(k)}\big(\hat\bdelta^{(k)}-\bv^{(k)}\big) \leq C'p^{-\kappa/d}c_{\Sigma}\norm{\hat\bdelta^{(k)}-\bv^{(k)}}_2.\]
		Combining the above inequalities leads to 
		\begin{align*}
			&\quad \frac{1}{4c_{\Sigma}}\norm{\hat\bdelta^{(k)}-\bv^{(k)}}^2_2 \\
			&\leq \lambda_{\delta}\left(\norm{\bv^{(k)}}_1-\norm{\hat\bdelta^{(k)}}_1\right)+\frac{\lambda_{\delta}}{2}\norm{\hat\bdelta^{(k)}-\bv^{(k)}}_1+4c^2_{\Sigma}\norm{\hat\bw-\bw}^2_2+C'p^{-\kappa/d}c_{\Sigma}\norm{\hat\bdelta^{(k)}-\bv^{(k)}}_2.
		\end{align*}
		Among the three terms $\lambda_{\delta}\left(\norm{\bv^{(k)}}_1-\norm{\hat\bdelta^{(k)}}_1\right)+\frac{\lambda_{\delta}}{2}\norm{\hat\bdelta^{(k)}-\bv^{(k)}}_1$, $4c^2_{\Sigma}\norm{\hat\bw-\bw}^2_2$, and $C'p^{-\kappa/d}c_{\Sigma}\norm{\hat\bdelta^{(k)}-\bv^{(k)}}_2$ on the RHS of the above inequality, if
		\begin{enumerate}[label=(\alph*)]
			\item 	$\lambda_{\delta}\left(\norm{\bv^{(k)}}_1-\norm{\hat\bdelta^{(k)}}_1\right)+\frac{\lambda_{\delta}}{2}\norm{\hat\bdelta^{(k)}-\bv^{(k)}}_1$ is the dominant one, we first have that
			\[\frac{1}{c_{\Sigma}}\norm{\hat\bdelta^{(k)}-\bv^{(k)}}^2_2  \lesssim \lambda_{\delta} \norm{\hat\bdelta^{(k)}-\bv^{(k)}}_1 \lesssim \lambda_{\delta} \sqrt{p}\norm{\hat\bdelta^{(k)}-\bv^{(k)}}_2,\]
			yielding that
			\[\norm{\hat\bdelta^{(k)}-\bv^{(k)}}_2 \lesssim \sqrt{p}\lambda_{\delta}.\]
			Moreover, using the fact that $\norm{\hat\bdelta^{(k)}}_1 \geq  \norm{\hat\bdelta^{(k)}-\bv^{(k)}}_1-\norm{\bv^{(k)}}_1 $, we obtain
			\[\frac{1}{c_{\Sigma}}\norm{\hat\bdelta^{(k)}-\bv^{(k)}}^2_2 \lesssim 2\lambda_{\delta}\norm{\bv^{(k)}}_1-\frac{\lambda_{\delta}}{2}\norm{\hat\bdelta^{(k)}-\bv^{(k)}}_1.\]
			Therefore,
			\[\frac{1}{c_{\Sigma}}\norm{\hat\bdelta^{(k)}-\bv^{(k)}}^2_2 \lesssim  \lambda_{\delta} \norm{\bv^{(k)}}_1,\]
			and
			\[\frac{\lambda_{\delta}}{2}\norm{\hat\bdelta^{(k)}-\bv^{(k)}}_2 \lesssim \frac{\lambda_{\delta}}{2}\norm{\hat\bdelta^{(k)}-\bv^{(k)}}_1 \lesssim  \lambda_{\delta} \norm{\bv^{(k)}}_1.\]
			In conclusion, we have 
			\[\norm{\hat\bdelta^{(k)}-\bv^{(k)}}_2 \lesssim \min\braces{\sqrt{p}\lambda_{\delta}, \norm{\bv^{(k)}}_1, \sqrt{\lambda_{\delta}\norm{\bv^{(k)}}_1}}. \]
			\item $4c^2_{\Sigma}\norm{\hat\bw-\bw}^2_2$ is the dominant one, then we have
			\[\norm{\hat\bdelta^{(k)}-\bv^{(k)}}_2 \lesssim \norm{\hat\bw-\bw}_2. \]
			\item $C'p^{-\kappa/d}c_{\Sigma}\norm{\hat\bdelta^{(k)}-\bv^{(k)}}_2$ is the dominant one, then we have
			\[\norm{\hat\bdelta^{(k)}-\bv^{(k)}}_2 \lesssim p^{-\kappa/d}.\] 
		\end{enumerate}
		Recall that $\bv^{(k)}=\bdelta^{(k)}-\bdelta$, which implies $\norm{\bv^{(k)}}_1 \lesssim h$. By Lemma \ref{lem:step-1-error}, \[\norm{\hat\bw-\bw}_2=O_{\PP}\paran{\sqrt{\frac{p}{n_{\calK}}}+h\sqrt{\frac{p\log^2 (n_{\calK})}{n_{\calK}}}+p^{-\kappa/d}}.\] 
		With $\lambda_{\delta} \asymp \sqrt{\log p/ n_{k}}$, we finally obtain that
		\[\norm{\hat\bdelta^{(k)}-\bv^{(k)}}_2=O_{\PP}\paran{\sqrt{\frac{p}{n_{\calK}}}+p^{-\kappa/d}+ \sqrt{\frac{p \log p}{n_k}} \wedge 
			\sqrt{h}\left(\frac{\log p}{n_k}\right)^{1/4}\wedge h}.\]   
	\end{proof}
	
	By Lemmas \ref{lem:step-1-error} and \ref{lem:step-2-error}, since $\hat\bbeta^{(0)}_{\tau}-\bbeta^{(0)}_{\tau}=\hat\bw_{\tau}+\hat\bdelta^{(0)}_{\tau}-\bbeta_{\tau}-\bdelta^{(0)}_{\tau}=\hat\bw_{\tau}-\bbeta_{\tau}-\bdelta_{\tau}+\hat\bdelta^{(0)}_{\tau}-\bv^{(0)}_{\tau}$, we have
	\[\norm{\hat\bbeta^{(0)}_{\tau} - \bbeta^{(0)}_{\tau}}_2 
	= O_{\PP}\left(\sqrt{\frac{p}{n_{\calK}}}+p^{-\kappa/d}+\sqrt{\frac{p \log p}{n_k}} \wedge 
	\sqrt{h}\left(\frac{\log p}{n_k}\right)^{1/4}\wedge h\right).\]
	By \eqref{eq:Emub-homo-0}, we have
	\begin{equation}\label{eq:Emub-homo-1}
		\begin{aligned}
			&\quad\EE_{\widetilde{\mu},\widetilde{\pi}}\abs{Q^{*}-\hat Q^{(0)}_{{\Upsilon}}}(\bx, a)\\
			&\leq \sum_{\tau=0}^{\Upsilon-2}\gamma^{\Upsilon-1-\tau}\EE_{\widetilde{\mu},\widetilde{\pi}}\left[\sup_{\omega_1,\dots,\omega_{\Upsilon-1-\tau}}P^{\omega_{\Upsilon-1-\tau}}P^{\omega_{\Upsilon-\tau-2}}\cdots P^{\omega_1}\abs{\calT^{(0)} \hat Q^{(0)}_{\tau}-\hat Q^{(0)}_{\tau+1}} \right]\\
			&\quad 
			+ \EE_{\widetilde{\mu},\widetilde{\pi}} \abs{\calT^{(0)} \hat Q^{(0)}_{\Upsilon-1}-\hat Q^{(0)}_{\Upsilon}} +\frac{2\gamma^{\Upsilon}R_{\max}}{1-\gamma}.
		\end{aligned}
	\end{equation}
	where $\omega_1,\dots,\omega_{\Upsilon-\tau-1}$ are policy functions and $\EE_{\widetilde{\mu}, \widetilde{\pi}}$ denotes the expectation with respect to $(\bx, a) \sim \widetilde\mu(\bx) \widetilde\pi(a \,|\, \bx)$, for any distribution $\widetilde{\mu}$ with bounded density and policy $\widetilde{\pi}$. By definition, for any function $f$,
	\begin{align*}
		&\quad P^{\omega_{J}}\cdots P^{\omega_1}f(\bx, a)\\
		&=\int_{\cX \times \cA}\cdots\int_{\cX \times \cA}g_1(\bx_2, a_2;f)\rho( \bx_2 \mid  \bx_{3}, a_{3})\omega_2(a_2\mid \bx_2)\mathrm{d}(\bx_{2}, a_2) \cdots \rho( \bx_J \mid  \bx, a)\omega_J(a_J\mid \bx_J)\mathrm{d}(\bx_J, a_J).
	\end{align*}
	where $g_1(\bx_2, a_2;f):=\int_{\cX \times \cA}f(\bx_{1}, a_1)\rho( \bx_1 \mid  \bx_2, a_2)\omega_1(a_1\mid \bx_1)\mathrm{d}(\bx_{1}, a_1)$. Assumption \ref{assum:kappa-smooth} implies that $\rho \leq \overline{\rho}$ for some constant $\overline{\rho}$, and thus, with probability approaching one,
	\begin{align*}
		&\quad g_1\left(\bx, a; \abs{\calT^{(0)} \hat Q^{(0)}_{\tau-1}-\hat Q^{(0)}_{\tau}} \right)\\
		&\lesssim  \int_{\calX} \sum_{a' \in \calA} \left[\bxi^{\top}(\bx', a')\left(\hat\bbeta_{\tau}^{(0)}-\bbeta_{\tau}^{(0)}\right) + p^{-\kappa/d}\right]\rho( \bx' \mid  \bx, a)\omega_1(a'\mid \bx') \mathrm{d}\bx'\\
		&\lesssim p^{-\kappa/d} + \sqrt{\left(\hat\bbeta_{\tau}^{(0)}-\bbeta_{\tau}^{(0)}\right)^{\top}\int_{\cX} \sum_{a'}\bxi(\bx', a')\bxi^{\top}(\bx', a')\rho( \bx' \mid  \bx, a)\omega_1(a'\mid \bx') \mathrm{d}\bx'\left(\hat\bbeta_{\tau}^{(0)}-\bbeta_{\tau}^{(0)}\right) }\\
		&\lesssim p^{-\kappa/d} +  \sup_{\tau \geq 1}\norm{\hat\bbeta_{\tau}^{(0)}-\bbeta_{\tau}^{(0)}}_2,
	\end{align*}
	which, together with \eqref{eq:Emub-homo-1},  implies that 
	\[\EE_{\widetilde{\mu},\widetilde{\pi}}\abs{Q^{*}-\hat Q^{(0)}_{{\Upsilon}}}(\bx, a) \lesssim \frac{1}{1-\gamma} \left(\sqrt{\frac{p}{n_{\calK}}}+p^{-\kappa/d}+\sqrt{\frac{p \log p}{n_k}} \wedge 
	\sqrt{h}\left(\frac{\log p}{n_k}\right)^{1/4}\wedge h\right)+\frac{2\gamma^{\Upsilon}R_{\max}}{1-\gamma},\]
	for any $\widetilde{\mu},\widetilde{\pi}$. Then the desired result for $v^*-v^{\widehat{\pi}_{\Upsilon}}$ is obtained by applying Lemma 13 in \cite{chen2019information}. 
\end{proof}

\paragraph{Proof of Theorem \ref{thm:Q-converge-hetero}.}

\begin{proof}
	
	%\wenbo{I cannot figure out where $\frac{s\log p}{n_0}$ comes from in Li Sai's paper.}
	
	The proof of Theorem \ref{thm:Q-converge-hetero} is partly analogous to the that of Theorem \ref{thm:Q-converge-homo}. We will focus on the different parts between them.
	
	Firstly, for the distribution of $\calS^{(k)}=\left\{\big(\bX_{i, t}^{(k)}, A_{i, t}^{(k)}, R_{i, t}^{(k)}, \bX_{i, t+1}^{(k)}\big)\right\}$ for $i \in [I^{(k)}]$, $t\in \{0\} \cup [T]$, the initial distribution $\nu^{(k)}$,  the underlying behavior policy $b^{(k)}(\cdot \,|\, \bx)$, the transition kernel $P^{(k)}\big(\cdot \,|\, \bx, a\big)$, and its density $\rho^{(k)}(\bx' \,|\, \bx, a)$ all depend on the task index $k$ in the heterogeneous transition setting. Still, the reward $R_{i,t}^{(k)}=r^{(k)}\left(\bX^{(k)}_{i,t}, A^{(k)}_{i,t}\right) + \eta^{(k)}_{i,t}$. %The tuple set $\Big\{\big(\bX^{(k)}_{1,t}, A^{(k)}_{1,t}, R_{1,t}^{(k)}, \bX^{(k)}_{1,t+1}\big)\Big\}_{t=0}^{T-1}$ forms a trajectory, and there are multiple i.i.d. trajectories $\Big\{\big(\bX^{(k)}_{i,t}, A^{(k)}_{i,t}, R_{i,t}^{(k)}, \bX^{(k)}_{i,t+1}\big)\Big\}$ in one task. 
	%We further split $\calS^{(k)}$ into $\tau$ subsets, $\left\{\calS_{\tau}^{(k)}\right\}_{\tau \in [\Upsilon]}$, each with sample size $n_{k}$, and use subset $\calS_{\tau}^{(k)}$ in iteration $\tau$, which ensures that the data used in different iterations are independent. %For simplicity, we put the tuples of all of the trajectories together and let $n_k=I^{(k)}T$ be the sample size of task $k$. 
	Assumption \ref{assum:distribution} ensures that, for any $i$, $\left\{\left(\bX^{(k)}_{i,t}, A^{(k)}_{i,t} \right)\right\}_{t =0}^{T-1}$ is a Markov chain on $\calX \times \calA$ with the transition kernel $P^{(k)}(\bx^{\prime}\,|\, \bx, a)b^{(k)}(a^{\prime}\,|\,\bx^{\prime})$, and  $\left\{\left(\bX^{(k)}_{i,t}\right)\right\}_{t =0}^{T-1}$ is also a Markov chain with the transition kernel $\sum_{a \in \calA} b^{(k)}(a \,|\,\bx)P^{(k)}(\bx^{\prime}\,|\, \bx, a)$. The latter chain has an invariant distribution $\mu^{(k)}(\bx)$, which implies that the former chain has an invariant distribution $\mu^{(k)}(\bx) b^{(k)}(a \,|\, \bx)$. %Furthermore, since $\nu^{(0)}=\mu^{(0)}$, the target task follows the invariant distribution, i.e.,  $\left(\bX^{(0)}_{i,t}, A^{(0)}_{i,t} \right) \sim \mu^{(0)}(\bx) b^{(0)}(a \mid \bx)$ for all $i, t$. %Also notice that the continuity of %$\calX$ is bounded in $L_2$ norm,$\mu^{(0)}(\cdot)$ implies that it is bounded away from 0 and $\infty$ on $\supp(\mu^{(k)}(\bx))$, since $\supp(\mu^{(k)}(\bx))$ is bounded. %, and $\mu^{(0)}(\cdot)$ is uniformly bounded away from zero and infinity. 
	%Moreover, $\inf_{a, \bx} b^{(0)}(a \,|\, \bx) \geq \underline{b} > 0$. 
	
	Redefine the Bellman optimality operator $\calT^{(k)}$ on the $k$-th task by
	\begin{equation}
		\calT^{(k)} Q \paran{\bx, a} = r^{(k)}\paran{\bx, a} + \gamma\cdot\EE\brackets{ \underset{a'\in\calA}{\max}\,Q\paran{\bx', a'} \,\big|\, \bx'\sim P^{(k)}\paran{\cdot\,|\,\bx, a} }.
	\end{equation}
	%We have $\calT^{(k)} Q^{*, (k)} = Q^{*, (k)}$.
	For a function $f(\bx, a)$ defined on $\calS \times \calA$ and a policy function $\omega(\cdot \mid \bx)$, redefine the operator $P^{\omega}$ by
	\[P^{\omega}f(\bx, a):=\EE\left[f(\bx^{\prime}, a^{\prime}) \,|\, \bx^{\prime} \sim P^{(0)}(\cdot \,|\, \bx, a), a^{\prime} \sim \omega(\cdot \,|\, \bx^{\prime}) \right].\]
	
	%\begin{proof}
	
	%Let $Q^*$ be the optimal action-value function on the target dataset. The iterative algorithm gives a series of estimators $\hat Q^{(0)}_\tau$ for $Q^*$ and finally an estimated optimal policy $\pi_\Upsilon$. By the proof of Theorem 6.1 in \cite{fan2020theoretical}, % Equation (C.27) and the equation under (C.27) 
	Similar to \eqref{eq:Emub-homo}, it holds that
	
	\begin{equation}\label{eq:Emub-hetero}
		\begin{aligned}
			&\quad\abs{Q^{*}-\hat Q^{(0)}_{{\Upsilon}}}\\
			&\leq \sum_{\tau=0}^{\Upsilon-2}\gamma^{\Upsilon-1-\tau}\sup_{\omega_1,\dots,\omega_{\Upsilon-1-\tau}}P^{\omega_{\Upsilon-1-\tau}}P^{\omega_{\Upsilon-\tau-2}}\cdots P^{\omega_1}\abs{\calT^{(0)} \hat Q^{(0)}_{\tau}-\hat Q^{(0)}_{\tau+1}} \\
			&\quad 
			+  \abs{\calT^{(0)} \hat Q^{(0)}_{\Upsilon-1}-\hat Q^{(0)}_{\Upsilon}} +\frac{2\gamma^{\Upsilon}R_{\max}}{1-\gamma}.
		\end{aligned}
	\end{equation}
	where the supreme is taken over all possible policies $\omega_1,\dots,\omega_{\Upsilon-\tau+j}$.%, and $\EE_{\mu^{(0)}, b^{(0)}}$ denotes the expectation with respect to the invariant distribution $\mu^{(0)}(\bx) b^{(0)}(a \,|\, \bx)$. %In the following proof, we focus on bounding $\abs{\calT^{(0)} \hat Q^{(0)}_{\tau-1} - \hat Q^{(0)}_\tau}(\bx, a)$ for a fixed $ \tau  \in [\Upsilon]$, where $\hat Q^{(k)}_{\tau}(\bx, a)=\bxi(\bx, a)^{\top}\hbbeta^{(k)}_{\tau}$ in our algorithm.

	For the same reason as in the homogeneous setting, we rewrite
	\begin{equation}
		Y^{(k), \tau}_{i,t} = \calT^{(k)}\hat Q_{\tau-1}^{(k)}\paran{\bX^{(k)}_{i,t}, A^{(k)}_{i,t}} + e^{(k), \tau}_{i,t}, 
	\end{equation}
	where %$\calT^{(k)}\hat Q_{\tau-1}^{(k)}:\calX\times\calA\mapsto\RR$ is an unknown regression function to be estimated. Recall that $R^{(k)}_{i,t}=r^{(k)}\paran{\bX_{i,t}^{(k)}, A_{i,t}^{(k)}}+\eta^{(k)}_{i,t}$ and 
	\[\paran{\calT^{(k)} \hat Q^{(k)}_{\tau-1}}\paran{\bX^{(k)}_{i,t}, A_{i,t}^{(k)}}=r^{(k)}\paran{\bX_{i,t}^{(k)}, A_{i,t}^{(k)}}+\gamma \E{\underset{a \in \calA}{\max} \hat Q^{(k)}_{\tau-1}\paran{\bx', a} \big| \bx' \sim P^{(k)}\paran{\cdot \mid \bX_{i,t}^{(k)}, A_{i,t}^{(k)}}  }.\]
	The regression noise term $e^{(k), \tau}_{i,t}$ is zero-mean sub-Gaussian, since
	\[
	e^{(k), \tau}_{i,t} = \eta^{(k)}_{i, t} + \gamma \paran{1-\mathbb E}\left[ \underset{a \in \calA}{\max} \hat Q^{(k)}_{\tau-1}\paran{\bx', a} \big| \bx' \sim P^{(k)}\paran{\cdot \mid \bX_{i,t}^{(k)}, A_{i,t}^{(k)}}  \right],\]
	where $\eta^{(k)}_{i, t}$ is zero-mean sub-Gaussian, and the second term is bounded. 
	
	%Let $\calF_p$ be a finite dictionary of functions $\phi_j:\calX\times\calA\mapsto\RR$, $j\in[p]$. 
	%By Lemma \ref{thm:bellman-opt-close} and the Assumption that $\calF_p$ is a collection of basis functions (e.g., wavelets, splines with fixed knots, step functions), the unknown function $f$ can be well approximated by a member of the span of $\calF_p$. 
	%In this paper, we have in mind the situation where $p >> n$, and $f$ can be estimated reasonably only because it can be approximated by a linear combination of a small number of members of $\calF_p$, or, in other words, it has a sparse approximation in the span of $\calF_p$. \wenbo{How to state this as an assumption?}
	%In each step $\tau$, we use $\hat Q^{(k)}_{\tau}(\bx, a)=\bxi^{\top}(\bx, a)\hbbeta_\tau^{(k)}$ to estimate $\paran{\calT^{(k)} \hat Q^{(k)}_{\tau-1}}\paran{\bx, a}$, where 
	%\[\bxi(\bx, a):=\left[\bphi^{\top}(\bx)\II(a=1),\bphi^{\top}(\bx)\II(a=2),\dots,\bphi^{\top}(\bx)\II(a=m)\right]^{\top},\]
	%and $\bphi(\cdot)=\paran{\phi_1(\cdot),\dots, \phi_p(\cdot)}^{\top}$ is a set of sieve basis functions.  
	With heterogeneous transition kernel,  
	\[\paran{\calT^{(k)} \hat Q^{(k)}_{\tau-1}}\paran{\bx, a}=r^{(k)}\paran{\bx, a}+\gamma \int_{\calX}\brackets{\underset{a' \in \calA}{\max} \hat Q^{(k)}_{\tau-1}\paran{\bx', a'}}  \rho^{(k)}\paran{\bx'\mid \bx, a } \mathrm{d} \bx'.\]
	%The estimation is guaranteed by the following property: 
	%For any function $f(\bx, a)$ that satisfies $f(\cdot, a) \in \Lambda(\kappa, c)$ for all $a\in\calA$,  there exists a set of vectors $\braces{\bbeta_a\in\RR^p}_{a \in \calA}$ that 
	%\begin{equation*}
	%	\underset{\bx\in\calX, a\in \calA}{\sup} \abs{  f\paran{\bx, a} - \bbeta_a^\top\bphi(\bx)} 
	%	\le C p^{-\kappa/d},
	%\end{equation*}
	%for some positive constant $C$. Let $\bbeta=\paran{\bbeta_1^{\top},\dots,\bbeta_m^{\top}}^{\top}$, and then we have
	%\[	\underset{\bx\in\calX, a\in \calA}{\sup} \abs{  f\paran{\bx, a} - \bbeta^\top\bxi(\bx, a)} 
	%\le C p^{-\kappa/d}.\]
	By Assumption \ref{assum:kappa-smooth} and the definition of $h$, there exist $\left\{\bbeta^{(k)}_{r}\right\}_{k \in \calK}$, $\left\{\bbeta^{(k)}_{\rho}(\bx')\right\}_{\bx' \in \calX, k \in \calK}$ such that 
	\[\sup_{\bx, a, k} \abs{r^{(k)}\paran{\bx, a} - \bxi^{\top}(\bx, a)\bbeta^{(k)}_{r}} \leq Cp^{-\kappa/d},\]
	\[\sup_{\bx, a, \bx', k} \abs{\rho^{(k)}\paran{\bx' \,|\, \bx, a} - \bxi^{\top}(\bx, a)\bbeta^{(k)}_{\rho}(\bx')} \leq Cp^{-\kappa/d},\]
	and 
	\[\int_{\calX}\norm{\bbeta^{(k)}_{\rho}(\bx')-\bbeta^{(0)}_{\rho}(\bx')}_1 \mathrm{d}\bx' \leq h, \quad  \norm{\bbeta^{(k)}_{r}-\bbeta^{(0)}_{r}}_1 \leq h. \]%\[\quad \norm{\bbeta_{\rho}^{(0)}(\bx')}_1 \leq C_{\bbeta}.\]
	Redefine 
	\[\bbeta_{\tau}^{(k)}:=\bbeta^{(k)}_{r}+\gamma \int \brackets{\max_{a' \in \calA}\hat Q^{(k)}_{\tau-1}\paran{\bx', a'}}\bbeta^{(k)}_{\rho}(\bx')   \mathrm{d} \bx'.\]
	Since $\max\limits_{a' \in \calA}\hat Q^{(k)}_{\tau-1}\paran{\bx', a'} \lesssim V_{\max}$, it follows that 
	\begin{equation}
		\label{eq:bias_f-beta}
		\sup_{\bx, a, k}\abs{\paran{\calT^{(k)} \hat Q^{(k)}_{\tau-1}}\paran{\bx, a}-\bxi^{\top}(\bx, a)\bbeta_{\tau}^{(k)}} \leq C^{\prime}p^{-\kappa/d},
	\end{equation}
	for some absolute constant $C'$. 
	
	Moreover, note that
	\begin{equation}
		\label{eq:betak-beta0}
		\begin{aligned}
			\bbeta_{\tau}^{(k)}-\bbeta_{\tau}^{(0)}&= \bbeta_r^{(k)}-\bbeta_r^{(0)}+\gamma \int \brackets{\max_{a' \in \calA}\hat Q^{(k)}_{\tau-1}\paran{\bx', a'}}\paran{\bbeta^{(k)}_{\rho}(\bx')-\bbeta^{(0)}_{\rho}(\bx')}   \mathrm{d} \bx'\\
			&\quad + \gamma \int \brackets{\max_{a' \in \calA}\hat Q^{(k)}_{\tau-1}\paran{\bx', a'}- \max_{a' \in \calA}\hat Q^{(0)}_{\tau-1}\paran{\bx', a'}}\bbeta^{(0)}_{\rho}(\bx')   \mathrm{d} \bx'.
		\end{aligned} 
	\end{equation}
	The $\ell_1$-norm of the first two terms on the RHS of \eqref{eq:betak-beta0} can be bounded by a constant times $h$. For the third term, we first have 
	\begin{align*}
		&\quad \int_{\calX}\abs{\underset{a' \in \calA}{\max} \hat Q^{(k)}_{\tau-1}\paran{\bx', a'}-\underset{a' \in \calA}{\max} \hat Q^{(0)}_{\tau-1}\paran{\bx', a'} } \mathrm{d} \bx'\\
		&\leq \int_{\calX}  \abs{\max_{a'} \bxi^{\top}(\bx', a')\paran{\hat\bbeta_{\tau-1}^{(k)}-\hat\bbeta_{\tau-1}^{(0)}}} \mathrm{d} \bx'\\
		&\leq \sqrt{\paran{\hat\bbeta_{\tau-1}^{(k)}-\hat\bbeta_{\tau-1}^{(0)}}^{\top}\int_{\mathcal{X}}\brackets{\bPhi(\bx')\bPhi^{\top}(\bx')}\mathrm{d}\bx'\paran{\hat\bbeta_{\tau-1}^{(k)}-\hat\bbeta_{\tau-1}^{(0)}}}\\
		& \lesssim \norm{\hat\bbeta_{\tau-1}^{(k)}-\hat\bbeta_{\tau-1}^{(0)}}_2,
	\end{align*}
	where $\bPhi(\bx):=\left[\bphi^{\top}(\bx),\bphi^{\top}(\bx),\dots,\bphi^{\top}(\bx)\right]^{\top}$ is the concatenation of $|\calA|$ copies of the basis functions $\bphi(\bx)$. %The last inequality follows from Lemma 2 in \cite{shi2020statistical}, which shows that
	%\[\lam_{\max}\brackets{\int_{\cal X} \bphi(\bx)\bphi^{\top}(\bx) \mathrm{d} x} < \infty.\]
	%and $\rho^{(0)}$ is uniformly bounded away from infinity.
	%The assumption that $\mu^{(0)}$ is continuous and bounded away from 0 implies that $\mathrm{supp}(\mu^{(0)})$ is compact. Since $\mu^{(0)}$ is the distribution of all $\bX_{i, t}^{(0)}$, we can assume without loss of generality that $\calX=\mathrm{supp}(\mu^{(0)})$. Then $\rho^{(0)}$ is also uniformly bounded away from zero. Adding that $\norm{\bbeta_P^{(0)}(\bx')}_1$ is bounded, 
	Then we obtain that the $\ell_1$-norm of the third term of \eqref{eq:betak-beta0} is upper bounded by a constant times $\gamma\norm{\hat\bbeta_{\tau-1}^{(k)}-\hat\bbeta_{\tau-1}^{(0)}}_2$. Therefore, 
	\[\norm{\bbeta_{\tau}^{(k)}-\bbeta_{\tau}^{(0)}}_1 \lesssim h + \gamma \norm{\hat\bbeta_{\tau-1}^{(k)}-\hat\bbeta_{\tau-1}^{(0)}}_2 \lesssim h, \forall k.\]

	%Note that 
	%\begin{equation}
	%	\begin{aligned}
		%		&\quad \left(\bbeta^{(k)}-\bbeta^{(0)}\right)^{\top}\E{\bxi(\bx, a)\bxi^{\top}(\bx, a)}	\left(\bbeta^{(k)}-\bbeta^{(0)}\right)\\&=\E{\paran{f_{\bbeta^{(k)}}(\bx, a)-f_{\bbeta^{(0)}}(\bx, a)}^2} \\
		%		&\lesssim p^{-2\kappa/d} + \E{\left(f^{(k)}(\bx, a)-f^{(0)}(\bx, a)\right)^2}\\
		%		&\lesssim p^{-2\kappa/d} + 
		%	\end{aligned}
	%\end{equation}
	%Recall that
	%\[f^{(k)}(\bx, a):=\paran{\calT^{(k)} \hat Q^{(k)}_{\tau-1}}\paran{\bx, a}=r^{(k)}(\bx, a)+\gamma\int_{\cal X}\left[\max_{a' \in \calA} \bxi^{\top}(\bx', a') \hat\bbeta_{\tau-1}^{(k)}\right] P^{(k)}(\bx' \,|\, \bx, a)\mathrm{d} \bx'.\]
	%\begin{align*}
	%	&\quad \E{\left(f^{(k)}(\bx, a)-f^{(0)}(\bx, a)\right)^2} \\
	%	&\lesssim \E{\paran{r^{(k)}(\bx, a)-r^{(0)}(\bx, a)}^2} + \gamma\E{\int \max_{a'} \abs{\bxi^{\top}(\bx', a')\left(\hat\bbeta_{\tau-1}^{(k)}-\hat\bbeta_{\tau-1}^{(0)}\right)}P^{(k)}(\bx' \,|\, \bx, a)\mathrm{d} \bx'}\\
	%	&\quad + 
	%\end{align*}
	
	Without loss of generality, we assume that $\bbeta_{\tau}^{(k)}=\bbeta_{\tau}+\bdelta_{\tau}^{(k)}$ with $\norm{\bdelta_{\tau}^{(k)}}_1 \leq h$ for $k=0,1,\dots, K$. %, where 
	%\[\widetilde h :=  h \vee \gamma \sup_k\norm{\hat\bbeta_{\tau-1}^{(k)}-\hat\bbeta_{\tau-1}^{(0)}}_2.\] 
	%Let $\hat f_{\bbeta} = \hat \bbeta^{\top}\left[\bphi(\bx)I(a=0),\bphi(\bx)I(a=1),\dots,\bphi(\bx)I(a=m)\right]$ be the estimator given by Algorithm \ref{algo:fitted-q-transfer}, which is refreshed as follows.
	
	%Till now, we have already established the formal definition and properties of the parameter of interest $\bbeta_{\tau}^{(k)}$.
	%In the sequel, we will provide an upper bound for the $\ell_2$ error $\norm{\hat\bbeta_{\tau}^{(k)}-\bbeta_{\tau}^{(k)}}_2$, which is closely related to $\abs{\calT^{(0)} \hat Q^{(0)}_{\tau-1} - \hat Q^{(0)}_\tau}$. We first simplify some notations for a clear presentation.
	Define $f_{\tau}^{(k)}$, $\bZ_{\tau}^{(k)}$, $\by_{\tau}^{(k)}$, $\bff_{\tau}^{(k)}$, and $\be_{\tau}^{(k)} $ the same as in the proof of Theorem \ref{thm:Q-converge-homo}. %$f_{\tau}^{(k)}=\calT^{(k)}\hat Q_{\tau-1}^{(k)}$ and the  $k$-th sample matrix $\bZ_\tau^{(k)}\in \RR^{n_k \times mp}$ has rows are $ \bxi^{\top}\big(\bX_{i,t}^{(k)}, A_{i,t}^{(k)}\big)$, for $(i, t) \in \calS_{\tau}^{(k)}$. Likewise, define 
	%\[\by_{\tau}^{(k)}: = \brackets{Y_{i_1,0}^{(k), \tau}, \dots, Y_{i_1,T-1}^{(k), \tau}, Y_{i_2,0}^{(k), \tau}, \dots, Y^{(k), \tau}_{i_{n_k/T},T-1}}^\top, (i_1,\dots,i_{n_k/T} \in \calS_{\tau}^{(k)})\] 
	%\[\bff_{\tau}^{(k)}:=\brackets{f_{\tau}^{(k)}\big(\bX_{i_1,0}^{(k)}, \bA_{i_1,0}^{(k)}\big), \dots, f_{\tau}^{(k)}\big(\bX_{i_{n_k/T},T-1}^{(k)}, \bA_{i_{n_k/T},T-1}^{(k)}\big)}^{\top},\] 
	%and $\be_{\tau}^{(k)} := \brackets{e_{i_1, 0}^{(k), \tau}, \dots, e_{i_{n_k/T}, T-1}^{(k), \tau}}^\top$. Using these notations, 
	Still, we have 
	\[\by_{\tau}^{(k)}=\bZ^{(k)}\bbeta_{\tau}^{(k)}+\brackets{\bff_{\tau}^{(k)}-\bZ^{(k)}\bbeta_{\tau}^{(k)}}+\be_{\tau}^{(k)},\]
	where $\be_{\tau}^{(k)}$ is sub-Gaussian, and by \eqref{eq:bias_f-beta}, 
	\[\norm{\bff_{\tau}^{(k)}-\bZ^{(k)}\bbeta_{\tau}^{(k)}}_{\infty} \leq C^{\prime}p^{-\kappa/d}.\]
	
	Moreover, recall that $n_{\calK}=\sum_{0\le k\le K}n_k$, $\hat\bSigma^{(k)} := \paran{\bZ^{(k)}}^\top\bZ^{(k)}/n_k$,
	$\hat\bSigma := \sum_{0\le k \le K} \alpha_k \hat\bSigma^{(k)}$ with $\alpha_k = n_k / n_{\calK}$. For the heterogeneous transition setting, we further define 
	$\bSigma^{(k)}:=\E{\hat\bSigma^{(k)}}$, $\overline\bSigma:=\E{\hat\bSigma}$, and 
	\[C_{\Sigma}:=1+\max_{k}\norm{\overline\bSigma^{-1}\paran{\bSigma^{(k)}-\overline\bSigma}}_1.\]
	
	By definition, 
	\[ \bSigma^{(k)} = \frac{1}{T}\EE \left[\sum_{t=0}^{T-1}\bxi\big(\bX_{i,t}^{(k)}, A_{i,t}^{(k)}\big)\bxi^{\top}\big(\bX_{i,t}^{(k)}, A_{i,t}^{(k)}\big)\right].\]% = \int_{\calX} \sum_{a \in \calA} \bxi\big(\bx, a\big)\bxi^{\top}\big(\bx, a\big)\mu^{(k)}(\bx)b^{(k)}(a \mid \bx) \mathrm{d} \bx.\]
	By Assumption \ref{assum:distribution}, there exists a constant $c_{\Sigma} \geq 1$ such that $c_{\Sigma}^{-1}<\lambda_{\min}\paran{ \bSigma^{(k)}}<\lambda_{\max}\paran{\bSigma^{(k)}} < c_{\Sigma}$ for all $k$ and thus, $c_{\Sigma}^{-1}<\lambda_{\min}\paran{\overline\bSigma}<\lambda_{\max}\paran{\overline\bSigma} < c_{\Sigma}$. By the proof of Lemma 4 in \cite{shi2020statistical}, there exists a constant $c'_{\Sigma} \geq 1$ such that $c_{\Sigma}^{'-1}<\lambda_{\min}\paran{\hat \bSigma^{(k)}}<\lambda_{\max}\paran{\hat \bSigma^{(k)}} < c'_{\Sigma}$ for all $k \in \calK$ and $c_{\Sigma}^{'-1}<\lambda_{\min}\paran{\hat \bSigma}<\lambda_{\max}\paran{\hat \bSigma} < c'_{\Sigma}$, with probability tending to one. Without less of generality, we suppose $c_{\Sigma}=c'_{\Sigma}$. Therefore, we have that $\norm{\hat \bSigma^{-1}}_2=O_{\PP}(1)$ and $\sup_{k} \norm{\paran{\hat \bSigma^{(k)}}^{-1}}_2=O_{\PP}(1)$. 
	
	Now we establish the convergence rate for Step \rom{1}. Let
	\begin{equation}
		\label{eq:def:w}
		\bw_{\tau}: = \bbeta_{\tau} + \bdelta_{\tau}, 
	\end{equation}
	where 
	\begin{equation}
		\label{eq:def:bdelta}
		\bdelta_{\tau}: = \overline\bSigma^{-1} \paran{\sum_{0\le k\le K} \alpha_k\bSigma^{(k)} \bdelta_{\tau}^{(k)}}.  
	\end{equation}
	
	We will first prove that
	\begin{equation}	\norm{\bw_{\tau}-\hat\bw_{\tau}}_2 =O_{\PP}\paran{\sqrt{\frac{p\log^2(n_{\calK})}{n_{\calK}}}+p^{-\kappa/d}}.
		\label{eq:step-1-error-hetero}
	\end{equation} 
	For simplicity, we omit the subscript $\tau$ in the following proof when there is no ambiguity.

	%By the proof of Theorem 1 in \cite{shi2020statistical} (more precisely, Lemma 4 in Section E.5), there exists a constant $c_{\Sigma}>0$ such that $c_{\Sigma}^{-1}<\lambda_{\min}\paran{\hat \bSigma^{(k)}}<\lambda_{\max}\paran{\hat \bSigma^{(k)}} < c_{\Sigma}$ for all $k \in \calK$ ($|\calK|$ is finite) and $c_{\Sigma}^{-1}<\lambda_{\min}\paran{\hat \bSigma}<\lambda_{\max}\paran{\hat \bSigma} < c_{\Sigma}$, with probability tending to one. Therefore, we also have that $\norm{\hat \bSigma^{-1}}_2=O_{\PP}(1)$ and $\sup_{k} \norm{\paran{\hat \bSigma^{(k)}}^{-1}}_2=O_{\PP}(1)$. 
	%The quantity $\bdelta$ defined in \eqref{eq:def:bdelta} is a weighted sum of the sparse difference vectors $\bdelta^{(k)}$ between the tasks. In the homogeneous setting, i.e., $\bSigma^{(k)}=\bSigma$ for all $k$, we have $\bdelta=\sum_{0\le k \le K} \frac{n_k}{n_{\calK}} \bdelta^{(k)}$, which satisfies that $\norm{\bdelta}_1 \leq h$. In the heterogeneous setting, we can also show that $\norm{\bdelta}_1 \leq C_{\Sigma} h$ for a constant $C_{\Sigma}$, under some assumptions.\wenbo{to be done}  Therefore, without loss of generality, we can assume that $\bdelta=\bzero$. Otherwise, let 
	
	Recall that $\bbeta^{(k)}=\bbeta+\bdelta^{(k)}$ and \[\by^{(k)}=\bZ^{(k)}\bbeta^{(k)}+\brackets{\bff^{(k)}-\bZ^{(k)}\bbeta^{(k)}}+\be^{(k)}.\] Then the error of the estimator $\hat \bw$ in Step I can be decomposed as
	\begin{equation}
		\begin{aligned}
			\hat \bw - \bw &= \hat \bSigma^{-1}\paran{\frac{1}{n_{\calK}}\sum_{0\le k \le K}\paran{\bZ^{(k)}}^{\top}\by^{(k)}}-\bbeta-\bdelta\\
			&= \underbrace{\paran{\hat\bSigma^{-1}\sum_{0\le k \le K} \alpha_k \hat \bSigma^{(k)}\bdelta^{(k)}-\bdelta}}_{\bE_1} + \underbrace{\hat \bSigma^{-1}\paran{\frac{1}{n_{\calK}}\sum_{0\le k \le K}\paran{\bZ^{(k)}}^{\top}\brackets{\bff^{(k)}-\bZ^{(k)}\bbeta^{(k)}}}}_{\bE_2}\\
			&\quad +\underbrace{\hat \bSigma^{-1}\paran{\frac{1}{n_{\calK}}\sum_{0\le k \le K}\paran{\bZ^{(k)}}^{\top}\be^{(k)}}}_{\bE_3}.
		\end{aligned}
	\end{equation}
	The bounds for $\bE_2$ and $\bE_3$ are identical to the proof of Theorem \ref{thm:Q-converge-homo}. For $\bE_1$, we have that
	\[\bE_1=\hat\bSigma^{-1}\brackets{\paran{ \overline\bSigma - \hat\bSigma} \bdelta + \Big(\sum_{0\le k \le K} \alpha_k \paran{\hat \bSigma^{(k)}-\bSigma^{(k)}}\bdelta^{(k)}\Big)}.\]
	By the proof of Lemma 3 and Lemma 4 in \cite{shi2020statistical}, we have
	\[\norm{ \overline\bSigma - \hat\bSigma}_2=O_{\PP}\paran{\sqrt{\frac{p \log^2 (n_{\calK})}{n_{\calK}}}}, \norm{\bSigma^{(k)} - \hat\bSigma^{(k)}}_2=O_{\PP}\paran{ \sqrt{\frac{p\log^2 (n_{k})}{n_{k}}}}.\]
	Therefore,
	\begin{equation}
		\begin{aligned}
			\norm{\bE_1}_2 & \leq \norm{\hat\bSigma^{-1}}_2 \brackets{\norm{ \overline\bSigma - \hat\bSigma}_2 \norm{\bdelta}_2 + \sum_{0\le k \le K} \alpha_k\norm{ \paran{\hat \bSigma^{(k)}-\bSigma^{(k)}}}_2 \norm{\bdelta^{(k)}}_2}\\
			%&\lesssim \sqrt{\frac{p\log^2 (n_{\calK})}{n_{\calK}}}\paran{1+\sum_{0\le k \le K}\sqrt{\frac{n_k}{n_{\calK}}}}\\
			&\lesssim \sqrt{\frac{p}{n_{\calK}}}.
		\end{aligned}
	\end{equation}

	Combining the bounds on $\bE_1, \bE_2$ and $\bE_3$, we obtain that
	\begin{equation}
		\norm{\bw-\hat\bw}_2 =O_{\PP}\paren{\sqrt{\frac{p}{n_{\calK}}}+p^{-\kappa/d}}.
	\end{equation}

	The analysis of Step \rom{2} is the same as that for Theorem \ref{thm:Q-converge-homo}, while %The quantity $\bdelta$ defined in \eqref{eq:def:bdelta} is a weighted sum of the sparse difference vectors $\bdelta^{(k)}$ between the tasks. %In the homogeneous setting, i.e., $\bSigma^{(k)}=\bSigma$ for all $k$, we have $\bdelta=\sum_{0\le k \le K} \frac{n_k}{n_{\calK}} \bdelta^{(k)}$, which satisfies that $\norm{\bdelta}_1 \leq \widetilde{h}$. In the heterogeneous setting, 
	by the definition of $\bdelta$ and $C_{\Sigma}$, it holds $\norm{\bdelta}_1 \leq C_{\Sigma} h$.  Repeating the same procedure leads to the error rate that
	\[\norm{\hat\bbeta^{(0)} - \bbeta^{(0)}}_2 
	= O_{\PP}\paren{\sqrt{\frac{p}{n_{\calK}}}+p^{-\kappa/d}+\sqrt{\frac{p \log p}{n_k}} \wedge 
		\sqrt{C_{\Sigma}h}\left(\frac{\log p}{n_k}\right)^{1/4}\wedge C_{\Sigma}h}.\]
	%In particular, when $\frac{hp\log^2 (n_{\calK})}{n_{\calK}}\sqrt{\frac{n_0}{\log p}}=o(1)$, the second term is dominated by the fourth term, and hence
	%  \[\sup_{\tau}\norm{\hat\bbeta_{\tau}^{(0)} - \bbeta_{\tau}^{(0)}}_2 
	% = \Op{
		% 	\sqrt{\frac{p}{n_{\calK}}}+p^{-\kappa/d}+C_{\Sigma}h \wedge \sqrt{C_{\Sigma}h}\paran{\frac{\log %p}{n_0}}^{1/4}}.\]
	By \eqref{eq:Emub-hetero}, we have
	\begin{equation}\label{eq:Emub-homo-1}
		\begin{aligned}
			&\quad\EE_{\widetilde{\mu},\widetilde{\pi}}\abs{Q^{*}-\hat Q^{(0)}_{{\Upsilon}}}(\bx, a)\\
			&\leq \sum_{\tau=0}^{\Upsilon-2}\gamma^{\Upsilon-1-\tau}\EE_{\widetilde{\mu},\widetilde{\pi}}\left[\sup_{\omega_1,\dots,\omega_{\Upsilon-1-\tau}}P^{\omega_{\Upsilon-1-\tau}}P^{\omega_{\Upsilon-\tau-2}}\cdots P^{\omega_1}\abs{\calT^{(0)} \hat Q^{(0)}_{\tau}-\hat Q^{(0)}_{\tau+1}} \right]\\
			&\quad 
			+ \EE_{\widetilde{\mu},\widetilde{\pi}} \abs{\calT^{(0)} \hat Q^{(0)}_{\Upsilon-1}-\hat Q^{(0)}_{\Upsilon}} +\frac{2\gamma^{\Upsilon}R_{\max}}{1-\gamma}.
		\end{aligned}
	\end{equation}
	where $\omega_1,\dots,\omega_{\Upsilon-\tau-1}$ are policy functions and $\EE_{\widetilde{\mu}, \widetilde{\pi}}$ denotes the expectation with respect to $(\bx, a) \sim \widetilde\mu(\bx) \widetilde\pi(a \,|\, \bx)$, for any distribution $\widetilde{\mu}$ with bounded density and policy $\widetilde{\pi}$. By definition, for any function $f$,
	\begin{align*}
		&\quad P^{\omega_{J}}\cdots P^{\omega_1}f(\bx, a)\\
		&=\int_{\cX \times \cA}\cdots\int_{\cX \times \cA}g_1(\bx_2, a_2;f)\rho( \bx_2 \mid  \bx_{3}, a_{3})\omega_2(a_2\mid \bx_2)\mathrm{d}(\bx_{2}, a_2) \cdots \rho( \bx_J \mid  \bx, a)\omega_J(a_J\mid \bx_J)\mathrm{d}(\bx_J, a_J).
	\end{align*}
	where $g_1(\bx_2, a_2;f):=\int_{\cX \times \cA}f(\bx_{1}, a_1)\rho( \bx_1 \mid  \bx_2, a_2)\omega_1(a_1\mid \bx_1)\mathrm{d}(\bx_{1}, a_1)$. Assumption \ref{assum:kappa-smooth} implies that $\rho \leq \overline{\rho}$ for some constant $\overline{\rho}$, and thus, with probability approaching one,
	\begin{align*}
		&\quad g_1\left(\bx, a; \abs{\calT^{(0)} \hat Q^{(0)}_{\tau-1}-\hat Q^{(0)}_{\tau}} \right)\\
		&\lesssim  \int_{\calX} \sum_{a' \in \calA} \left[\bxi^{\top}(\bx', a')\left(\hat\bbeta_{\tau}^{(0)}-\bbeta_{\tau}^{(0)}\right) + p^{-\kappa/d}\right]\rho( \bx' \mid  \bx, a)\omega_1(a'\mid \bx') \mathrm{d}\bx'\\
		&\lesssim p^{-\kappa/d} + \sqrt{\left(\hat\bbeta_{\tau}^{(0)}-\bbeta_{\tau}^{(0)}\right)^{\top}\int_{\cX} \sum_{a'}\bxi(\bx', a')\bxi^{\top}(\bx', a')\rho( \bx' \mid  \bx, a)\omega_1(a'\mid \bx') \mathrm{d}\bx'\left(\hat\bbeta_{\tau}^{(0)}-\bbeta_{\tau}^{(0)}\right) }\\
		&\lesssim p^{-\kappa/d} +  \sup_{\tau \geq 1}\norm{\hat\bbeta_{\tau}^{(0)}-\bbeta_{\tau}^{(0)}}_2,
	\end{align*}
	which, together with \eqref{eq:Emub-homo-1},  implies that 
	\begin{align*}
		&\quad \EE_{\widetilde{\mu},\widetilde{\pi}}\abs{Q^{*}-\hat Q^{(0)}_{{\Upsilon}}}(\bx, a) \\ & \lesssim \frac{1}{1-\gamma} \left(\sqrt{\frac{p}{n_{\calK}}}+p^{-\kappa/d}+\sqrt{\frac{p \log p}{n_k}} \wedge 
		\sqrt{C_{\Sigma}h}\left(\frac{\log p}{n_k}\right)^{1/4}\wedge C_{\Sigma}h\right)+\frac{2\gamma^{\Upsilon}R_{\max}}{1-\gamma},
	\end{align*}
	for any $\widetilde{\mu},\widetilde{\pi}$. Then the desired result for $v^*-v^{\widehat{\pi}_{\Upsilon}}$ is obtained by applying Lemma 13 in \cite{chen2019information}.
\end{proof}

\paragraph{Proof for Theorem \ref{thm:p-select}}

\begin{proof}
	The proof of Theorem \ref{thm:Q-converge-homo} ensures that, with probability approaching one,  \[\EE_{\widetilde{\mu},\widetilde{\pi}}\abs{Q^{*}-\hat Q^{(0)}_{{\Upsilon, g}}}(\bx, a) \lesssim \frac{1}{1-\gamma} \left(\sqrt{\frac{p_g}{n_{\calK}}}+p_g^{-\kappa/d}+\sqrt{\frac{p_g \log p_g}{n_0}} \wedge 
	\sqrt{h_g}\left(\frac{\log p_g}{n_0}\right)^{1/4}\wedge h_g\right),\]
	for any distribution $\widetilde{\mu}$ with bounded density, policy $\widetilde{\pi}$, and sufficiently large $\Upsilon$, where $p_g:=2^g$, and $h_g$ denotes the corresponding task-discrepancy. Therefore, for any $p_{g'}>p_{g}$,
	\[\EE_{\widetilde{\mu},\widetilde{\pi}}\abs{\hat Q^{(0)}_{{\Upsilon, g'}}-\hat Q^{(0)}_{{\Upsilon, g}}}(\bx, a) \lesssim \frac{1}{1-\gamma} \left(\sqrt{\frac{p_{g'}}{n_{\calK}}}+p_{g}^{-\kappa/d}+\sqrt{\frac{p_{g'} \log p_{g'}}{n_0}} \wedge (\Delta_{p_g} \vee \Delta_{p_{g'}})\right),\]
	which implies that 
	\begin{equation}\label{eq:error-bound-rho}
		\int_{\calX}\sum_{a\in \calA}\abs{\hat Q^{(0)}_{{\Upsilon, g'}}-\hat Q^{(0)}_{{\Upsilon, g}}}(\bx, a)\mathrm{d}\bx \lesssim \frac{1}{1-\gamma} \left(\sqrt{\frac{p_{g'}}{n_{\calK}}}+p_{g}^{-\kappa/d}+\sqrt{\frac{p_{g'} \log p_{g'}}{n_0}}\wedge(\Delta_{p_g} \vee \Delta_{p_{g'}})\right),\
	\end{equation}
	by picking $\widetilde{\mu}$ to be the uniform distribution on $\calX$ and $\widetilde{\pi}$ to be the random policy. 
	
	Let  $g^*:=\Big\lceil\log_2\big(n_{\calK}^{\frac{d}{d+2\kappa}}\big)\Big\rceil$. For any $g \in \widetilde{\calP}$, it holds that $g\geq g^*$. Otherwise, if $g\leq g^*-1$, then
	\[p_{g}^{-\kappa/d}\geq p_{g^*-1}^{-\kappa/d}> n_{\calK}^{-\kappa/(d+2\kappa)} > \sqrt{p_{g^*-1}/n_{\calK}} \geq \sqrt{p_{g}/n_{\calK}},\]
	contradicting the definition of $\widetilde\calP$. Therefore, we have $g^*\leq \underline{g} \leq \widetilde{g}_{\max}$.
	%By the assumption $\widetilde{g}_{\max} \in \widetilde\calP$, we have $g^* \leq \widetilde{g}_{\max} \leq g_{\max}$; otherwise, if $g^* > \widetilde{g}_{\max}$, then $\sqrt{p_{\widetilde{g}_{\max}}/n_{\calK}} \ll p_{\widetilde{g}_{\max}}^{-\kappa/d}$, contradicting the definition of $\widetilde\calP$.% and $\Delta_{p_{\overline{g}-1}}\leq\Delta_{p_{\overline{g}}} \leq C\sqrt{p_{\overline{g}}/n_{\calK}}$ for some constant $C$. Therefore, with probability approaching one, 
	%\[\int_{\calX}\sum_{a\in \calA}\abs{\hat Q^{(0)}_{{\Upsilon, \overline{g}}}-\hat Q^{(0)}_{{\Upsilon, \overline{g}-1}}}(\bx, a)\mathrm{d}\bx \lesssim \frac{1}{1-\gamma} \sqrt{\frac{p_{\overline{g}}}{n_{\calK}}},\]
	%and thus $\widetilde{\calG} \neq \emptyset$, with $\widetilde{g}_{\max}=\max \widetilde{\calG}\geq \overline{g}  \geq g^*$. 
	
	If $\underline{g} = g^*$, we have  $\{p_{g^*}, p_{g^*+1}, \dots, p_{\widetilde{g}_{\max}}\} \subset \widetilde\calP$. For any $g'$ such that $\widetilde{g}_{\max} \geq g' \geq g^*$,
	\[p_{g^*}^{-\kappa/d} \leq n_{\calK}^{-\kappa/(d+2\kappa)} \leq  \sqrt{p_{g^*}/n_{\calK}} \leq \sqrt{p_{g'}/n_{\calK}},\]
	and, by the definition of $\widetilde{\calP}$, it holds that $\sqrt{p_{g'}/n_{\calK}} \geq \Delta_{p_{g'}}$ and $\sqrt{p_{g'}/n_{\calK}} \geq\sqrt{p_{g^*}/n_{\calK}} \geq \Delta_{p_{g^*}}$. Hence, by \eqref{eq:error-bound-rho},
	\[\int_{\calX}\sum_{a\in \calA}\abs{\hat Q^{(0)}_{{\Upsilon, g^*}}-\hat Q^{(0)}_{{\Upsilon, g'}}}(\bx, a)\mathrm{d}\bx \lesssim \frac{1}{1-\gamma} \sqrt{\frac{p_{g'}}{n_{\calK}}}.\]
	% then the above inequalities imply that  
	%\[\int_{\calX}\sum_{a\in \calA}\abs{\hat Q^{(0)}_{{\Upsilon, g^*}}-\hat Q^{(0)}_{{\Upsilon, g'}}}(\bx, a)\mathrm{d}\bx \lesssim \frac{1}{1-\gamma}\sqrt{\frac{p_{g'}}{n_{\calK}}},\]
	%for any $g'$ such that $\widetilde{g}_{\max} \geq g' \geq g^*$. 
	Then by the definition of $\widehat{g}$, we have $\widehat{g} \leq g^*$ with probability approaching one, and hence
	\begin{equation}\label{eq:error-g-hat-1}
		\begin{aligned}
			\EE_{\widetilde{\mu},\widetilde{\pi}}\abs{Q^{*}-\hat Q^{(0)}_{{\Upsilon, \widehat{g}}}}(\bx, a) &\lesssim \EE_{\widetilde{\mu},\widetilde{\pi}}\abs{Q^{*}-\hat Q^{(0)}_{{\Upsilon, g^*}}}(\bx, a) + \EE_{\widetilde{\mu},\widetilde{\pi}}\abs{Q^{(0)}_{{\Upsilon, g^*}}-\hat Q^{(0)}_{{\Upsilon, \widehat{g}}}}(\bx, a)\\
			&\lesssim\frac{1}{1-\gamma} \left(\sqrt{\frac{p_{g^*}}{n_{\calK}}}+p_{g^*}^{-\kappa/d}+ \overline{\Delta}\right) + \frac{1}{1-\gamma} \sqrt{\frac{p_{g^*}}{n_{\calK}}}\\
			&\lesssim \frac{1}{1-\gamma} \cdot n^{-\kappa/(2\kappa+d)}.
		\end{aligned}
	\end{equation}
	On the other hand, if $\underline{g} > g^*$, %then let $g_{\overline{\Delta}}:=\max \left\{g: \sqrt{p_g/n_{\calK}} \leq \Delta_p, g \leq \widetilde{g}_{\max}\right\}$, which satisfies $g_{\overline{\Delta}} \geq g^*$ since $\sqrt{p_{g^*}/n_{\calK}} \leq \sqrt{2}n_{\calK}^{-\kappa/(2d+\kappa)}< \overline{\Delta}$. %, and $g_{\overline{\Delta}}< \widetilde{g}_{\max}$ by \eqref{eq:p-select} holds trivially. Otherwise, 
	for any $g'$ such that $\widetilde{g}_{\max} > g'>\underline{g}$, it holds that
	$\sqrt{\frac{p_{g'}}{n_{\calK}}} \geq\sqrt{\frac{p_{\underline{g}}}{n_{\calK}}} \geq p_{\underline{g}}^{-\kappa/d}$. The definition of $\widetilde{\calP}$ still ensures that $\sqrt{p_{g'}/n_{\calK}} \geq \Delta_{p_g'}$.  and thus
	\[\int_{\calX}\sum_{a\in \calA}\abs{\hat Q^{(0)}_{{\Upsilon, g'}}-\hat Q^{(0)}_{{\Upsilon, \underline{g}}}}(\bx, a)\mathrm{d}\bx \lesssim \frac{1}{1-\gamma} \sqrt{\frac{p_{g'}}{n_{\calK}}}.\]
	Therefore, it holds that $\widehat{g} \leq \underline{g}$ with probability approaching one, and similar to \eqref{eq:error-g-hat-1}, 
	\[	\EE_{\widetilde{\mu},\widetilde{\pi}}\abs{Q^{*}-\hat Q^{(0)}_{{\Upsilon, \widehat{g}}}}(\bx, a) \lesssim \frac{1}{1-\gamma} \sqrt{\frac{p_{\underline{g}}}{n_{\calK}}}\leq \sqrt{2}\sqrt{\frac{p_{\underline{g}-1}}{n_{\calK}}}.\]
	Moreover, since $\underline{g} -1 \notin \widetilde{\calP}$, it holds that \[\sqrt{\frac{p_{\underline{g}-1}}{n_{\calK}}}< \big(\Delta_{p_{\underline{g}-1}} \vee p_{\underline{g}-1}^{-\kappa/d}\big).\]
	Note that $\underline{g} -1 \geq g^*$ implies $p_{\underline{g}-1}^{-\kappa/d} \leq p_{g^*}^{-\kappa/d} \leq \sqrt{p_{\underline{g}-1}/n_{\calK}}$, which yields that $\sqrt{\frac{p_{\underline{g}-1}}{n_{\calK}}}<\Delta_{p_{\underline{g}-1}}$.
	Combining these inequalities with \eqref{eq:error-g-hat-1} leads to $\EE_{\widetilde{\mu},\widetilde{\pi}}\abs{Q^{*}-\hat Q^{(0)}_{{\Upsilon, \widehat{g}}}}(\bx, a) \lesssim \frac{1}{1-\gamma} \left(n^{-\frac{\kappa}{d+2\kappa}}\vee\Delta_{p_{\underline{g}-1}}\right)$. The proof is then concluded using the same argument in the proof of Theorem \ref{thm:Q-converge-homo}.
\end{proof}

\section{Details for the Real Data Analysis}\label{sec:supp-real}

In this section, we provide implementation details for the real data analysis presented in Section \ref{sec:appl}. For tasks $k=0, 1$, we create calibrated environments using the complete datasets, with sample sizes $n^{(0)}=156,064$ and $n^{(1)}=122,534$. We estimate the reward function $r^{(k)}(\bx, a)$ using kernel ridge regression with observable response $R_{i, t}^{(k)}$ and predictors $\big(\bX_{i, t}^{(k)}, A_{i, t}^{(k)}\big)$. For the transition kernel $\rho^{(k)}(\bx' \,|\, \bx, a)$, we employ the least-squares approach to conditional density estimation proposed by \cite{sugiyama2010conditional} and implemented by \cite{rothfuss2019conditional}. For the initial state distribution $\nu^{(k)}$, we use a multivariate normal distribution whose mean and covariance matrix match the sample mean and covariance matrix of all states in task $k$.

To evaluate the performance of different methods, we generate target and source tasks from the calibrated environments with target size $I^{(0)}\in\{100, 500\}$ and source size $I^{(1)} \in [250, 2000]$, and obtain the estimated optimal policies. We then estimate the value of an estimated policy $\widehat{\pi}$ by averaging the cumulative returns from 500 independent trajectories generated from the calibrated environment according to $\widehat{\pi}$. For baseline comparison, we estimate the optimal policy $\pi^*$ using FQI on the complete target dataset and view it as the ground truth. Across all experiments, we set the discount parameter to $\gamma=0.6$, use 10-dimensional B-splines as basis functions $\bphi(\bx)$, and select the regularization parameters ${\lambda_{\delta}^{(k)}}$ through cross-validation.

\end{document}